%% file: main.tex
\documentclass[10pt,twocolumn,letterpaper]{article}

\usepackage[pagenumbers]{wacv} %

\usepackage{graphicx}
\usepackage{amsmath}
\usepackage{amssymb}
\usepackage{booktabs}
\usepackage{wrapfig}	
\usepackage{chngcntr}
\usepackage[dvipsnames]{xcolor}

\usepackage[accsupp]{axessibility}  %

\usepackage{xcolor}  %
\usepackage{multirow} %
\usepackage{tikz}
\usetikzlibrary{calc}
\usepackage{titletoc}     %

\usepackage{amssymb}%
\usepackage{pifont}%
\newcommand{\cmark}{\ding{51}}%
\newcommand{\xmark}{\ding{55}}%

\usepackage{microtype}
\usepackage[accsupp]{axessibility} 

\usepackage[pagebackref,breaklinks,colorlinks]{hyperref}

\usepackage[capitalize]{cleveref}
\crefname{section}{Sec.}{Secs.}
\Crefname{section}{Section}{Sections}
\Crefname{table}{Table}{Tables}
\crefname{table}{Tab.}{Tabs.}

\def\model{LIME} \title{\model{}: Localized Image Editing via Attention Regularization \\ in Diffusion Models}

\author{\vspace{1mm}
    Enis Simsar$^{1}$
    \hspace{0.5cm}
    Alessio Tonioni$^{3}$
    \hspace{0.5cm}
    Yongqin Xian$^{3}$
    \hspace{0.5cm}
    Thomas Hofmann$^{1}$
    \hspace{0.5cm}
    Federico Tombari$^{2,3}$
    \\
    $^1$ETH Zürich - DALAB
    \hspace{2.5em} 
    $^2$Technical University of Munich
    \hspace{2.5em} 
    $^3$Google Switzerland
}
\begin{document}

\newlength{\defaultcolumnsep}
\setlength{\defaultcolumnsep}{\columnsep}

\input{tex_figures/teaser}

\input{sections/0_abstract}    
\input{sections/1_intro}

\input{sections/2_related_work}

\input{sections/3_background}

\input{sections/4_method}
\input{sections/5_experiments}

\input{sections/6_conclusion}

{\small
\bibliographystyle{ieee_fullname}
\bibliography{main}
}

\clearpage
\setlength{\columnsep}{\defaultcolumnsep}

\maketitlesupplementary

\input{sections/7_suppl}

\clearpage

\end{document}

%% file: tex_figures/teaser.tex
\twocolumn[{

    \small        
    \renewcommand\twocolumn[1][]{#1}%
    \maketitle

    \vspace{-2.em}

    \centering
    \tikz[remember picture,overlay] \node [anchor=base] (linebase) at ([xshift=0\linewidth]0,0) {};

    \begin{minipage}{.15\textwidth}
        \centering
        \textbf{Input}
    \end{minipage}
    \begin{minipage}{.15\textwidth}
        \centering
        \textbf{IP2P}
    \end{minipage}
    \begin{minipage}{.15\textwidth}
        \centering
        \textbf{+ \model{}}
    \end{minipage}
    \hfill
    \begin{minipage}{.045\textwidth}
        \centering
    \end{minipage}
    \hfill
    \begin{minipage}{.15\textwidth}
        \centering
        \textbf{Input}
    \end{minipage}
    \begin{minipage}{.15\textwidth}
        \centering
        \textbf{IP2P w/MB}
    \end{minipage}
    \begin{minipage}{.15\textwidth}
        \centering
        \textbf{+ \model{}}
    \end{minipage}
    
    \vspace{0.2em}
    
    \begin{minipage}{.15\textwidth}
        \centering
        \includegraphics[width=\textwidth]{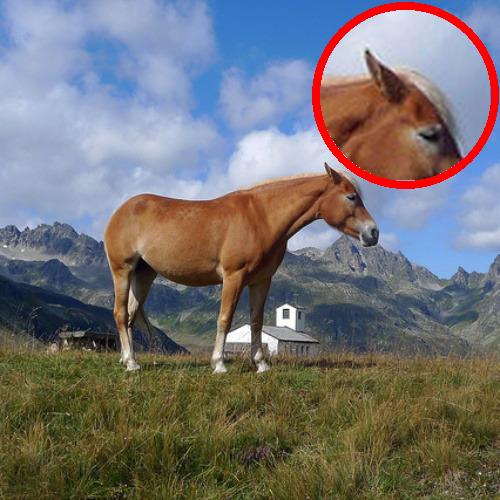}
    \end{minipage}
    \begin{minipage}{.15\textwidth}
        \centering
        \includegraphics[width=\textwidth]{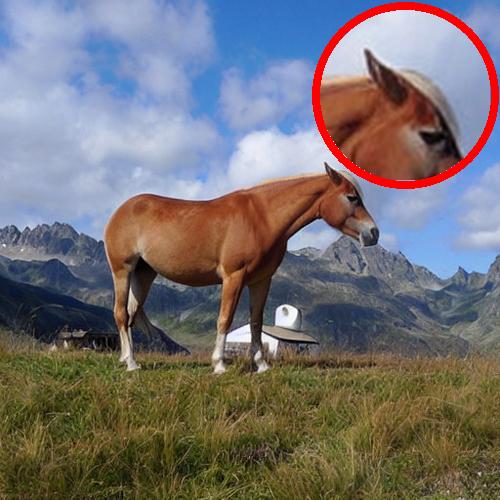}
    \end{minipage}
    \begin{minipage}{.15\textwidth}
        \centering
        \includegraphics[width=\textwidth]{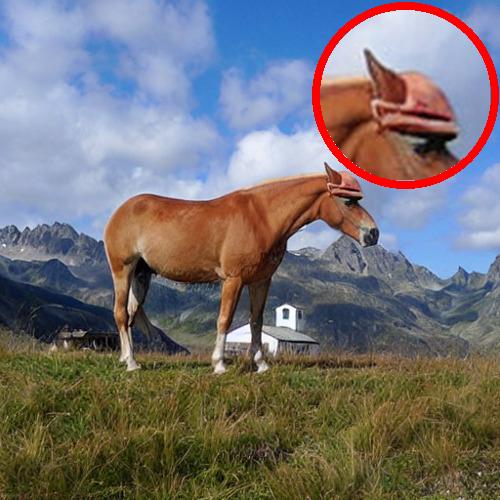}
    \end{minipage}
    \hfill
    \begin{minipage}{.045\textwidth}
        \centering
    \end{minipage}
    \hfill
    \begin{minipage}{.15\textwidth}
        \centering
        \includegraphics[width=\textwidth]{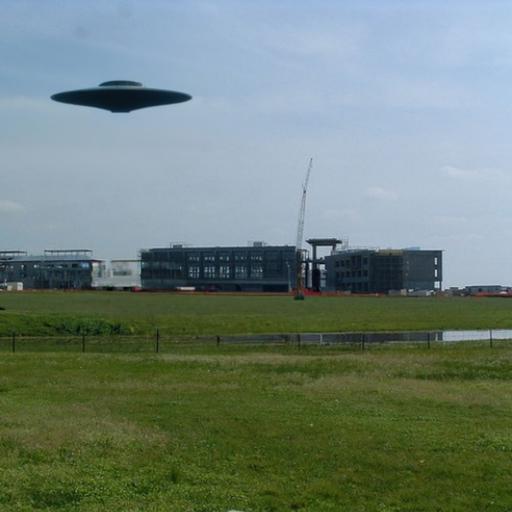}
    \end{minipage}
    \begin{minipage}{.15\textwidth}
        \centering
        \includegraphics[width=\textwidth]{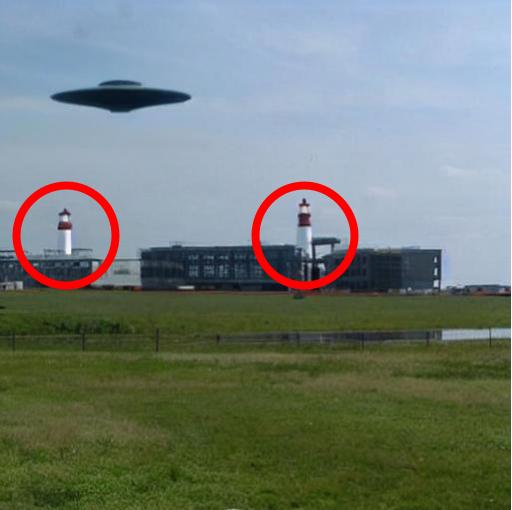}
    \end{minipage}
    \begin{minipage}{.15\textwidth}
        \centering
        \includegraphics[width=\textwidth]{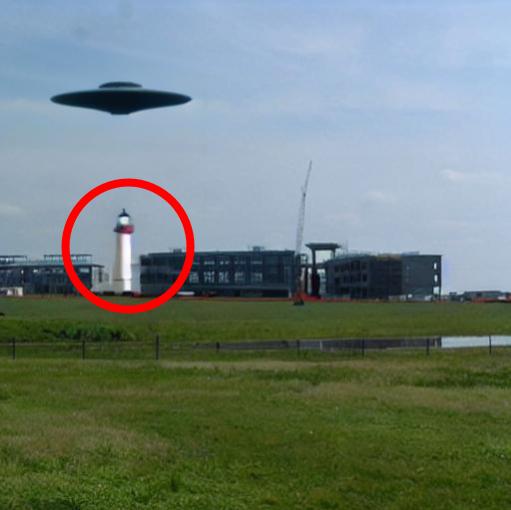}
    \end{minipage}
    
    \vspace{0.4em}
    
    \begin{minipage}{.48\textwidth}
        \centering
        (a) \textit{What if the horse were wearing a hat?}
    \end{minipage}
    \hfill
    \begin{minipage}{.045\textwidth}
        \centering
    \end{minipage}
    \hfill
    \begin{minipage}{.48\textwidth}
        \centering
        (b) \textit{Put a lighthouse under UFO.}    
    \end{minipage}
    
    \vspace{-0.25em}
    
    \begin{tikzpicture}
        \draw[dashed, very thick] (0,0) -- (\textwidth,0);
    \end{tikzpicture}

    \begin{minipage}{.15\textwidth}
        \centering
        \textbf{Input}
    \end{minipage}
    \begin{minipage}{.15\textwidth}
        \centering
        \textbf{IP2P}
    \end{minipage}
    \begin{minipage}{.15\textwidth}
        \centering
        \textbf{+ \model{}}
    \end{minipage}
    \hfill
    \begin{minipage}{.045\textwidth}
        \centering
    \end{minipage}
    \hfill
    \begin{minipage}{.15\textwidth}
        \centering
        \textbf{Input}
    \end{minipage}
    \begin{minipage}{.15\textwidth}
        \centering
        \textbf{IP2P w/MB}
    \end{minipage}
    \begin{minipage}{.15\textwidth}
        \centering
        \textbf{+ \model{}}
    \end{minipage}
    
    \vspace{0.2em}
    
    \begin{minipage}{.15\textwidth}
        \centering
        \includegraphics[width=\textwidth]{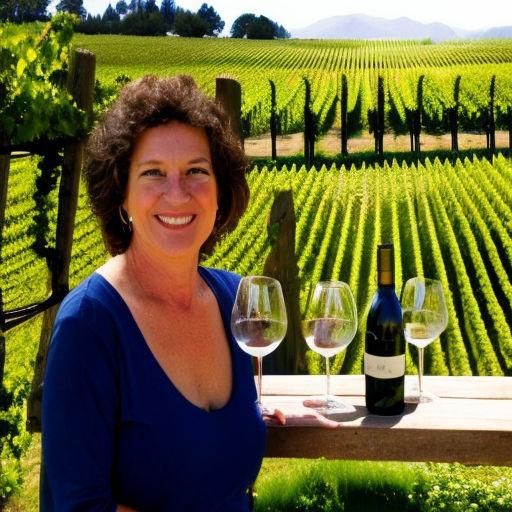}
    \end{minipage}
    \begin{minipage}{.15\textwidth}
        \centering
        \includegraphics[width=\textwidth]{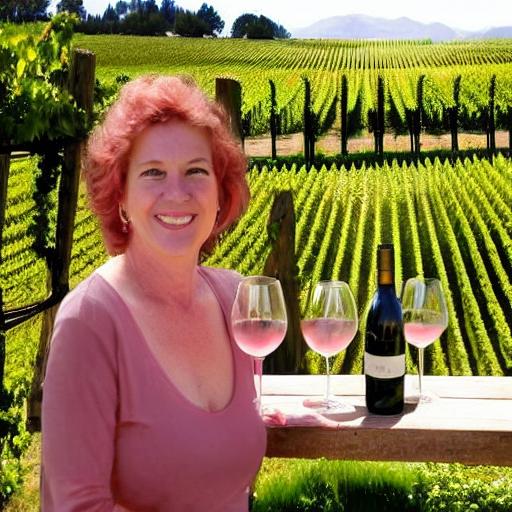}
    \end{minipage}
    \begin{minipage}{.15\textwidth}
        \centering
        \includegraphics[width=\textwidth]{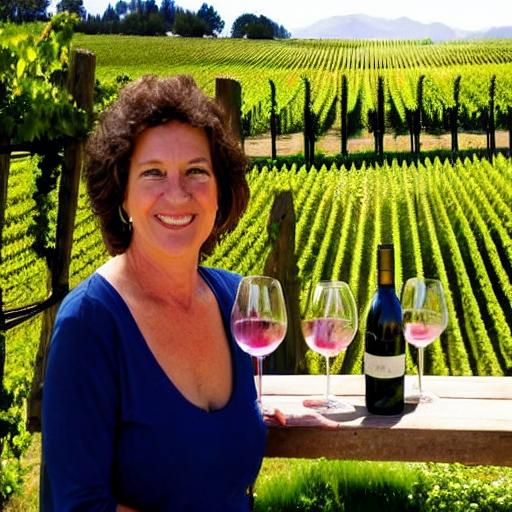}
    \end{minipage}
    \hfill
    \begin{minipage}{.045\textwidth}
        \centering
    \end{minipage}
    \hfill
    \begin{minipage}{.15\textwidth}
        \centering
        \includegraphics[width=\textwidth]{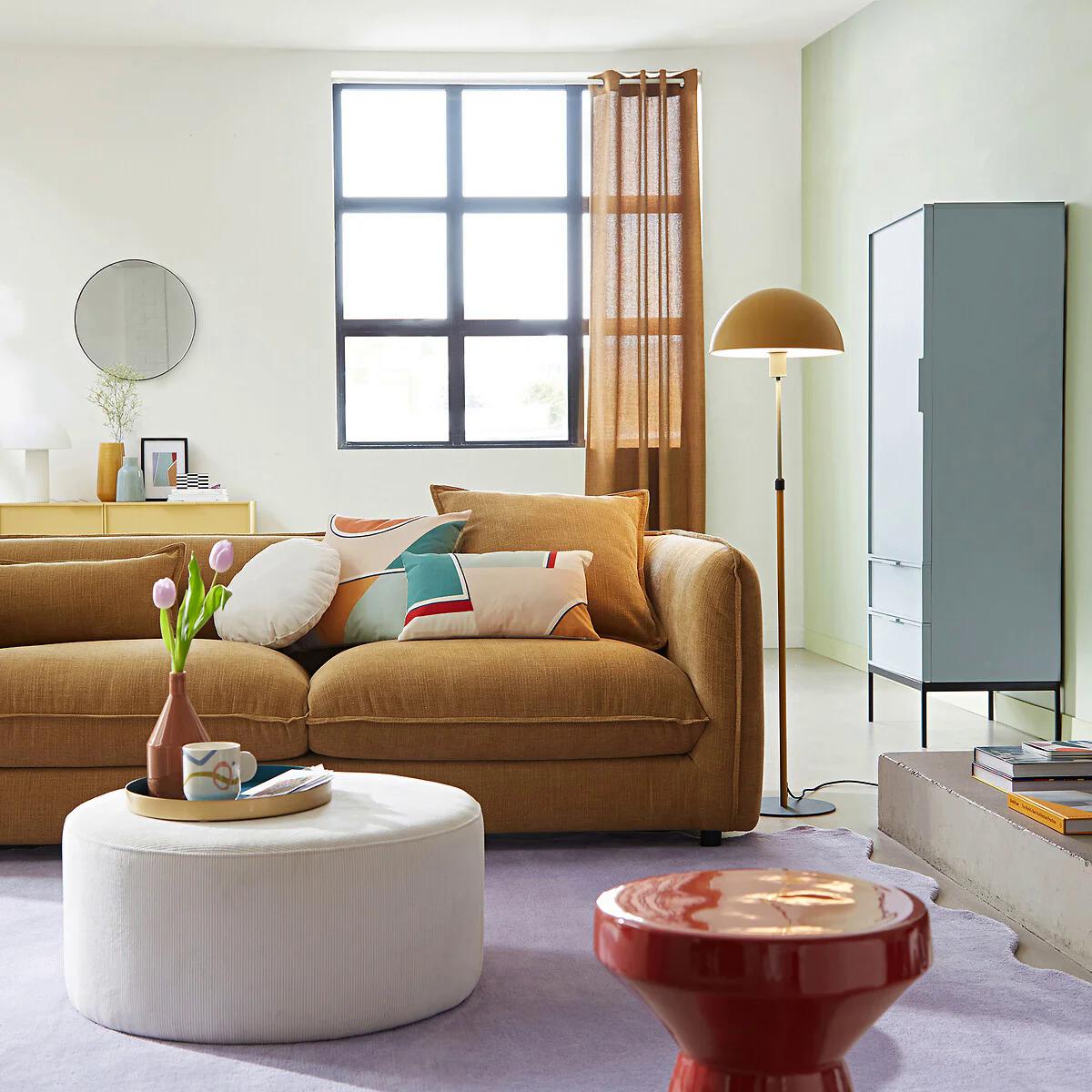}
    \end{minipage}
    \begin{minipage}{.15\textwidth}
        \centering
        \includegraphics[width=\textwidth]{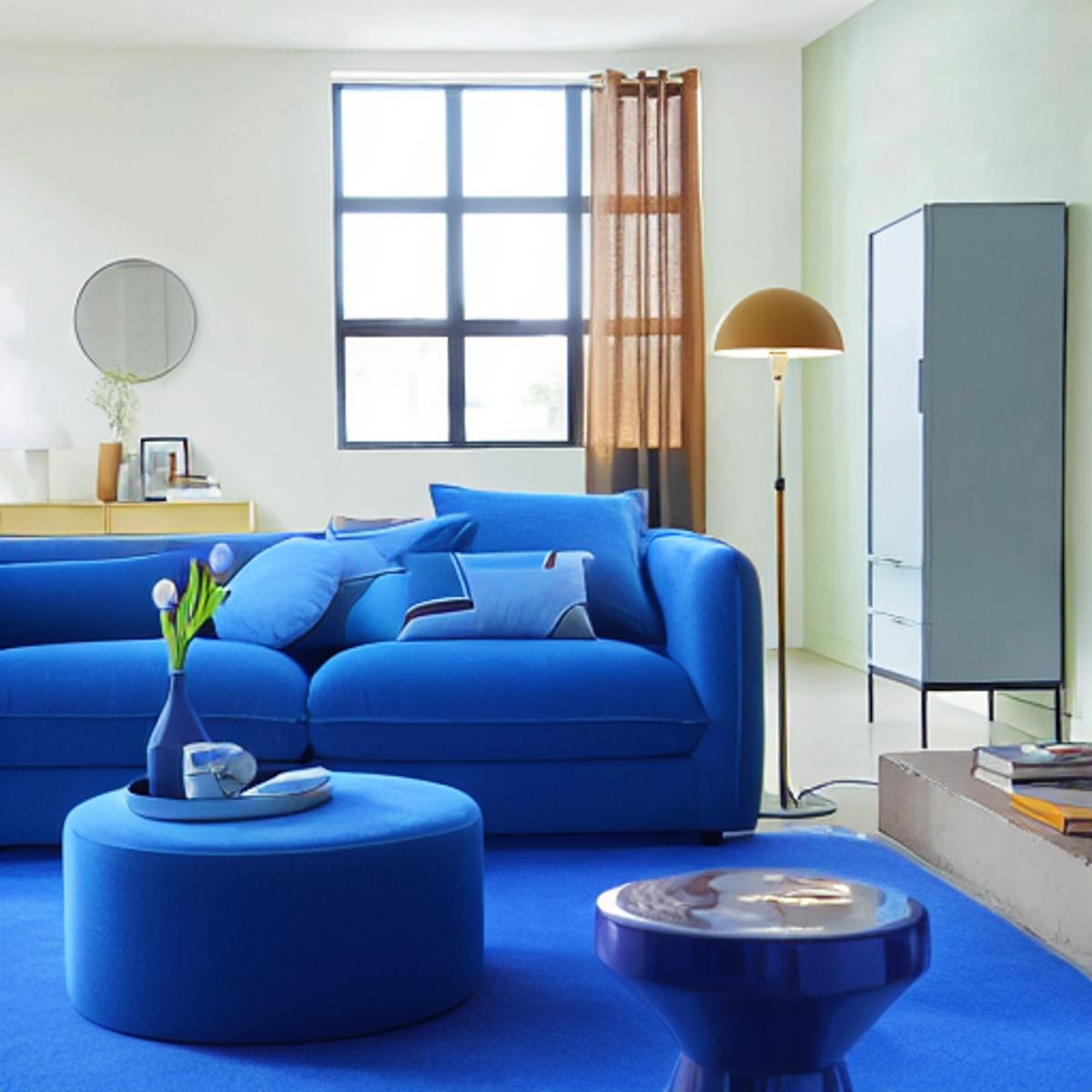}
    \end{minipage}
    \begin{minipage}{.15\textwidth}
        \centering
        \includegraphics[width=\textwidth]{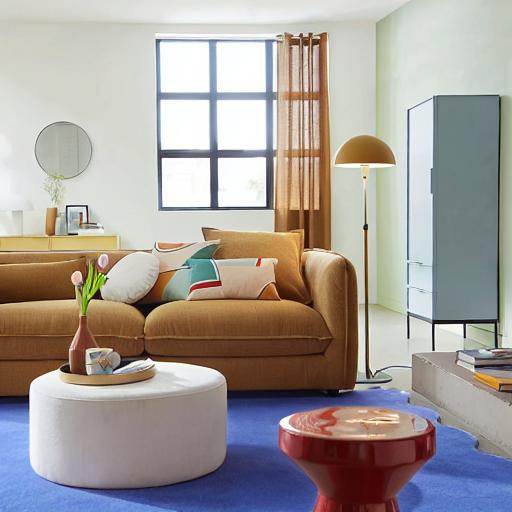}
    \end{minipage}
    
    \vspace{0.4em}
    
    \begin{minipage}{.48\textwidth}
        \centering
        (c) \textit{Change to a rosé.}
    \end{minipage}
    \hfill
    \begin{minipage}{.045\textwidth}
        \centering
    \end{minipage}
    \hfill
    \begin{minipage}{.48\textwidth}
        \centering
        (d) \textit{Change color of carpet to dark blue.} %
    \end{minipage}
    
    \begin{tikzpicture}[remember picture, overlay]
        \draw[dashed, very thick] (linebase) -- ++(0,-7.5); %
    \end{tikzpicture}
    
    \vspace{-1.2em}

    \captionof{figure}{\textbf{\model{}: \underline{L}ocalized \underline{IM}age \underline{E}diting.} \model{} edits an image based on an edit instruction without needing customized datasets, fine-tuning, or explicit information about the object of interest. The addition of \model{} improves InstructPix2Pix~(IP2P)~\cite{Brooks2022InstructPix2Pix} and its fine-tuned version on MagicBrush~(MB)~\cite{Zhang2023MagicBrush}, human-annotated, dataset and allows localized edits preserving the rest of the image untouched.} \vspace{1.5em}
    \label{fig:teaser}
}]

%% file: sections/0_abstract.tex
\begin{abstract}
    Diffusion models~(DMs) have gained prominence due to their ability to generate high-quality varied images with recent advancements in text-to-image generation. The research focus is now shifting towards the controllability of DMs. A significant challenge within this domain is localized editing, where specific areas of an image are modified without affecting the rest of the content. This paper introduces \model{} for localized image editing in diffusion models. \model{} does not require user-specified regions of interest~(RoI) or additional text input, but rather employs features from pre-trained methods and a straightforward clustering method to obtain precise editing mask. Then, by leveraging cross-attention maps, it refines these segments for finding regions to obtain localized edits. Finally, we propose a novel cross-attention regularization technique that penalizes unrelated cross-attention scores in the RoI during the denoising steps, ensuring localized edits. Our approach, without re-training, fine-tuning and additional user inputs, consistently improves the performance of existing methods in various editing benchmarks. The project page can be found at \href{https://enisimsar.github.io/LIME/}{\textit{https://enisimsar.github.io/LIME/}}.
\end{abstract}

%% file: sections/1_intro.tex
\section{Introduction}

Diffusion models~(DMs) have recently achieved remarkable success in generating images that are not only high-quality but also richly varied, thanks to advancements in text-to-image conversion~\cite{ho2020denoising, Saharia2022Imagen, Rombach2022StableDiffusion, Ramesh2022DALLE2}. Beyond their generative capabilities, there is a growing research interest in the controllability aspect of these models~\cite{Hertz2022Prompt2prompt, Brooks2022InstructPix2Pix, Zhang2023HIVE, couairon2023diffedit, Avrahami2022BlendedDiffusion, patashnik2023localizing}. This has led to the exploration of a variety of editing techniques, leveraging the power of DMs for tasks such as personalized image creation~\cite{Ruiz2022DreamBooth, wei2023elite, gal2022image}, context-aware inpainting \cite{lugmayr2022repaint, Nichol2022GLIDE, yang2023paint}, and image transformation in response to textual edits \cite{Avrahami2022BlendedDiffusion, Hertz2022Prompt2prompt, Brooks2022InstructPix2Pix, Meng2022SDEdit, Kawar2022Imagic, couairon2023diffedit}. These developments underscore the versatility of DMs and their potential to serve as foundational tools for various image editing applications.

In this paper, we address the task of text-guided image editing, explicitly focusing on localized editing, which refers to \textit{identifying} and \textit{modifying} any region of interest in an image while preserving the context of the surrounding regions. In particular, we focus in the settings where this needs to be done relying only on textual instructions. The difficulty arises from the intertwined nature of image representations within these models, where changes intended for one area can inadvertently affect others~\cite{Hertz2022Prompt2prompt, Meng2022SDEdit, Zhang2023HIVE, Brooks2022InstructPix2Pix}. Existing methods often depend on additional user input, such as masking the target area, \ie{,} Region of Interest~(RoI), or providing additional text information, \eg{,} objects of interest, to pinpoint the editing region~\cite{Avrahami2022BlendedDiffusion,couairon2023diffedit,guo2023focus}. However, these approaches introduce complexity and do not guarantee the precision necessary for seamless editing. \Cref{fig:teaser} highlights %
localized edits without altering the overall image, a balance that current methods have not yet struck. Advancing localized editing to be more effective remains a challenging research direction especially when minimizing the need for extensive re-training, fine-tuning, or additional user inputs.

We address the challenge of localized image editing by introducing \emph{\model{}}, that leverages pre-trained InstructPix2Pix~(IP2P)~\cite{Brooks2022InstructPix2Pix} without the need for additional supervision, user inputs, or model re-training/fine-tuning. Recent studies \cite{xu2023odise, PNVR2023Ldznet, tang2023emergent} have demonstrated that diffusion models are capable of encoding semantics within their intermediate features. \model{} utilizes those features to identify segments, then extracts RoI by utilizing attention scores derived from instructions. Other research \cite{chefer2023attend, Agarwal2023astar,meral2024conform} has shown the significant impact of attention-based guidance on image generation tasks. Accordingly, \model{} aims to restrict the scope of edits by regularizing attention scores to enable disentangled and localized edits. 
By incorporating these two lines of work, \model{} not only offers more effective localized editing as shown in \cref{fig:teaser} but also demonstrates a notable advancement. This advancement is quantitatively measured by outperforming current state-of-the-art methods on two different benchmark datasets.  While we primarily benchmark against IP2P, the method's principles can be adapted to other relevant approaches, see \textit{Supplementary Material}.

Our pipeline contains two steps. (i) It first finds semantic segments of the input image. This is achieved based on semantic information encoded in intermediate features. Then, we identify the area to be edited by combining the segments with significant cross-attention scores toward the edit instruction. (ii) Once we isolate the area to be edited, \ie{,} RoI, the proposed attention regularization is applied to the text tokens cross-attention scores to ensure that the edit is focused to the RoI, avoiding unintended changes to other parts of the image. 
This two-step approach, first identifying targeted areas and then editing within them, ensures that the edits are local and contextually coherent, simplifying the editing process while avoiding unintended alterations to the rest of the image. The core contributions of this study:

\begin{itemize}
    \item We introduce a localized image editing method, \model{}, that eliminates the need for fine-tuning, re-training or additional user inputs, ensuring efficient and precise localized edits.
    \item We propose a novel attention regularization strategy to ensure that our edits are not only localized but also contextually coherent within the RoI, resulting in seamless and targeted image modifications.
    \item Our approach leverages the pre-trained instruction-based editing model's intermediate features to segment the image and to identify the regions where modifications will be applied.
\end{itemize}

\noindent The experimental evaluation demonstrates \model{} outperforms existing methods in localized editing both qualitatively and quantitatively on two benchmark datasets~\cite{Zhang2023MagicBrush, Brooks2022InstructPix2Pix}.

%% file: sections/2_related_work.tex
\section{Related Work}

\noindent\textbf{Image editing with Diffusion Models.}
One direction for image editing is utilizing pre-trained diffusion models by first inverting the input image in the latent space and then applying the desired edit by altering the text prompt~\cite{Mokady2022NullTextInversion, Hertz2022Prompt2prompt, wang2023mdp, Meng2022SDEdit, couairon2023diffedit, ju2023direct, parmar2023zero, wang2023dynamic, wu2023uncovering}. For instance, the input image can be inverted to the diffusion model latent space via Null-text inversion~\cite{Mokady2022NullTextInversion}, PnPInversion~\cite{ju2023direct} or any other inversion methods and then Prompt-to-Prompt~\cite{Hertz2022Prompt2prompt} can be applied to obtain the desired edit. Prompt-to-Prompt also introduces an attention modulation technique to control the edit by \textit{globally} modifying the attention maps of specific tokens. In contrast, the attention regularization of \model{} relies on the RoI and thereby is \textit{local} in the image space. We show that our attention regularization is applicable to Prompt-to-Prompt for improved localized edits in \textit{Supplementary Material}. DiffEdit~\cite{couairon2023diffedit}, on the other hand, matches the differences in predictions for input and output captions that are initialized from the inversion to localize the edit yet it struggles with complex instructions. Moreover, edits are performed by interpolations in the noise space instead of the attention modulation we propose. The key challenge of these methods is the inversion of the real image because it may lose details of the input image.

On the other hand, another direction for image editing by using instructions that eliminates the need for inversion is training diffusion models on triplet data, which contains input image, instruction, and edited image~\cite{Brooks2022InstructPix2Pix, Zhang2023MagicBrush, Zhang2023HIVE, fu2023guiding}. The most common approach, IP2P~\cite{Brooks2022InstructPix2Pix}, performs better than previous methods but sometimes generates entangled edits. To tackle this problem, MagicBrush~(IP2P w/MB)~\cite{Zhang2023MagicBrush} fine-tunes IP2P on a more localized human-annotated dataset, and HIVE~\cite{Zhang2023HIVE} relies on human feedback on edited images to learn what users generally prefer and uses this information to fine-tune IP2P, aiming to align more closely with human expectations.
Concurrently to our work, \cite{guo2023focus} proposed addressing the same entanglement problem with localized image editing by using grounding properties of IP2P. 
However, compared to \model{}, it requires an additional user input, \eg{,} the object of interest to edit. Moreover, our method performs edits via attention regularization rather than manipulating the noise space~\cite{couairon2023diffedit, Avrahami2022BlendedDiffusion, mirzaei2023watch}.

\noindent\textbf{Semantics in Diffusion Models.} 
Intermediate features of diffusion models, as explored in studies like~\cite{tang2023emergent, xu2023odise, PNVR2023Ldznet, patashnik2023localizing}, have been shown to encode semantic information. Recent research such as LD-ZNet~\cite{PNVR2023Ldznet} and ODISE~\cite{xu2023odise} leverages intermediate features of these models for training networks for semantic segmentation. Localizing Prompt Mixing~(LPM)~\cite{patashnik2023localizing}, on the other hand, utilizes clustering on self-attention outputs for segment identification. Motivated by this success, our method leverages pre-trained intermediate features to achieve semantic segmentation and apply localized edits using edit instructions.

%% file: sections/3_background.tex
\section{Background}\label{sec:background}
\noindent\textbf{Latent Diffusion Models.}
Stable Diffusion~(SD)~\cite{Rombach2022StableDiffusion} is a Latent Diffusion Model~(LDM) designed to operate in a compressed latent space. This space is defined by a pre-trained variational autoencoder~(VAE) to enhance computational efficiency. Gaussian noise is introduced into the latent space, generating samples from a latent distribution $z_t$. A U-Net-based denoising architecture~\cite{dhariwal2021diffusion} is then employed for image reconstruction, conditioned on noise input ($z_t$) and text conditioning ($c_T$). This reconstruction is iteratively applied over multiple time steps, each involving a sequence of self-attention and cross-attention layers. Self-attention layers transform the current noised image representation, while cross-attention layers integrate text conditioning.

Every attention layer comprises three components: Queries ($Q$), Keys ($K$), and Values ($V$). For cross-attention layers, $Q$s are obtained by applying a linear transformation $f_Q$ to the result of the self-attention layer preceding the cross-attention layer (\ie{,} image features). Similarly, $K$s and $V$s are derived from text conditioning $c_T$ using linear transformations $f_K$ and $f_V$. \Cref{eq:attention} shows the mathematical formulation of an attention layer where $P$ denotes the attention maps and is obtained as the softmax of the dot product of $K$ and $Q$ normalized by the square root of dimension $d$ of $K$s and $Q$s. 

\vspace{-1.2em}

\begin{equation}
\begin{split}
    \text{Attention}&(Q,K,V) = P \cdot V, \\
    &\text{where } P=\text{Softmax} \left( \frac{QK^T}{\sqrt{d}} \right).
    \label{eq:attention}
\end{split}
\end{equation}
\vspace{-0.8em}

Intuitively, $P$ denotes which areas of the input features will be modified in the attention layer. For cross-attention, it corresponds to the affected area by one of the conditioning text tokens that define $c_T$.
Beyond the attention maps, our approach leverages the output of transformer layers, noted as \textit{intermediate features} $\phi(z_t)$, which contain rich semantic content, as highlighted in recent studies~\cite{tang2023emergent, xu2023odise, PNVR2023Ldznet}. 
In this work, we modify the cross-attention's $P$ and leverage the intermediate features $\phi(z_t)$ to localize edits. %

\noindent\textbf{InstructPix2Pix.}
Our method relies on IP2P~\cite{Brooks2022InstructPix2Pix}, an image-to-image transformation network trained for text-conditioned editing. IP2P builds on top of Stable Diffusion and incorporates a bi-conditional framework, which simultaneously leverages an input image $I$, and an accompanying text-based instruction $T$ to steer the synthesis of the image, with the conditioning features being $c_I$ for the image and $c_T$ for the text. The image generation workflow is modulated through a classifier-free guidance~(CFG) strategy~\cite{ho2021classifierfree} that employs two separate coefficients, $s_T$ for text condition and $s_I$ for image condition. 
The noise vectors predicted by the learned network $e_\theta$, which corresponds to the individual U-Net step, with different sets of inputs, are linearly combined as represented in \cref{eq:cfg2} to achieve score estimate $\tilde{e}_{\theta}$. Our method utilizes and modifies the processes for the terms with $c_I$ in~\cref{eq:cfg2} to apply localized image editing.

\vspace{-1.2em}

\begin{equation}
\begin{split}
    \tilde{e}_{\theta}(z_t, c_I, c_T) = &\: e_{\theta}(z_t, \varnothing, \varnothing) \\ &+ s_I \cdot (e_{\theta}(z_t, c_I, \varnothing) - e_{\theta}(z_t, \varnothing, \varnothing)) \\ &+ s_T \cdot (e_{\theta}(z_t, c_I, c_T) - e_{\theta}(z_t, c_I, \varnothing)).
    \label{eq:cfg2}
\end{split}
\end{equation}

\vspace{-1.2em}

%% file: sections/4_method.tex
\section{Method}

We aim to develop a localized editing method for a \emph{pre-trained} IP2P \emph{without re-training, fine-tuning or the need for additional user inputs besides an edit prompt}. The proposed method contains two components: \textit{(i)} \emph{Edit Localization} finds the RoI by using the input image and the edit instruction, see~\cref{sec:edit_localization}, and \textit{(ii)} \emph{Edit Application} applies the instruction to RoI in a disentangled and localized manner, see~\cref{sec:edit_application}. 

\subsection{Edit Localization} \label{sec:edit_localization}

\noindent\textbf{Segmentation}
Our study extends the established understanding that intermediate features of diffusion models encode essential semantic information. In contrast to previous methods that build upon Stable Diffusion~\cite{tang2023emergent, PNVR2023Ldznet, xu2023odise}, our approach works on IP2P and focuses on the features conditioned on the original image (\(z_t\), \(c_I\), and \(\varnothing\)) for segmentation as indicated in \cref{eq:cfg2}. Through experimental observation, we show that these features align well with segmentation objectives for editing purposes. To obtain segmentation maps, we extract features from multiple layers of the U-Net architecture, including both down- and up-blocks, to encompass a variety of resolutions and enhance the semantic understanding. %

We implement a multi-resolution fusion strategy to refine feature representations within our proposed model and cluster them to extract segments. This involves \textit{(i)} resizing feature maps from various resolutions to a common resolution by applying bi-linear interpolation, \textit{(ii)} concatenating and normalizing them along the channel dimension, and \textit{(iii)} finally, applying a clustering method, such as the K-means algorithm, on fused features. Following these steps, we aim to retain each feature set's rich, descriptive qualities. Moreover, each resolution in the U-Net step maintains a different granularity of the regions regarding semantics and sizes.

\Cref{fig:seg_res} demonstrates the resulting segmentation maps from different resolutions and our proposed fused features. Each resolution captures different semantic components of the image (\eg{,} field, racket, hat and dress). 
Although \emph{Resolution 64} can distinguish objects (\eg{,} skin and outfit), it does not provide consistent segment areas, \eg{,} two distinct clusters for lines in the field. 
On the other hand, lower resolutions, \emph{Resolution 16 and 32}, capture coarse segments like lines in the field and the racket. Fusing those features from different resolutions yields more robust features, enhancing the segmentation; see \cref{fig:seg_res} - \emph{Ours}. %
For the extraction of intermediate features, we use time steps between 30 and 50 out of 100 steps, as recommended by LD-ZNet~\cite{PNVR2023Ldznet}.

\input{tex_figures/method_roi}

\noindent\textbf{Localization}
Upon identifying the segments within the input image, the proposed method identifies the RoI for the edit using cross-attention maps conditioned on the input image and instruction (\(z_t\), \(c_I\), and \(c_T\)) as indicated in \cref{eq:cfg2}. These maps have dimensions of $H_b \times W_b \times D$, where $H_b$ and $W_b$ represent the height and width of the features of block $b^{th}$ (up and down blocks), respectively, and $D$ denotes the number of text tokens.
Following the same strategy for segmentation, the cross-attention maps are resized to a common resolution, which is $64 \times 64$, combined among the spatial dimensions, namely $H$ and $W$, and normalized among the token dimension, $D$. Then, our method discards unrelated tokens, \eg{,} \textit{padding}, \textit{stop words} defined by NLTK~\cite{nltk}, and \textit{\textless start of text\textgreater} (SoT) defined by CLIP~\cite{Radford2021CLIP}. \textit{SoT} is an unrelated token, as recommended by~\cite{chefer2023attend}, while \textit{\textless end of text\textgreater} (EoT) is a related token since it is used as the feature representation of the text~\cite{Radford2021CLIP}. We verify experimentally that filtering these tokens allows \model{} to focus on related tokens, accurately identifying the RoI in accordance with the edit instruction.
Then, we get the mean attention score among the tokens to generate a final attention map; see \cref{fig:seg_res} - \emph{Attention}, that should correlate with the area of the image where the edit will be applied. Subsequently, we identify the top N points with the highest attention score, \eg{,} $N=100$ points - approximately $2.5\%$ of $64 \times 64$ attention maps, ablated in \cref{tab:ablation}. Eventually, all segments that overlap at least one of those points are combined to obtain the RoI; see \cref{fig:seg_res} - \emph{Ours}, \emph{Attention}, and \emph{RoI}. 

\subsection{Edit Application} \label{sec:edit_application}
This component proposes a novel \textit{attention regularization} which manipulates attention scores corresponding to the RoI while ensuring the rest remains the same, thus preventing any unintended alterations outside the RoI, \ie{,} localized image editing. Specifically, this procedure uses the terms with \( z_t \), \( c_I \), and \( c_T \) using the notation of \cref{eq:cfg2}. 

\input{tex_figures/method_editing}

\input{tex_figures/comparison}

\noindent\textbf{Attention Regularization} Previous methods~\cite{couairon2023diffedit, mirzaei2023watch, Avrahami2022BlendedDiffusion} use the noise space instead of attention scores. In contrast, our method introduces targeted attention regularization for selectively reducing the influence of unrelated tokens within the RoI during the editing process. This approach regularizes attention scores for tokens that are unrelated to the editing task, such as \textit{SoT}, \textit{padding}, and \textit{stop words}, defined in~\cref{sec:edit_localization} (denoted as \( S \)). By adjusting the attention scores (\( QK^T \)) within the RoI, we aim to minimize the impact of these unrelated tokens during the softmax normalization process. As a result, the softmax function is more likely to assign higher attention probabilities within the RoI to tokens that align with the editing instructions, \ie{,} related tokens. This targeted approach ensures that edits are precisely focused on the desired areas, enhancing the accuracy and effectiveness of the edits while preserving the rest. Given the binary mask for RoI $M$, we modify the result of the dot product $QK^T$ of cross-attention layers for unrelated tokens to a regularization version $R(QK^T, M)$ as follows:

\vspace{-1em}

\begin{equation}
\begin{split}
    R(QK^T, M) = \begin{cases} 
    QK^T_{ijt} - \infty, & \text{if } M_{ij} = 1 \text{ and } t \in S\\ %
    QK^T_{ijt}, & \text{otherwise}.
    \end{cases}
    \label{eq:attention_penalization}
\end{split}
\end{equation}
\vspace{-0.8em}

Intuitively, we prevent unrelated tokens from attending to the RoI, while related tokens will be more likely to be selected in the RoI, leading to more accurate, localized, and focused edits, as shown in \cref{fig:attn_reg}. This method achieves an optimal balance between targeted editing within the intended areas and preserving the surrounding context, thus enhancing the overall effectiveness of the instruction.

By employing our regularization technique within the RoI significantly enhances IP2P. It elevates the degree of disentanglement and improves the localization of edits by tapping into the already-learned features of the model. This targeted approach circumvents the need for \textit{re-training or fine-tuning} , preserving computational resources and time. It harnesses the inherent strength of the pre-trained IP2P features, deploying them in a focused and effective manner. This precision ensures that edits are contained within the intended areas, underpinning the model's improved capability to execute complex instructions in a localized and controlled way without the necessity for additional rounds of training, fine-tuning or additional user inputs.

%% file: tex_figures/method_roi.tex
\begin{figure}[!ht]
    \footnotesize
    \centering

    \vspace{-0.6em}
    
    \begin{minipage}{.24\linewidth}
        \centering
        \textbf{Input}
    \end{minipage}
    \begin{minipage}{.24\linewidth}
        \centering
        \textbf{Resolution 16}
    \end{minipage}
    \begin{minipage}{.24\linewidth}
        \centering
        \textbf{Resolution 32}
    \end{minipage}
    \begin{minipage}{.24\linewidth}
        \centering
        \textbf{Resolution 64}
    \end{minipage}

    \vspace{0.3em}

    \begin{minipage}{.24\linewidth}
        \centering
        \includegraphics[width=\textwidth]{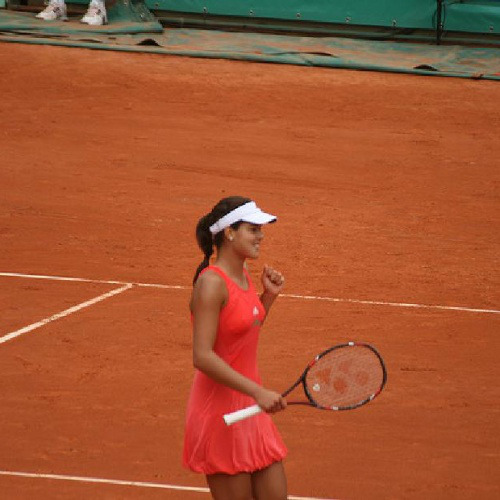}
    \end{minipage}
    \begin{minipage}{.24\linewidth}
        \centering
        \includegraphics[width=\textwidth]{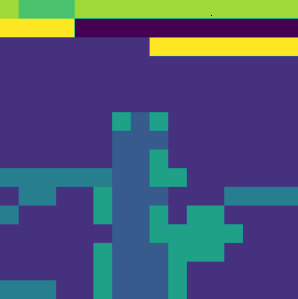}
    \end{minipage}
    \begin{minipage}{.24\linewidth}
        \centering
        \includegraphics[width=\textwidth]{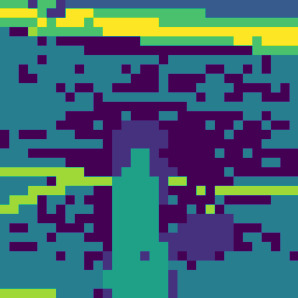}
    \end{minipage}
    \begin{minipage}{.24\linewidth}
        \centering
        \includegraphics[width=\textwidth]{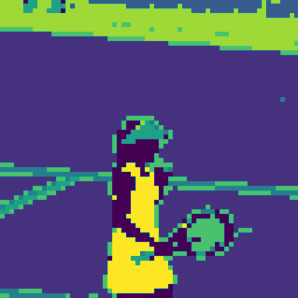}
    \end{minipage}

    \vspace{0.3em}

    \begin{minipage}{.24\linewidth}
        \centering
        \phantom{\textbf{Input}}
    \end{minipage}
    \begin{minipage}{.24\linewidth}
        \centering
        \textbf{Ours}
    \end{minipage}
    \begin{minipage}{.24\linewidth}
        \centering
        \textbf{Attention}
    \end{minipage}
    \begin{minipage}{.24\linewidth}
        \centering
        \textbf{RoI}
    \end{minipage}

    \vspace{0.3em}

    \begin{minipage}{.24\linewidth}
        \centering
        \textbf{Instruction:} \textit{Make her outfit black} \\ \textbf{\# of clusters:} 8
    \end{minipage}
    \begin{minipage}{.24\linewidth}
        \centering
        \includegraphics[width=\textwidth]{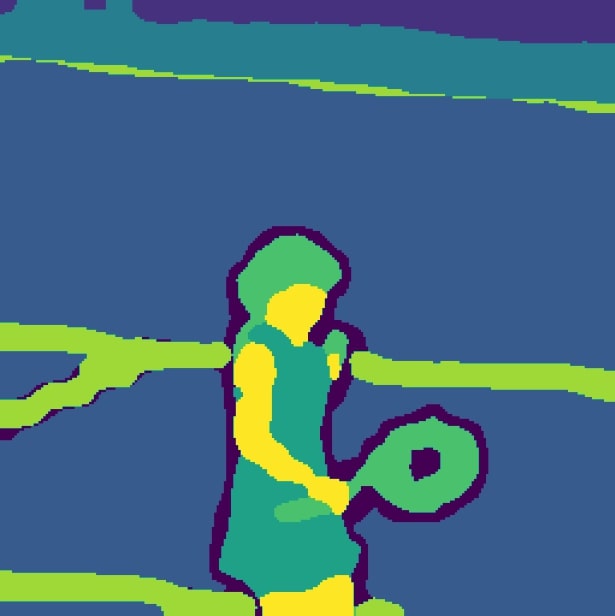}
    \end{minipage}
    \begin{minipage}{.24\linewidth}
        \centering
        \includegraphics[width=\textwidth]{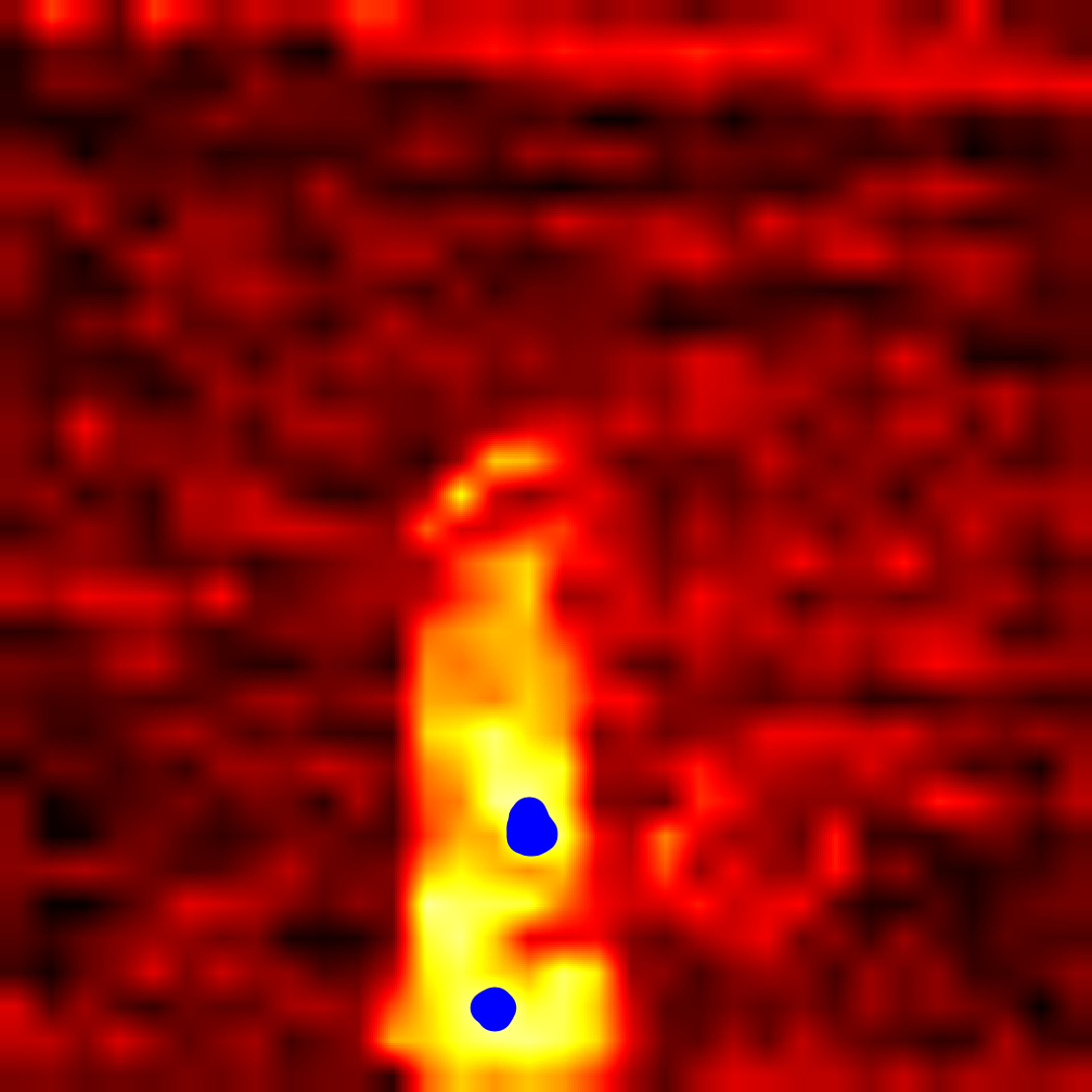}
    \end{minipage}
    \begin{minipage}{.24\linewidth}
        \centering
        \includegraphics[width=\textwidth]{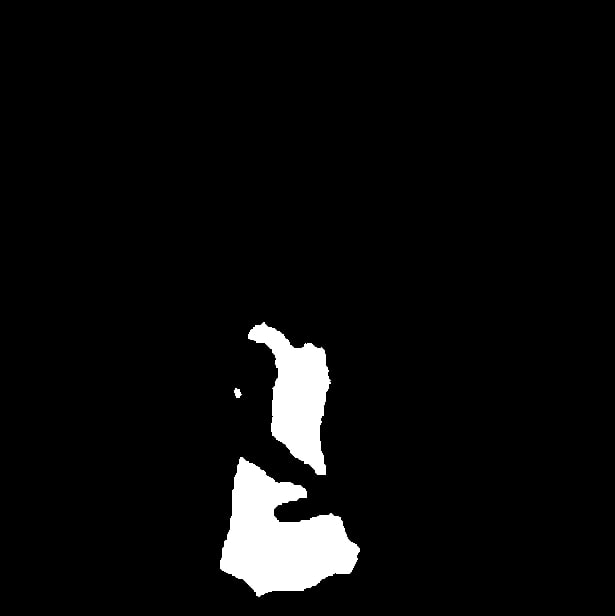}
    \end{minipage}
    \vspace{-.4em}

    \caption{\textbf{Segmentation and RoI finding.} \emph{Resolution X}s demonstrates segmentation maps from different resolutions, while \emph{Ours} shows the segmentation map from our multi-resolution fusion method explained above. For the cross-attention map, the color yellow indicates high probability, and blue dots mark the \textit{N} %
    points with the highest probability. The last image shows the extracted RoI using blue dots and \emph{Ours}. 
    \vspace{-1.2em}}
    \label{fig:seg_res}
\end{figure}

%% file: tex_figures/method_editing.tex
\begin{figure}[ht!]
    \centering
    \footnotesize
    \vspace{-0.6em}

    \begin{minipage}{\linewidth}
        \centering
        \textbf{Token-based cross-attention probabilities}
    \end{minipage}

    \vspace{0.5em}

    \begin{minipage}{.03\linewidth}
        \centering
        \rotatebox{90}{\textbf{Before}}
    \end{minipage}
    \begin{minipage}{.93\linewidth}
        \centering
        \includegraphics[width=\textwidth]{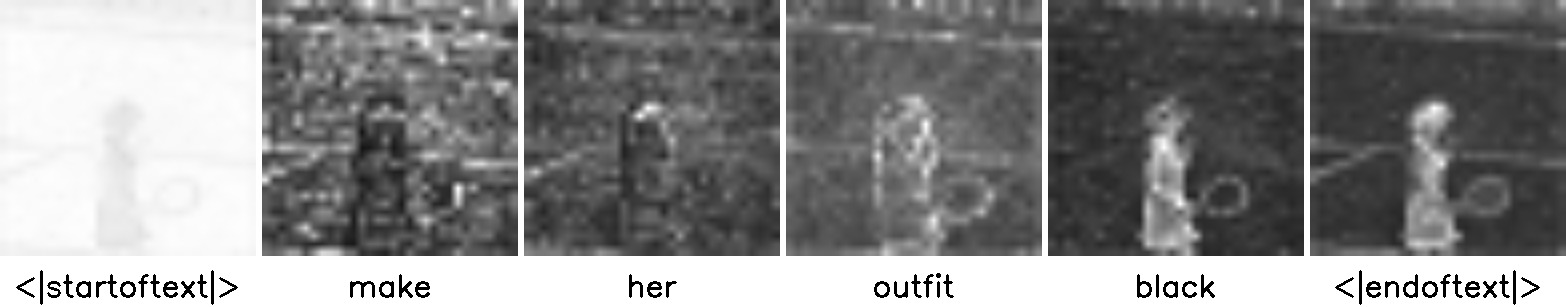}
    \end{minipage}

    \vspace{0.5em}

    \begin{minipage}{.03\linewidth}
        \centering
        \rotatebox{90}{\textbf{After}}
    \end{minipage}
    \begin{minipage}{.93\linewidth}
        \centering
        \includegraphics[width=\textwidth]{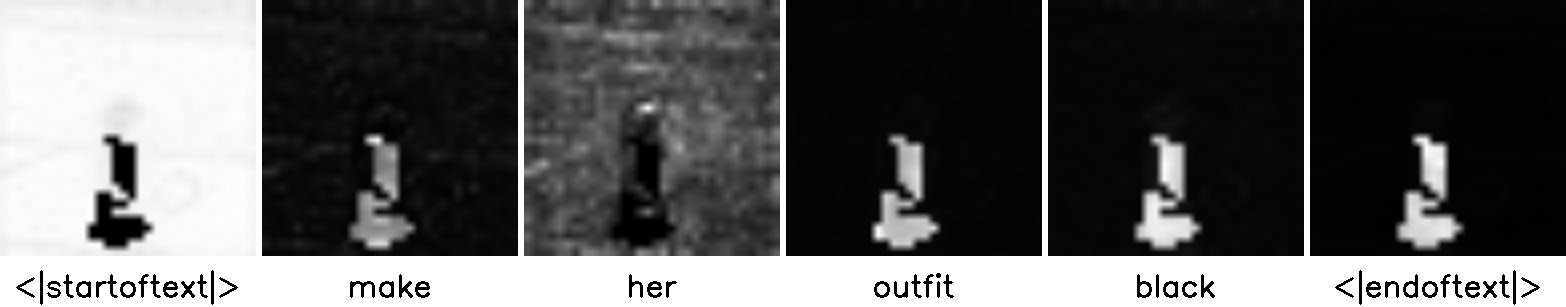}
    \end{minipage}
    \vspace{-.4em}

    \caption{\textbf{Attention Regularization.} Our method selectively regularizes unrelated tokens (\emph{SoT} and stop words: \textit{her}) within the RoI, ensuring precise, context-aware edits without the need for extra model training or user inputs. \textit{After} attention regularization, the probabilities for the related tokens are attending the RoI, as illustrated in the second row.}
    \vspace{-1.2em}

    \label{fig:attn_reg}
\end{figure}

%% file: tex_figures/comparison.tex
\begin{figure*}[!ht]
    \centering

    \scriptsize

    \tikz[remember picture,overlay] \node [anchor=base] (linebase) {};

    \begin{minipage}{.12\textwidth}
        \centering
        \textbf{Input Image}
    \end{minipage}
    \hfill
    \begin{minipage}{.12\textwidth}
        \centering
        \textbf{IP2P~\cite{Brooks2022InstructPix2Pix}}
    \end{minipage}
    \hfill
    \begin{minipage}{.12\textwidth}
        \centering
        \textbf{RoI}
    \end{minipage}
    \hfill
    \begin{minipage}{.12\textwidth}
        \centering
        \textbf{+ \model{}}
    \end{minipage}
    \hfill
    \begin{minipage}{.03\textwidth}
        \centering
    \end{minipage}
    \hfill    
    \begin{minipage}{.12\textwidth}
        \centering
        \textbf{Input Image}
    \end{minipage}
    \hfill
    \begin{minipage}{.12\textwidth}
        \centering
        \textbf{IP2P~\cite{Brooks2022InstructPix2Pix}}
    \end{minipage}
    \hfill
    \begin{minipage}{.12\textwidth}
        \centering
        \textbf{RoI}
    \end{minipage}
    \hfill
    \begin{minipage}{.12\textwidth}
        \centering
        \textbf{+ \model{}}
    \end{minipage}

    \vspace{0.3em}

    \begin{minipage}{.12\textwidth}
        \centering
        \includegraphics[width=\textwidth]{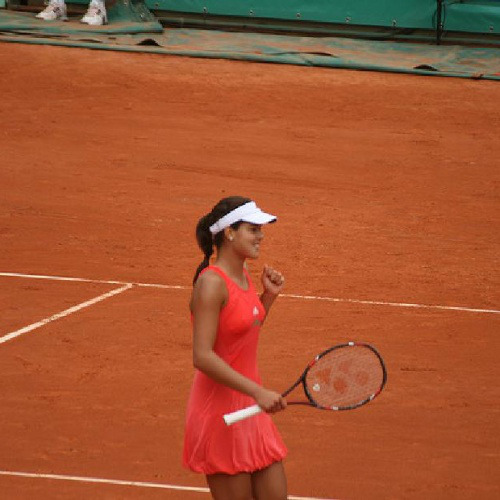}
    \end{minipage}
    \hfill
    \begin{minipage}{.12\textwidth}
        \centering
        \includegraphics[width=\textwidth]{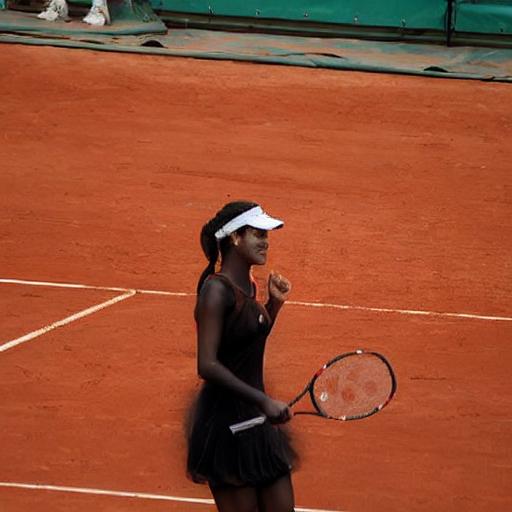}
    \end{minipage}
    \hfill
    \begin{minipage}{.12\textwidth}
        \centering
        \includegraphics[width=\textwidth]{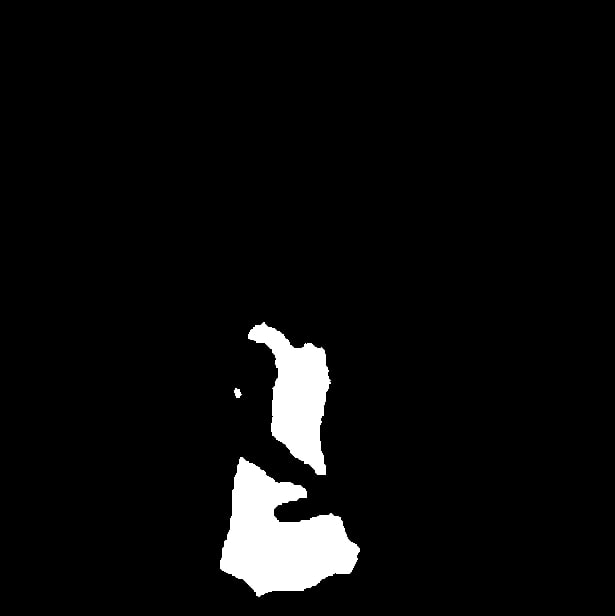}
    \end{minipage}
    \hfill
    \begin{minipage}{.12\textwidth}
        \centering
        \includegraphics[width=\textwidth]{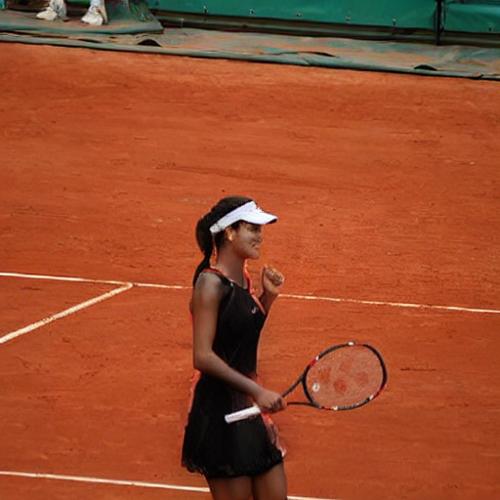}
    \end{minipage}
    \hfill
    \begin{minipage}{.03\textwidth}
        \centering
    \end{minipage}
    \hfill
    \begin{minipage}{.12\textwidth}
        \centering
        \includegraphics[width=\textwidth]{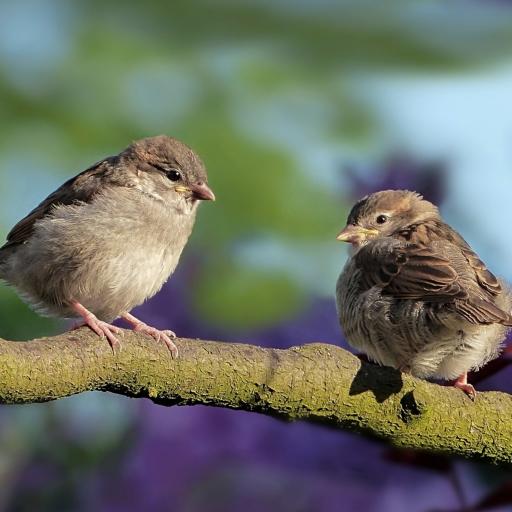}
    \end{minipage}
    \hfill
    \begin{minipage}{.12\textwidth}
        \centering
        \includegraphics[width=\textwidth]{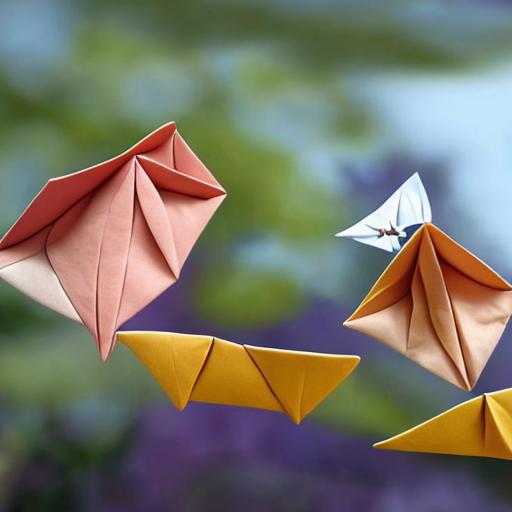}
    \end{minipage}
    \hfill
    \begin{minipage}{.12\textwidth}
        \centering
        \includegraphics[width=\textwidth]{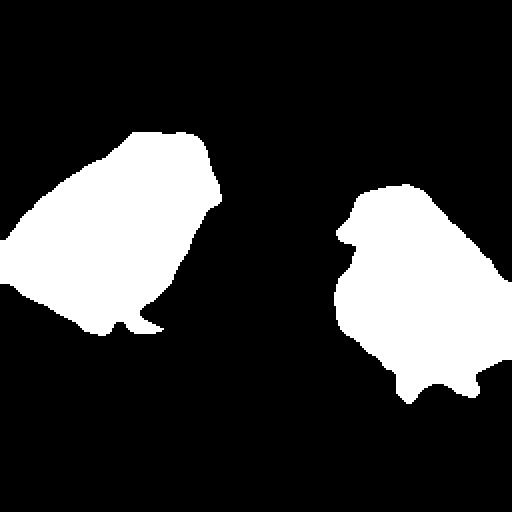}
    \end{minipage}
    \hfill
    \begin{minipage}{.12\textwidth}
        \centering
        \includegraphics[width=\textwidth]{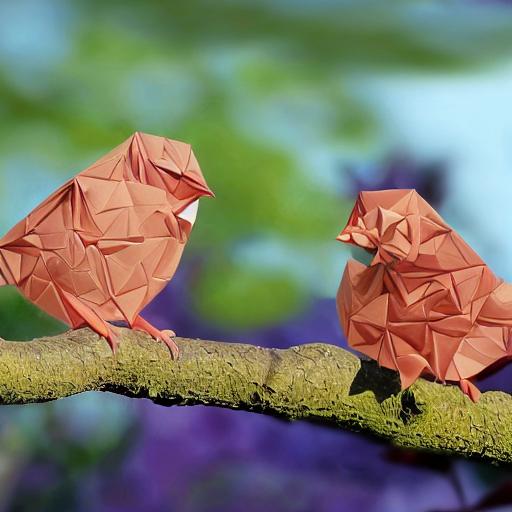}
    \end{minipage}

    \vspace{0.3em}

    \small{
        \begin{minipage}{.12\textwidth}
            \centering
            (a)
        \end{minipage}
        \hfill
        \begin{minipage}{.36\textwidth}
            \centering
            \textit{Make her outfit black.}
        \end{minipage}
        \hfill
        \begin{minipage}{.12\textwidth}
            \centering
            (b)
        \end{minipage}
        \hfill
        \begin{minipage}{.36\textwidth}
            \centering
            \textit{Turn the real birds into origami birds.}
        \end{minipage}
    }

    \vspace{0.3em}

    \begin{minipage}{.12\textwidth}
        \centering
        \includegraphics[width=\textwidth]{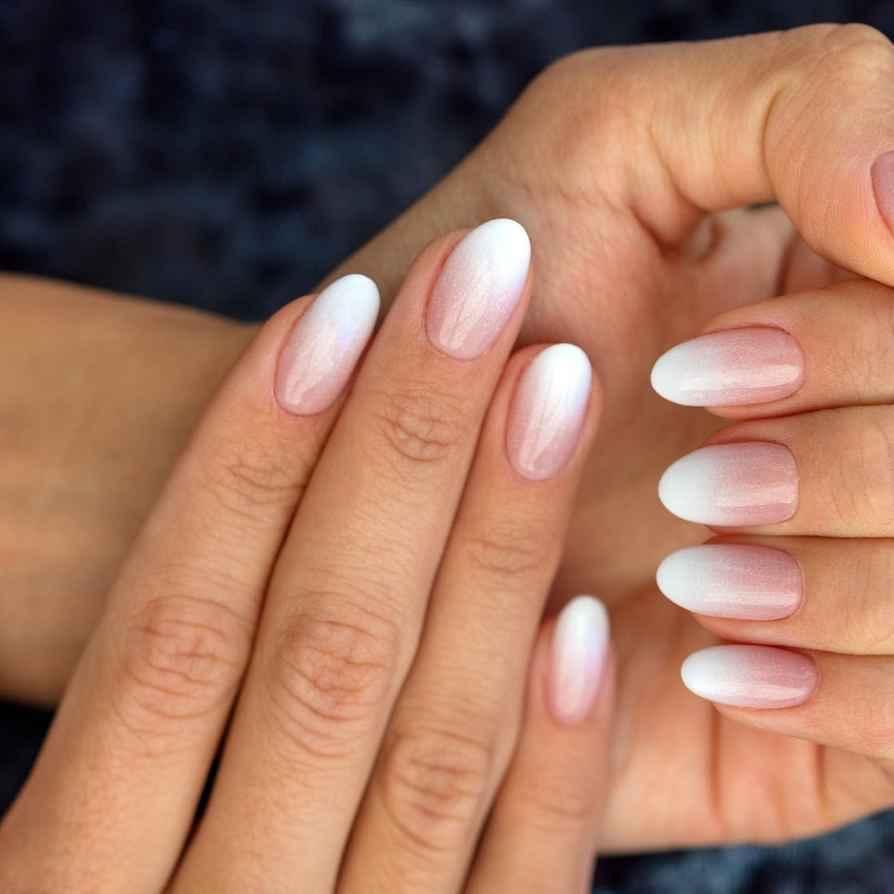}
    \end{minipage}
    \hfill
    \begin{minipage}{.12\textwidth}
        \centering
        \includegraphics[width=\textwidth]{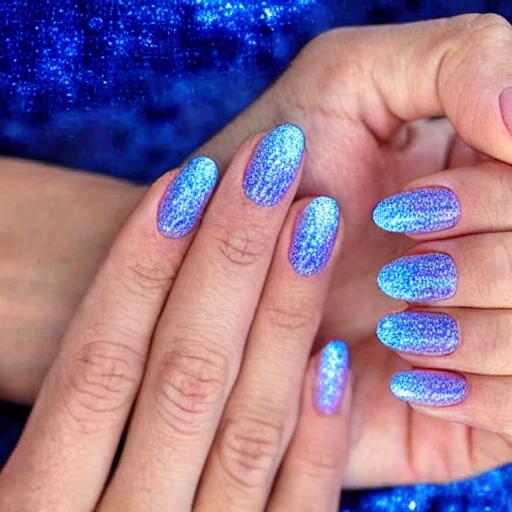}
    \end{minipage}
    \hfill
    \begin{minipage}{.12\textwidth}
        \centering
        \includegraphics[width=\textwidth]{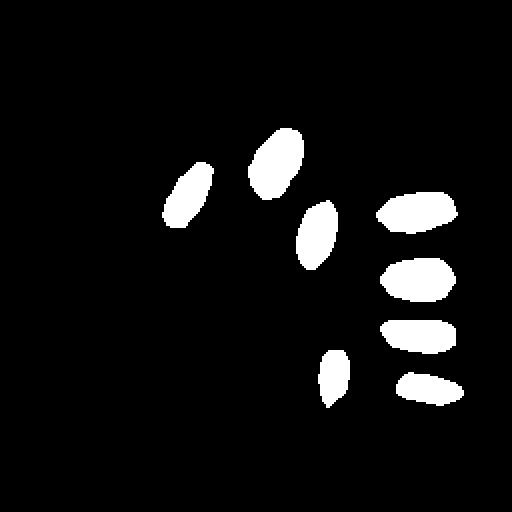}
    \end{minipage}
    \hfill
    \begin{minipage}{.12\textwidth}
        \centering
        \includegraphics[width=\textwidth]{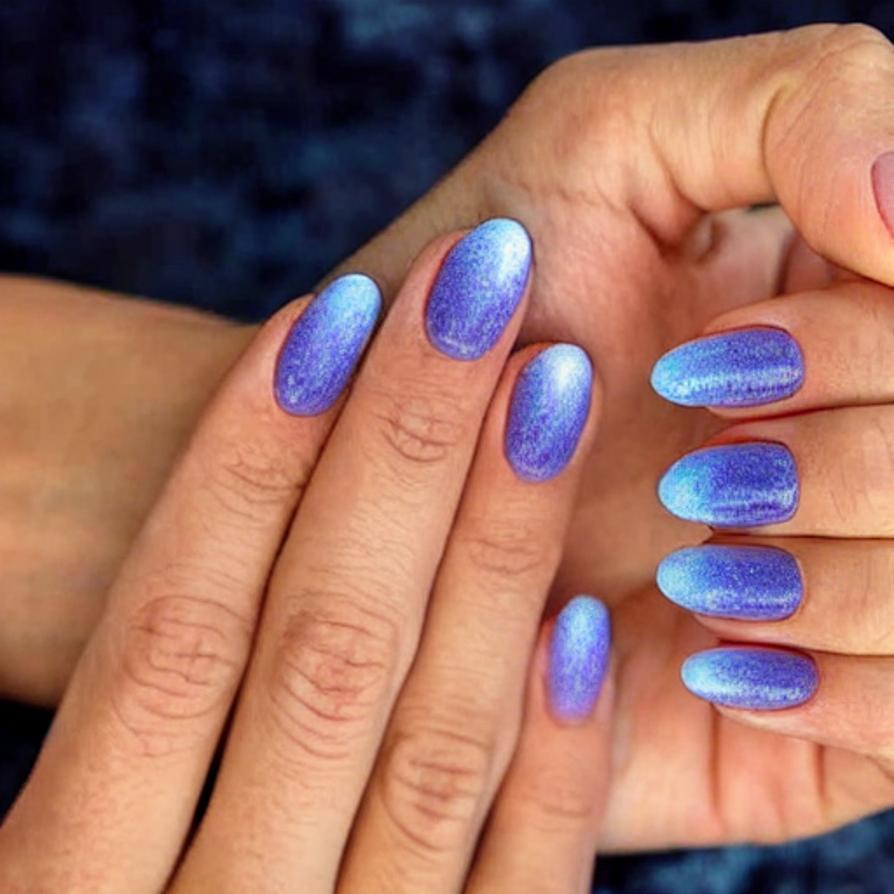}
    \end{minipage}
    \hfill
    \begin{minipage}{.03\textwidth}
        \centering
    \end{minipage}
    \hfill
    \begin{minipage}{.12\textwidth}
        \centering
        \includegraphics[width=\textwidth]{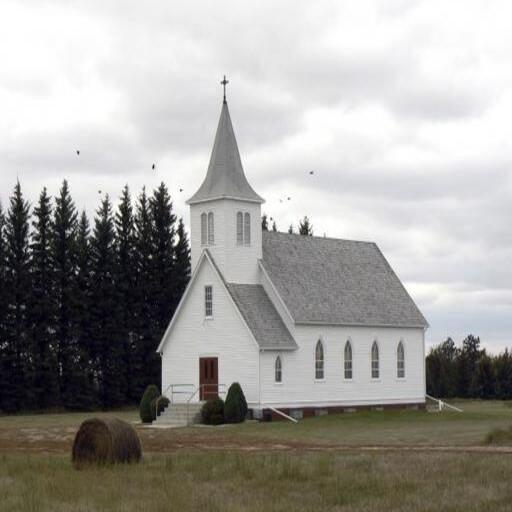}
    \end{minipage}
    \hfill
    \begin{minipage}{.12\textwidth}
        \centering
        \includegraphics[width=\textwidth]{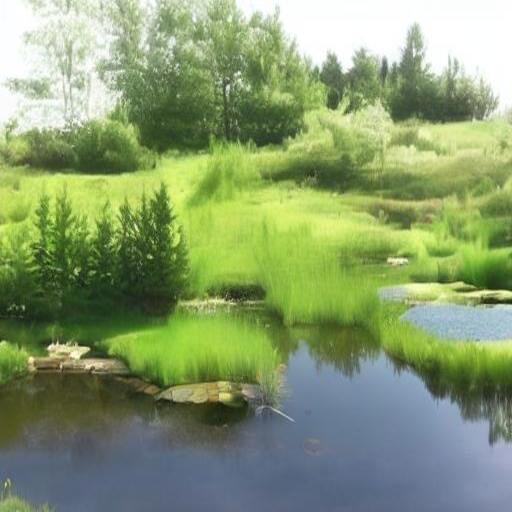}
    \end{minipage}
    \hfill
    \begin{minipage}{.12\textwidth}
        \centering
        \includegraphics[width=\textwidth]{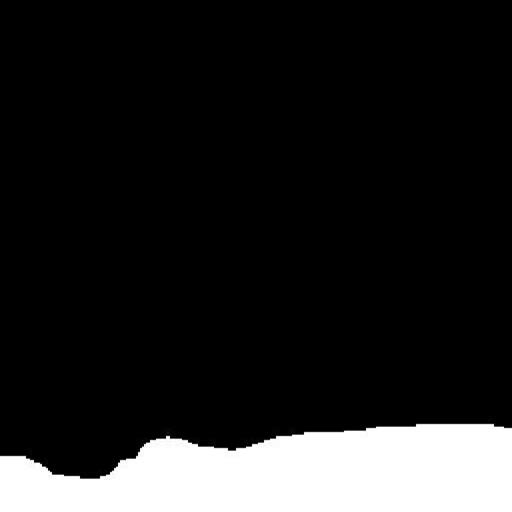}
    \end{minipage}
    \hfill
    \begin{minipage}{.12\textwidth}
        \centering
        \includegraphics[width=\textwidth]{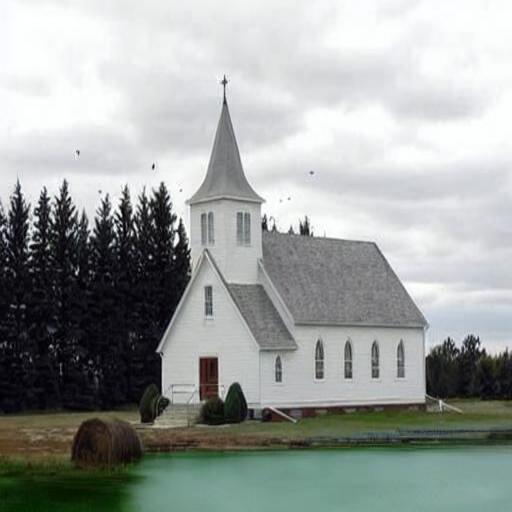}
    \end{minipage}

    \vspace{0.3em}

    \small{
        \begin{minipage}{.12\textwidth}
            \centering
            (c)
        \end{minipage}
        \hfill
        \begin{minipage}{.36\textwidth}
            \centering
            \textit{Put blue glitter on fingernails.}
        \end{minipage}
        \hfill
        \begin{minipage}{.12\textwidth}
            \centering
            (d)
        \end{minipage}
        \hfill
        \begin{minipage}{.36\textwidth}
            \centering
            \textit{Add a pond.}
        \end{minipage}
    }

    \vspace{-0.4em}

    \begin{tikzpicture}
        \draw[dashed, very thick] (0,0) -- (\textwidth,0);
    \end{tikzpicture}

    \vspace{0.1em}

    \scriptsize

    \begin{minipage}{.12\textwidth}
        \centering
        \textbf{Input Image}
    \end{minipage}
    \hfill
    \begin{minipage}{.12\textwidth}
        \centering
        \textbf{IP2P~\cite{Brooks2022InstructPix2Pix} w/MB~\cite{Zhang2023MagicBrush}}
    \end{minipage}
    \hfill
    \begin{minipage}{.12\textwidth}
        \centering
        \textbf{RoI}
    \end{minipage}
    \hfill
    \begin{minipage}{.12\textwidth}
        \centering
        \textbf{+ \model{}}
    \end{minipage}
    \hfill
    \begin{minipage}{.03\textwidth}
        \centering
    \end{minipage}
    \hfill    
    \begin{minipage}{.12\textwidth}
        \centering
        \textbf{Input Image}
    \end{minipage}
    \hfill
    \begin{minipage}{.12\textwidth}
        \centering
        \textbf{IP2P~\cite{Brooks2022InstructPix2Pix} w/MB~\cite{Zhang2023MagicBrush}}
    \end{minipage}
    \hfill
    \begin{minipage}{.12\textwidth}
        \centering
        \textbf{RoI}
    \end{minipage}
    \hfill
    \begin{minipage}{.12\textwidth}
        \centering
        \textbf{+ \model{}}
    \end{minipage}

    \vspace{0.3em}

    \begin{minipage}{.12\textwidth}
        \centering
        \includegraphics[width=\textwidth]{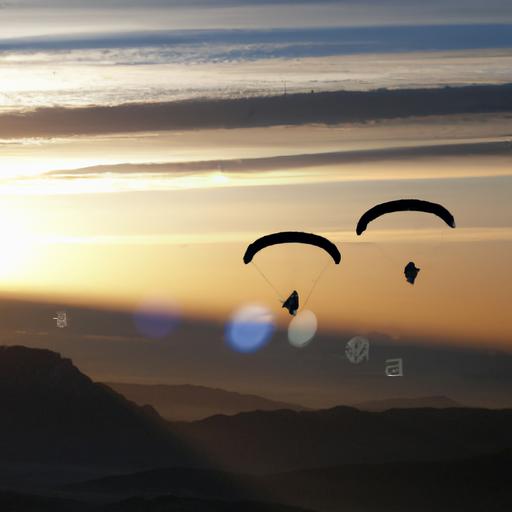}
    \end{minipage}
    \hfill
    \begin{minipage}{.12\textwidth}
        \centering
        \includegraphics[width=\textwidth]{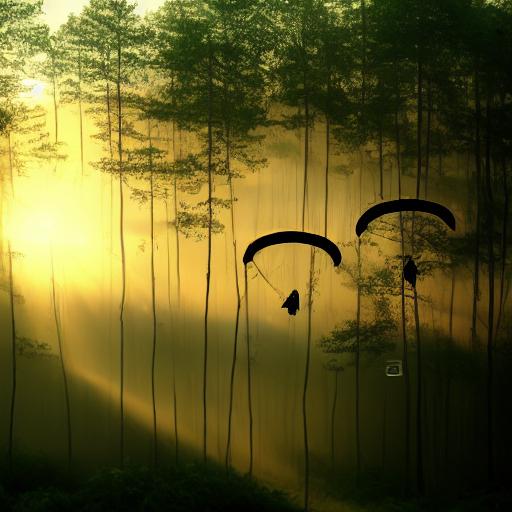}
    \end{minipage}
    \hfill
    \begin{minipage}{.12\textwidth}
        \centering
        \includegraphics[width=\textwidth]{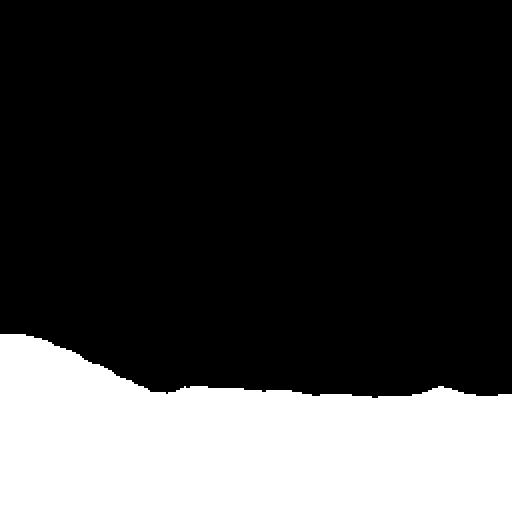}
    \end{minipage}
    \hfill
    \begin{minipage}{.12\textwidth}
        \centering
        \includegraphics[width=\textwidth]{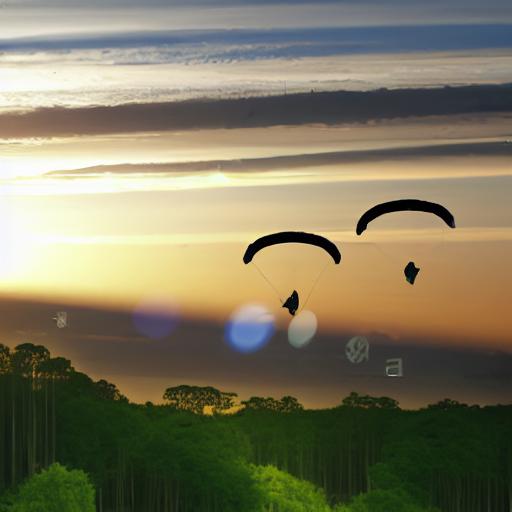}
    \end{minipage}
    \hfill
    \begin{minipage}{.03\textwidth}
        \centering
    \end{minipage}
    \hfill
    \begin{minipage}{.12\textwidth}
        \centering
        \includegraphics[width=\textwidth]{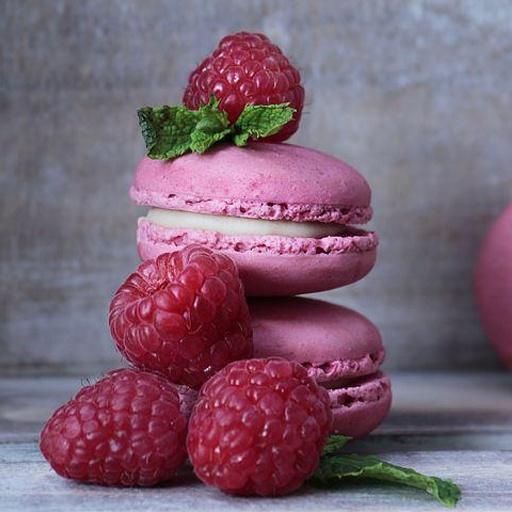}
    \end{minipage}
    \hfill
    \begin{minipage}{.12\textwidth}
        \centering
        \includegraphics[width=\textwidth]{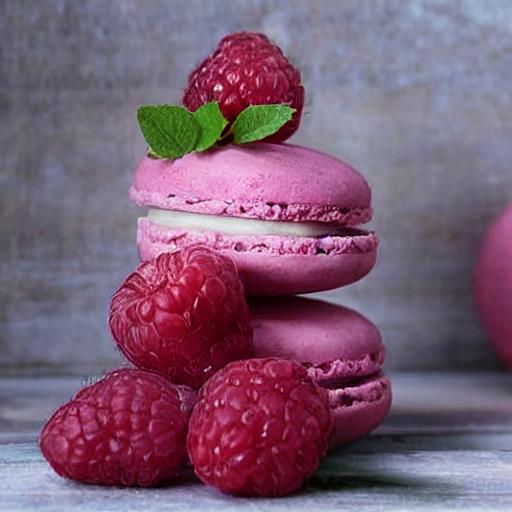}
    \end{minipage}
    \hfill
    \begin{minipage}{.12\textwidth}
        \centering
        \includegraphics[width=\textwidth]{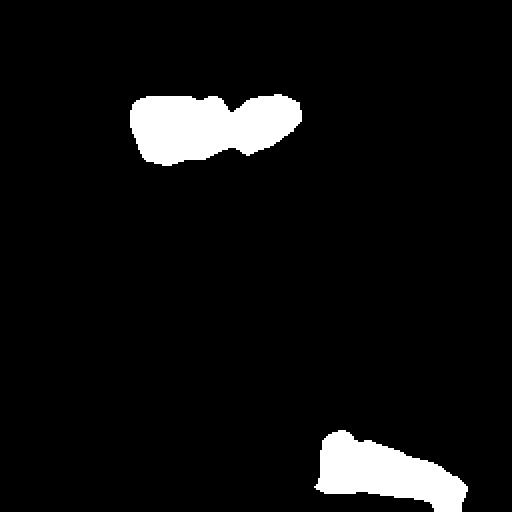}
    \end{minipage}
    \hfill
    \begin{minipage}{.12\textwidth}
        \centering
        \includegraphics[width=\textwidth]{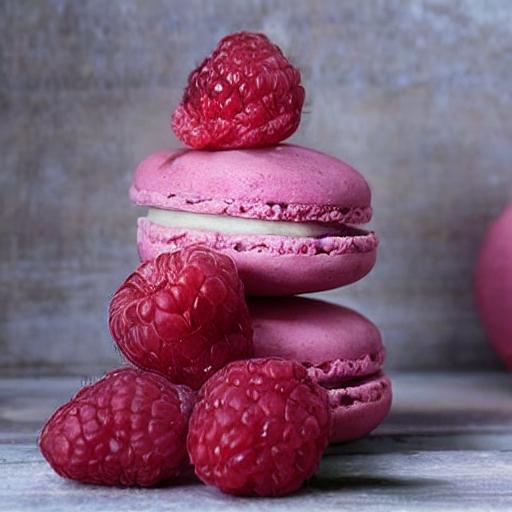}
    \end{minipage}

    \vspace{0.3em}

    \small{
        \begin{minipage}{.12\textwidth}
            \centering
            (e)
        \end{minipage}
        \hfill
        \begin{minipage}{.36\textwidth}
            \centering
            \textit{Replace the ground with a forest.}
        \end{minipage}
        \hfill
        \begin{minipage}{.12\textwidth}
            \centering
            (f)
        \end{minipage}
        \hfill
        \begin{minipage}{.36\textwidth}
            \centering
            \textit{Remove mint leaves.}
        \end{minipage}
    }

    \begin{tikzpicture}[remember picture, overlay]
        \draw[dashed, very thick] (linebase) -- ++(0,-8.7); %
    \end{tikzpicture}

    \vspace{-1.2em}
    
    \caption{\textbf{Qualitative Examples.} We test our method on different tasks: (a) editing a large segment, (b) altering texture, (c) editing multiple segments, (d) adding, (e) replacing, and (f) removing objects. 
    The integration of {\textbf{{\model{}}}} enhances the performance of baselines, enabling localized edits while preserving the remaining image areas.} 
    \label{fig:comparison}
    \vspace{-1.8em}

\end{figure*}
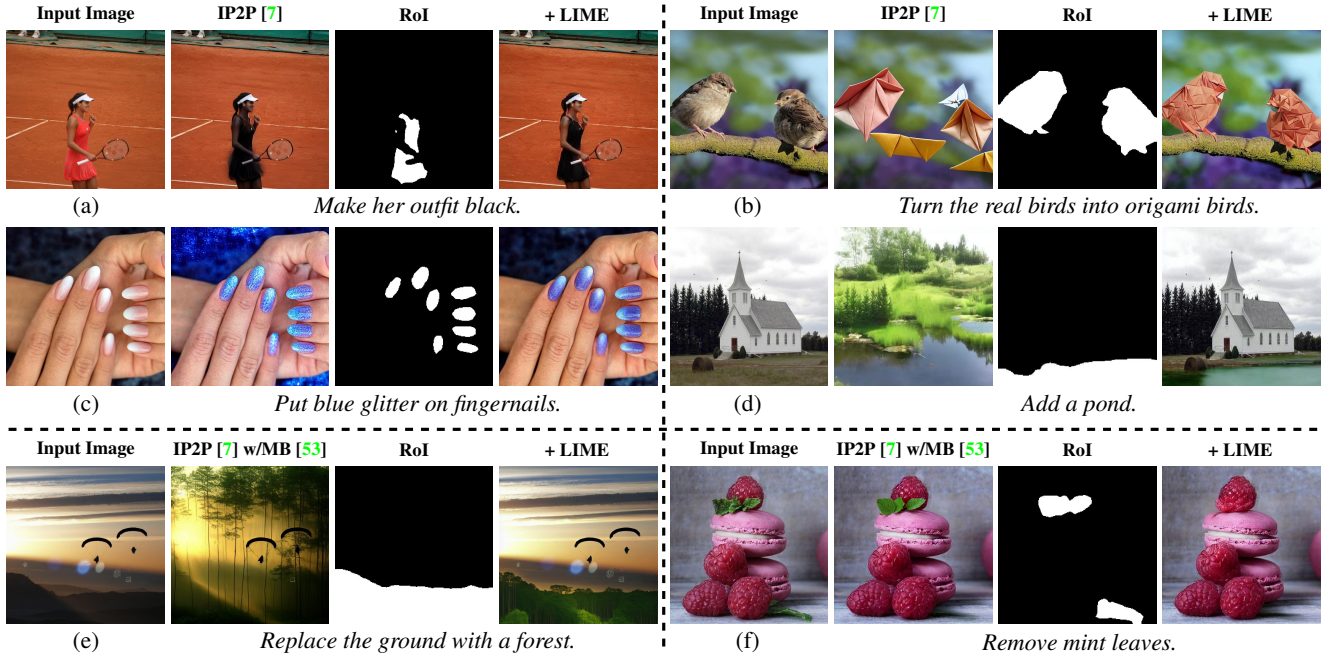

%% file: sections/5_experiments.tex
\section{Experiments}
To assess the effectiveness of our model, we integrate it into the widely recognized IP2P~\cite{Brooks2022InstructPix2Pix}, referenced as IP2P + \model{} throughout this section. The original IP2P model serves as our standard benchmark for all analyses.
Our evaluation involves a comprehensive quantitative comparison of our approach against other state-of-the-art methods, utilizing established datasets and metrics, complemented by a user study. 
While \model{} can be integrated with other models~\cite{Hertz2022Prompt2prompt,Zhang2023HIVE}, we limit the experimental analysis in the main paper to only IP2P for clarity. Additional results with other models will be provided in \textit{Supplementary Material}. %

\input{tex_figures/qualitative_examples}

\subsection{Evaluation Datasets and Metrics}
\noindent\textbf{MagicBrush~\cite{Zhang2023MagicBrush}} Its test split offers an evaluation pipeline with 535 sessions - refer to the source images used for iterative editing instructions - and 1053 turns - denote the individual editing steps within each session. It employs \textit{L1} and \textit{L2} norms to measure pixel accuracy, \textit{CLIP-I}, and \textit{DINO} embeddings for assessing image quality via cosine similarity, and \textit{CLIP-T} to ensure that the generated images align accurately with local textual descriptions.

\begin{figure*}
  \centering

  \begin{subfigure}{0.68\linewidth}
    \input{tables/quantitative_magicbrush}

  \end{subfigure}
  \hfill
  \begin{subfigure}{0.3\linewidth}
    \input{tex_figures/ip2p_quantitative}
  \end{subfigure}
\vspace{-0.6em}
  \caption{Evalutation on MagicBrush~\cite{Zhang2023MagicBrush} and IP2P Test Split~\cite{Brooks2022InstructPix2Pix}.}
  
  \vspace{-1.5em}
  \label{fig:short}
\end{figure*}

\noindent\textbf{InstructPix2Pix~\cite{Brooks2022InstructPix2Pix}}
We evaluate our method on IP2P test split with 5K image-instruction pairs on \textit{CLIP image similarity} for visual fidelity and \textit{CLIP text-image direction similarity} to measure adherence to the editing instructions.

\subsection{Implementation Details}
Our method adopts IP2P~\cite{Brooks2022InstructPix2Pix} as its base model and runs the model for 100 steps for each component in~\cref{sec:edit_localization,sec:edit_application}. Specifically, during \textit{Edit Localization}, intermediate representations are extracted between $30$ and $50$ out of $100$ steps, as suggested in LD-ZNet~\cite{PNVR2023Ldznet}. Moreover, those intermediate features are resized to $256 \times 256$. The number of clusters for segmenting is $8$ across all experiments, motivated by~\cref{sec:ablation}. We gather cross-attention maps from steps $1$ to $75$ and retain only related tokens. We extract $100$ points with the highest probabilities from the attention maps to identify RoI and determine overlapping segments. For \emph{Edit Localization}, the image scale $s_I$ and the text scale $s_T$ are set to $1.5$ and $7.5$, respectively. During \textit{Edit Application}, the attention regularization is employed between steps $1$ and $75$, targeting only unrelated tokens, \eg{,} \textit{SoT}, \textit{padding}, and \textit{stop words}. Throughout the editing process, the image scale, $s_I$, and the text scale, $s_T$, parameters are set to $1.5$ and $3.5$, respectively.

\subsection{Qualitative Results}
\Cref{fig:comparison} presents qualitative examples for various editing tasks: editing large segments, altering textures, editing multiple small segments simultaneously, and adding, replacing, or removing objects. The first column shows the input images, with the edit instructions below each image. The second column illustrates the results generated by the base models without our proposed method. The third and fourth columns report the RoI identified by our method and the edit result by integrating \model{} to base models considering the RoI. As shown in \cref{fig:comparison}, our method effectively implements the edit instructions while preserving the overall scene.
In all presented results, our method surpasses current state-of-the-art models, including their fine-tuned versions on manually annotated datasets, \eg{,} IP2P w/MB~\cite{Zhang2023MagicBrush}. %

\Cref{fig:qual_examples} provides further analyses. The integration of \model{} successfully handles diverse tasks, including (a) modifying multiple objects, (b) altering the color of object, (c) adding, (d) removing objects, and altering the texture of an animal (e). As depicted in \cref{fig:qual_examples}, our approach significantly surpasses existing baselines. Notably, while they tend to alter entire scene rather than the RoI, \eg{,} shelf or bird, our method targets and modifies the RoI (a), and (e). Furthermore, baseline models often inadvertently affect areas beyond the intended RoI, as seen in cases (b) and (c). In contrast, our method demonstrates precision by confining its operations within the RoI. Particularly in scenario (d), while baseline models such as IP2P struggle to maintain the integrity of the original image or, as in the case of IP2P w/MB, fail to remove objects effectively, our method excels by accurately removing the specified objects, underscoring its superiority in targeted image editing tasks. Additional comparisons and the applicability of \model{} on other methods are provided in \textit{Supplementary Material}.

\subsection{Quantitative Results}

\noindent\textbf{Results on MagicBrush}
Our method outperforms all other methods on both the single- and multi-turn editing tasks on MagicBrush~(MB) benchmark~\cite{Zhang2023MagicBrush}, as seen in ~\cref{tab:magicbrush}. Compared to the base models, our approach provides significant improvements and best results in terms of \textit{L1}, \textit{L2}, \textit{CLIP-I}, and \textit{DINO}. For the \textit{CLIP-T} metric, which compares the edited image and caption to the ground truth, our method comes very close to the oracle scores --- metric between GT edited image vs. caption --- of $0.309$ for multi-turn and $0.307$ for single-turn. This indicates that our edits accurately reflect the ground truth modifications. VQGAN-CLIP~\cite{Crowson2022VQGAN-CLIP} achieves the highest in \textit{CLIP-T} by directly using CLIP~\cite{Radford2021CLIP} for fine-tuning during inference. However, this can excessively alter images, leading to poorer performance in other metrics. Overall, the performance across metrics shows that our approach generates high-quality and localized image edits based on instructions, outperforming other methods.

\vspace{2pt}

\noindent\textbf{Results on InstructPix2Pix}
We evaluate our method using the same procedure as IP2P, presenting results on a synthetic evaluation dataset~\cite{Brooks2022InstructPix2Pix} as shown in ~\cref{fig:IP2P}. Our approach notably improves the base model, IP2P, optimizing the trade-off between the input image and edit instruction. Additionally, while an increase in text scale, \( s_T \), enhances the \textit{CLIP Text-Image Direction Similarity}, it adversely impacts \textit{ CLIP Image Similarity}. For both metrics, the higher, the better. The \textbf{black arrow} shows the default configuration for our study, which is the basis for this paper's reporting of all metrics with \( s_T = 3.5 \) and the number of clusters is 8.

\vspace{2pt}

\input{tables/user_study}

\noindent\textbf{User Study}
Although quantitative metrics provide a reasonable means of comparing baselines, evaluating the precision and impact of edits is challenging. Therefore, we conduct a user study, asking 53 anonymous individuals on the crowd-sourcing platform Prolific~\cite{prolific} to answer questions about the localization (Q1) and effectiveness (Q2) of edits for a random selection of 54 pairs of images and edit instructions. Each question has three options: (i) baseline, (ii) baseline + \model{} and (iii) none of them (Neither). The findings from our survey indicate that our approach outperforms baselines~\cite{Brooks2022InstructPix2Pix, Zhang2023MagicBrush} in both metrics, see~\cref{tab:user_sduty}.

\subsection{Ablation Study} \label{sec:ablation}

We run ablation studies both on alternative designs for both on components (\cref{sec:edit_localization} and \cref{sec:edit_application}) of \model{}, and on the impact of parameters of \model{}.

\input{tables/ablation}

\vspace{2pt}

\noindent\textbf{Mask type}
We ablate localization design alternative to \cref{sec:edit_localization}. Specifically, we employ ground-truth masks provided by MagicBrush~\cite{Zhang2023MagicBrush} and the diffusion-based segmentation from LPM approach~\cite{patashnik2023localizing} integrated into the IP2P framework. The results show that our proposed method outperforms the competing approach on the MagicBrush dataset, see \cref{tab:ablation}, and performs comparably or better to ground-truth. More analysis is provided in \textit{Supplementary Material} and further comparisons to alternatively utilizing an off-the-shelf segmentor for localization. 

\vspace{2pt}

\noindent\textbf{Edit type} Instead of \textit{attention regularization} in ~\cref{sec:edit_application}, editing can also be performed in noise space~\cite{couairon2023diffedit, Avrahami2022BlendedDiffusion, mirzaei2023watch}. This corresponds to blending of the input image and a reference image derived from the edit text in noise space, based on the RoI. However, alignment between the reference and input images in the edited area is crucial for targeting the RoI effectively. As shown in \cref{tab:ablation}, our method enhances local editing by employing \textit{attention regularization}.

\vspace{2pt}

\noindent\textbf{Number of points} 
We test with different values to understand the impact of the number of points, \(N\), for RoI finding. Using only $25$ points worsens performance, as it cannot capture multiple distinct segments within RoI. However, having more points includes excessive noise, causing more segments to improperly merge and expanding the RoI area. $100$ points provide better RoI, see \cref{tab:ablation}.

\vspace{2pt}

\noindent\textbf{Number of clusters} 
To show the robustness of our method in terms of the number of clusters, we perform an ablation study. A low cluster count (\eg{,} 4) merges into larger segments, hindering precise edits, while a high count (\eg{,} 16 or 32) fragments the RoI excessively. Our findings, highlighted in~\cref{tab:ablation}, indicate that 8 clusters strike an optimal balance. %

%% file: tex_figures/qualitative_examples.tex
\begin{figure*}[!hpt]
    \centering

    \tikz[remember picture,overlay] \node [anchor=base] (linebase) {};

    \begin{minipage}{.18\textwidth}
        \centering
        \textbf{Input Image}
    \end{minipage}
    \begin{minipage}{.18\textwidth}
        \centering
        \textbf{IP2P}
    \end{minipage}
    \begin{minipage}{.18\textwidth}
        \centering
        \textbf{+ \model{}}
    \end{minipage}
    \begin{minipage}{.18\textwidth}
        \centering
        \textbf{IP2P w/MB}
    \end{minipage}
    \begin{minipage}{.18\textwidth}
        \centering
        \textbf{+ \model{}}
    \end{minipage}

    \vspace{0.6em}

    \begin{minipage}{.18\textwidth}
        \centering
        \includegraphics[width=\textwidth]{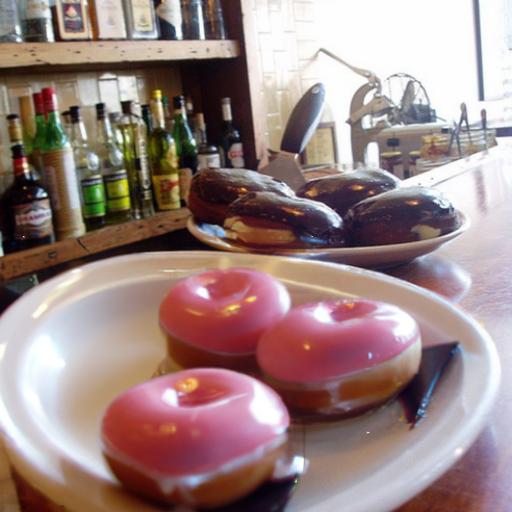}
    \end{minipage}
    \begin{minipage}{.18\textwidth}
        \centering
        \includegraphics[width=\textwidth]{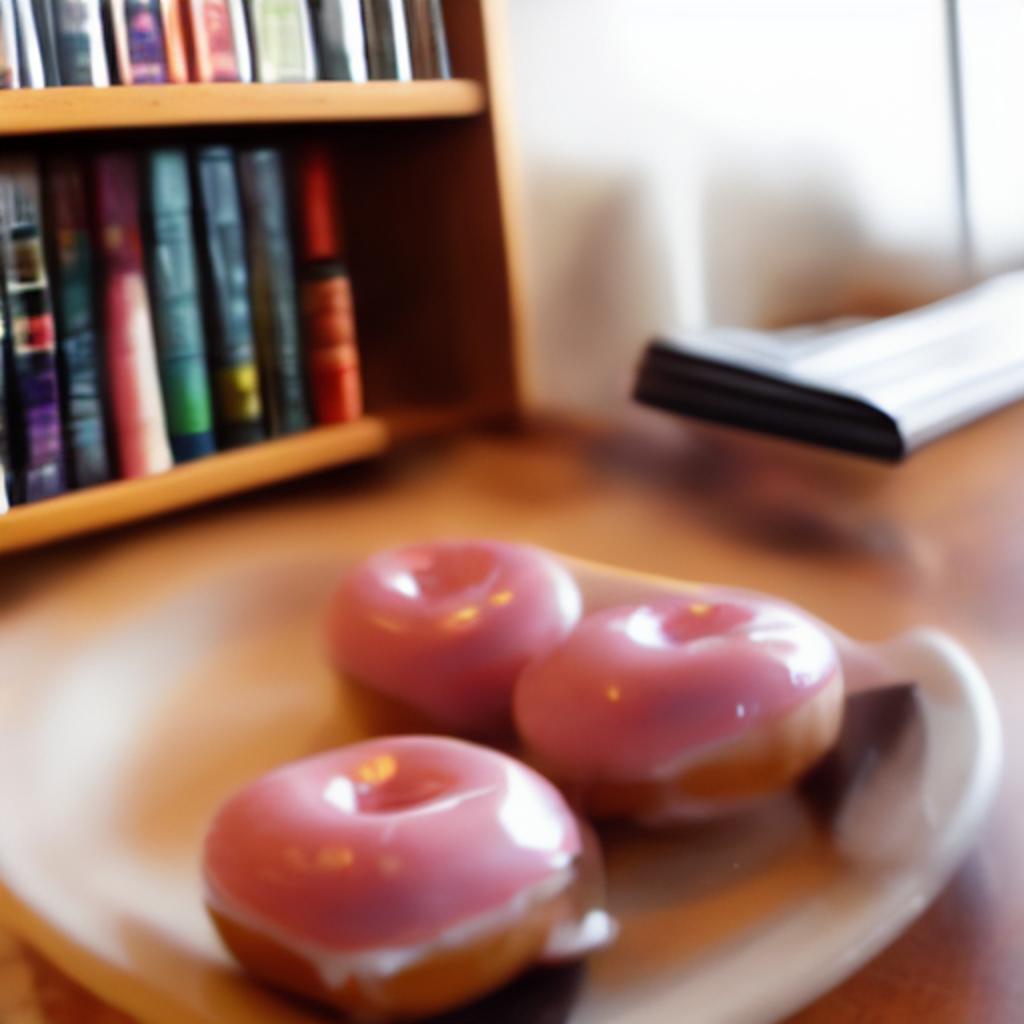}
    \end{minipage}
    \begin{minipage}{.18\textwidth}
        \centering
        \includegraphics[width=\textwidth]{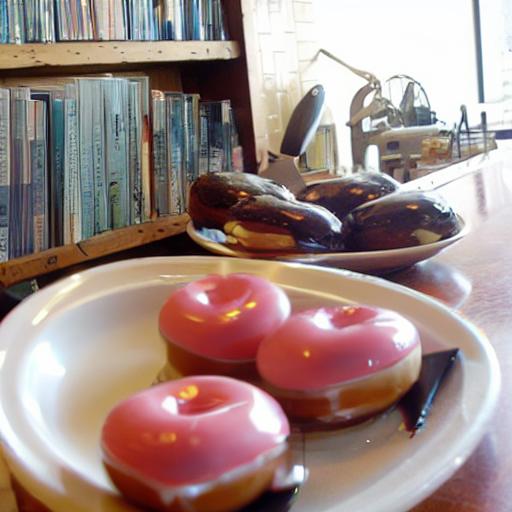}
    \end{minipage}
    \begin{minipage}{.18\textwidth}
        \centering
        \includegraphics[width=\textwidth]{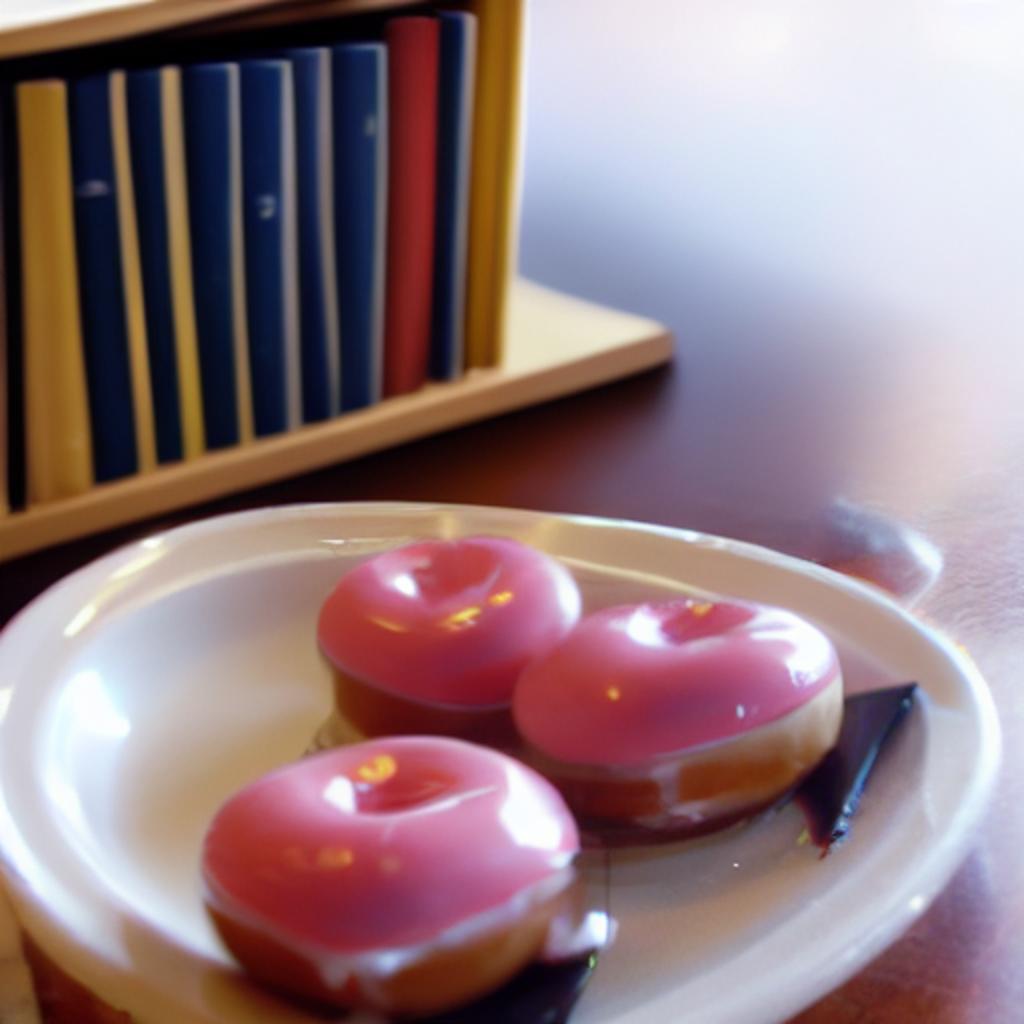}
    \end{minipage}
    \begin{minipage}{.18\textwidth}
        \centering
        \includegraphics[width=\textwidth]{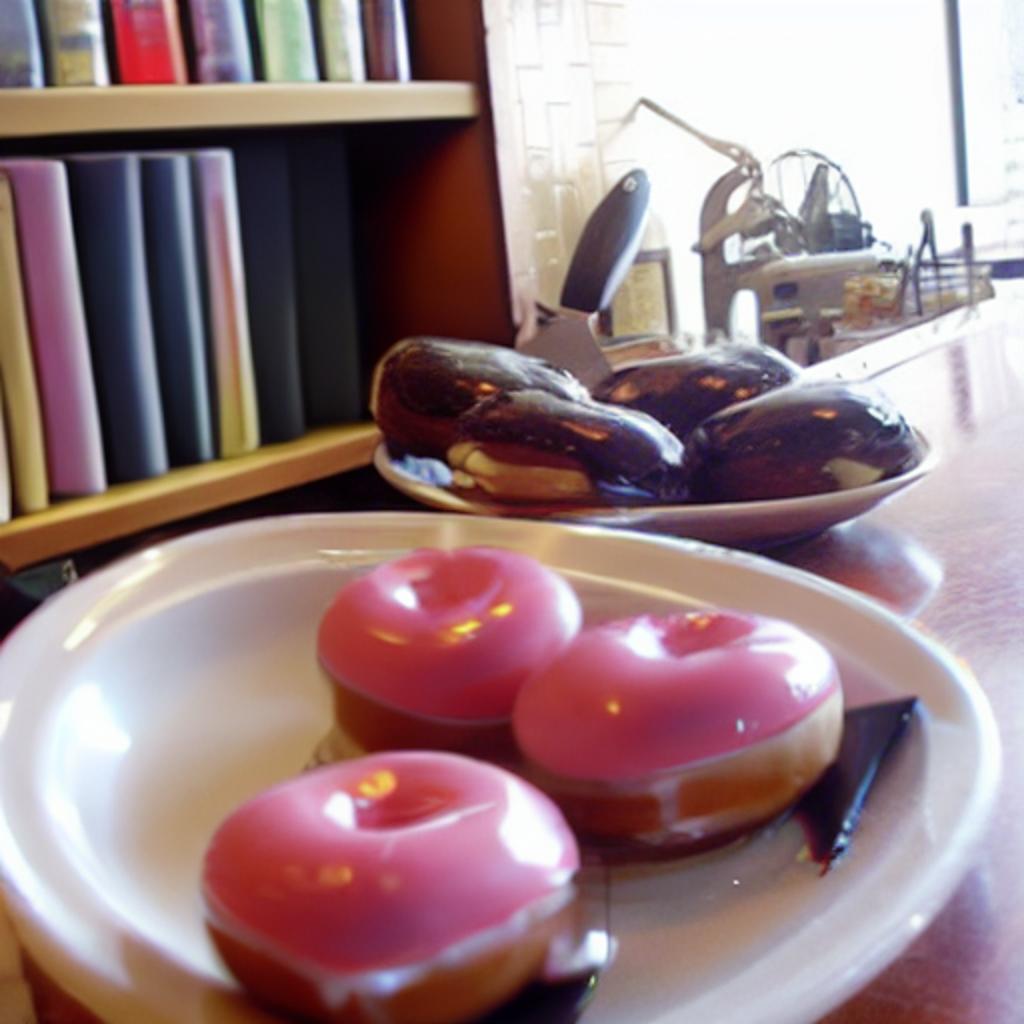}
    \end{minipage}

    \vspace{0.6em}

    \begin{minipage}{.18\textwidth}
        \centering
        (a)
    \end{minipage}
    \begin{minipage}{.72\textwidth}
        \centering
        \textit{Change the alcohol shelf into a bookshelf.}
    \end{minipage}
    
    \vspace{0.6em}

    \begin{minipage}{.18\textwidth}
        \centering
        \includegraphics[width=\textwidth]{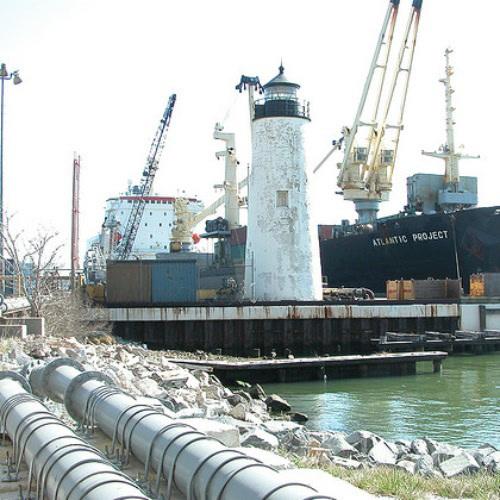}
    \end{minipage}
    \begin{minipage}{.18\textwidth}
        \centering
        \includegraphics[width=\textwidth]{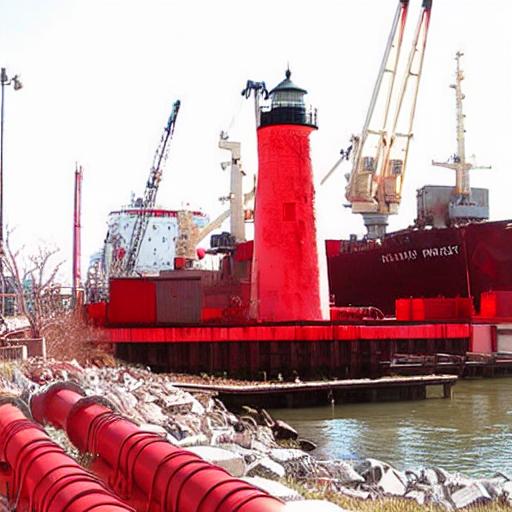}
    \end{minipage}
    \begin{minipage}{.18\textwidth}
        \centering
        \includegraphics[width=\textwidth]{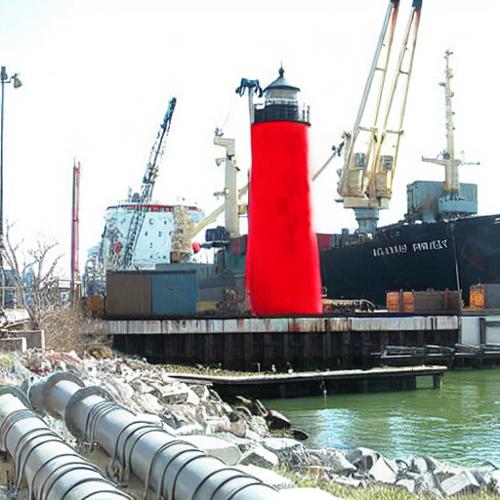}
    \end{minipage}
    \begin{minipage}{.18\textwidth}
        \centering
        \includegraphics[width=\textwidth]{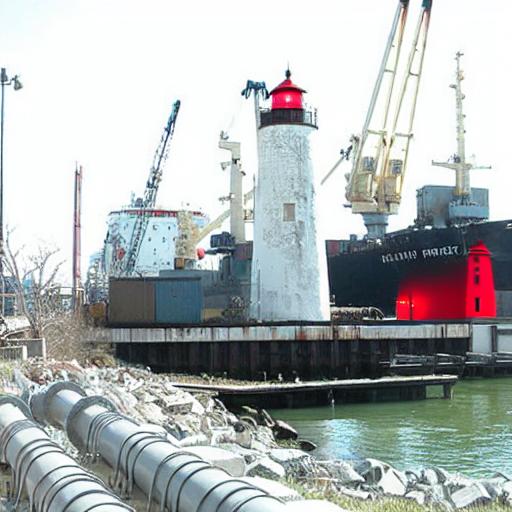}
    \end{minipage}
    \begin{minipage}{.18\textwidth}
        \centering
        \includegraphics[width=\textwidth]{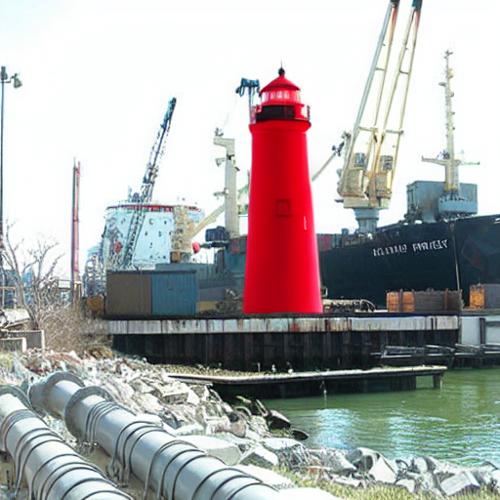}
    \end{minipage}

    \vspace{0.6em}

    \begin{minipage}{.18\textwidth}
        \centering
        (b)
    \end{minipage}
    \begin{minipage}{.72\textwidth}
        \centering
        \textit{Change the color of the lighthouse into completely red.}
    \end{minipage}
    
    \vspace{0.6em}

    \begin{minipage}{.18\textwidth}
        \centering
        \includegraphics[width=\textwidth]{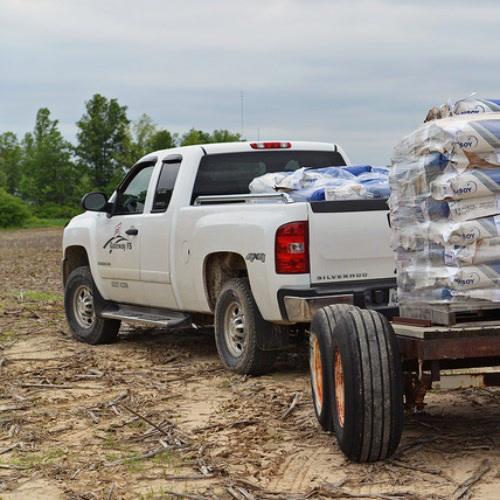}
    \end{minipage}
    \begin{minipage}{.18\textwidth}
        \centering
        \includegraphics[width=\textwidth]{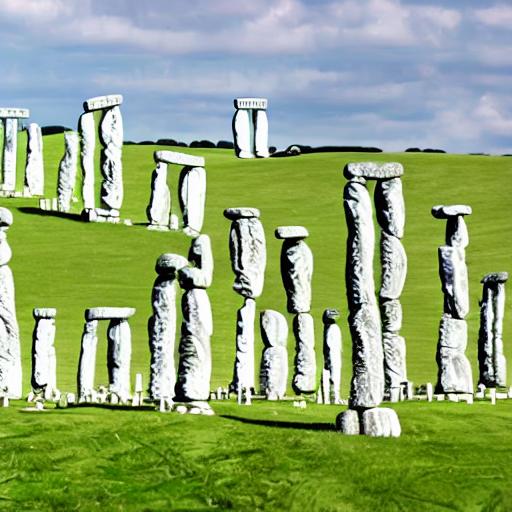}
    \end{minipage}
    \begin{minipage}{.18\textwidth}
        \centering
        \includegraphics[width=\textwidth]{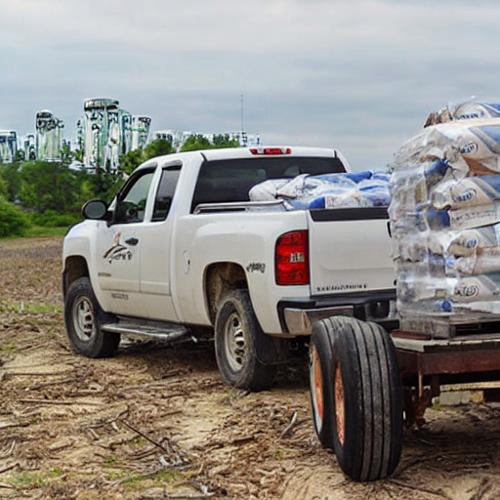}
    \end{minipage}
    \begin{minipage}{.18\textwidth}
        \centering
        \includegraphics[width=\textwidth]{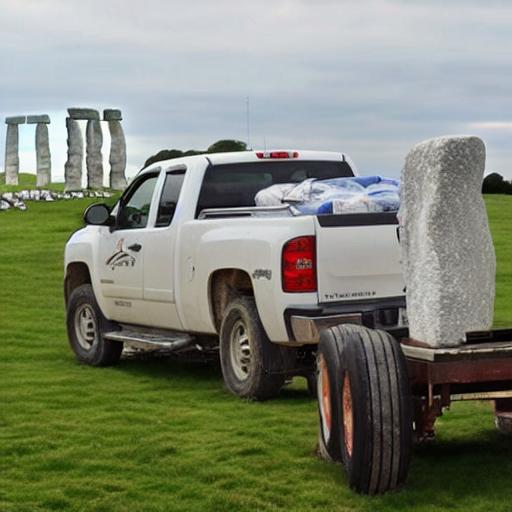}
    \end{minipage}
    \begin{minipage}{.18\textwidth}
        \centering
        \includegraphics[width=\textwidth]{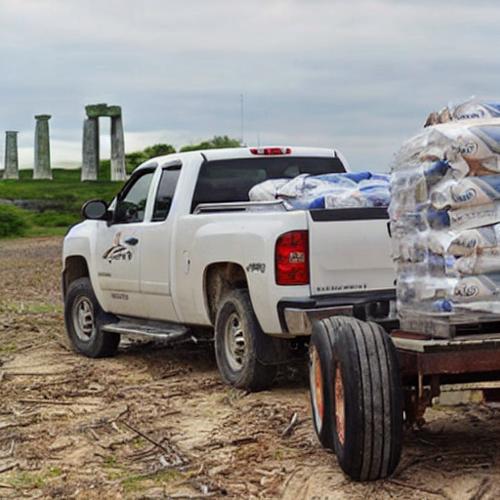}
    \end{minipage}

    \vspace{0.6em}

    \begin{minipage}{.18\textwidth}
        \centering
        (c)
    \end{minipage}
    \begin{minipage}{.72\textwidth}
        \centering
        \textit{Put Stone Henge as the background of the scene.}
    \end{minipage}

    \vspace{0.6em}

    \begin{minipage}{.18\textwidth}
        \centering
        \includegraphics[width=\textwidth]{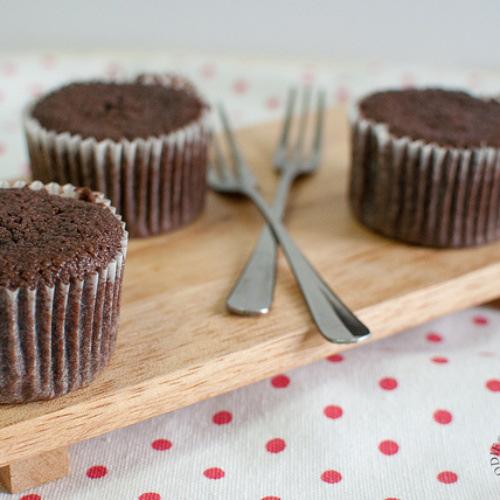}
    \end{minipage}
    \begin{minipage}{.18\textwidth}
        \centering
        \includegraphics[width=\textwidth]{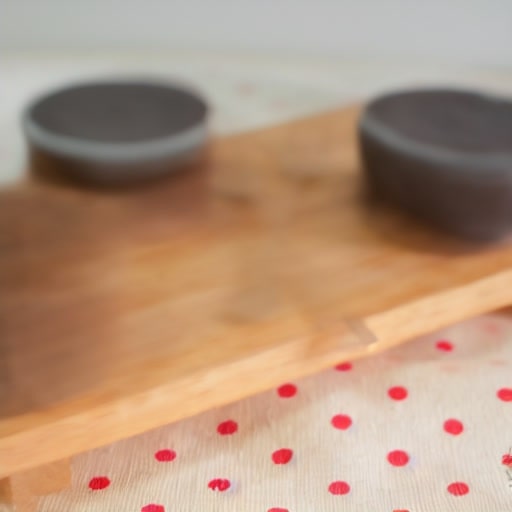}
    \end{minipage}
    \begin{minipage}{.18\textwidth}
        \centering
        \includegraphics[width=\textwidth]{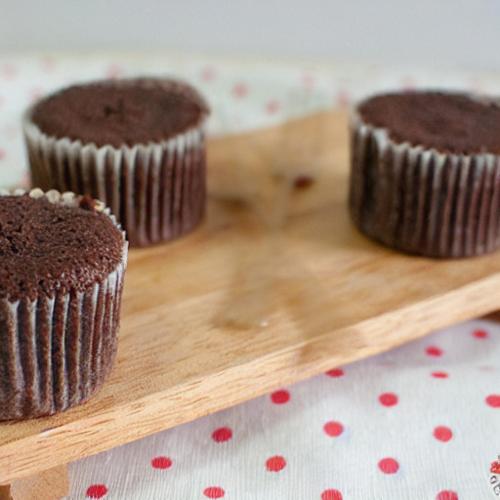}
    \end{minipage}
    \begin{minipage}{.18\textwidth}
        \centering
        \includegraphics[width=\textwidth]{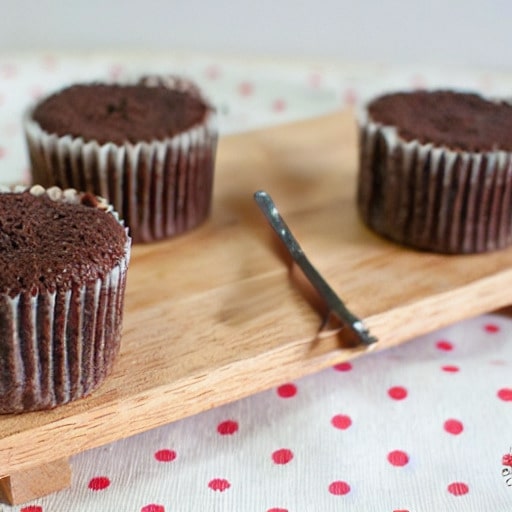}
    \end{minipage}
    \begin{minipage}{.18\textwidth}
        \centering
        \includegraphics[width=\textwidth]{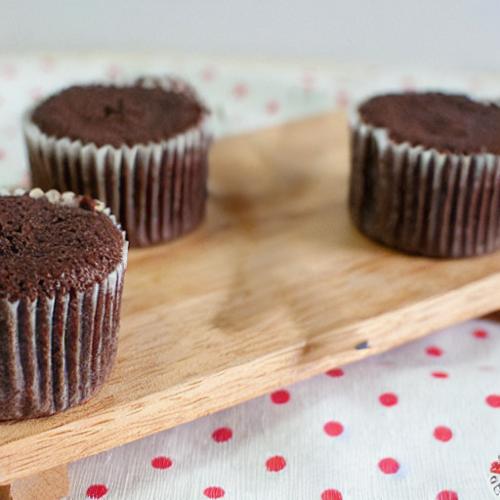}
    \end{minipage}

    \vspace{0.6em}

    \begin{minipage}{.18\textwidth}
        \centering
        (d)
    \end{minipage}
    \begin{minipage}{.72\textwidth}
        \centering
        \textit{Remove the forks from the shelf.}
    \end{minipage}

    \vspace{0.6em}

    \begin{minipage}{.18\textwidth}
        \centering
        \includegraphics[width=\textwidth]{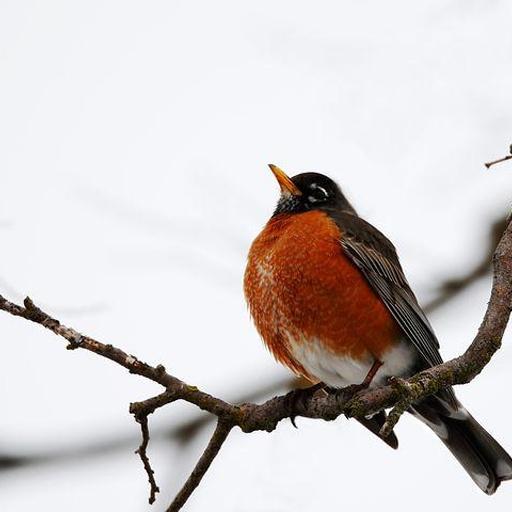}
    \end{minipage}
    \begin{minipage}{.18\textwidth}
        \centering
        \includegraphics[width=\textwidth]{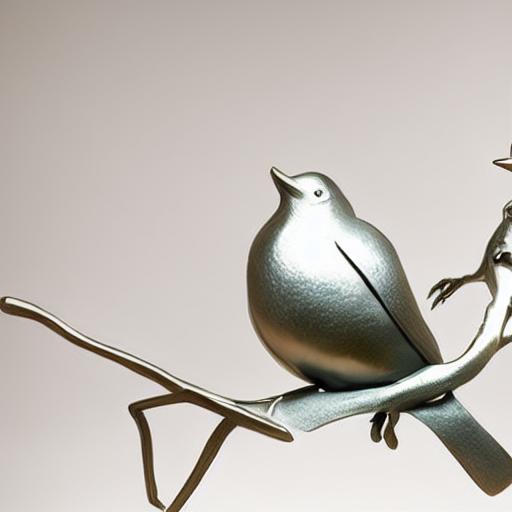}
    \end{minipage}
    \begin{minipage}{.18\textwidth}
        \centering
        \includegraphics[width=\textwidth]{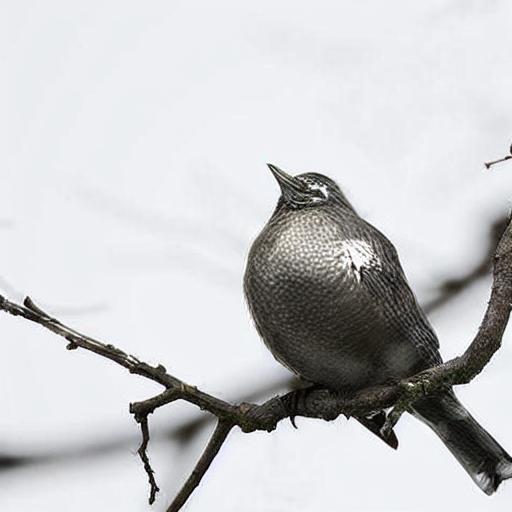}
    \end{minipage}
    \begin{minipage}{.18\textwidth}
        \centering
        \includegraphics[width=\textwidth]{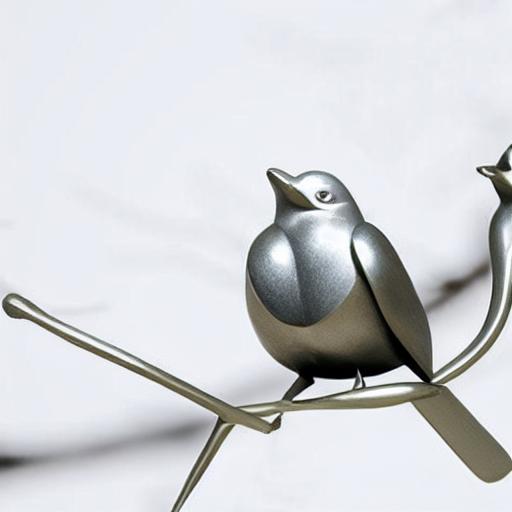}
    \end{minipage}
    \begin{minipage}{.18\textwidth}
        \centering
        \includegraphics[width=\textwidth]{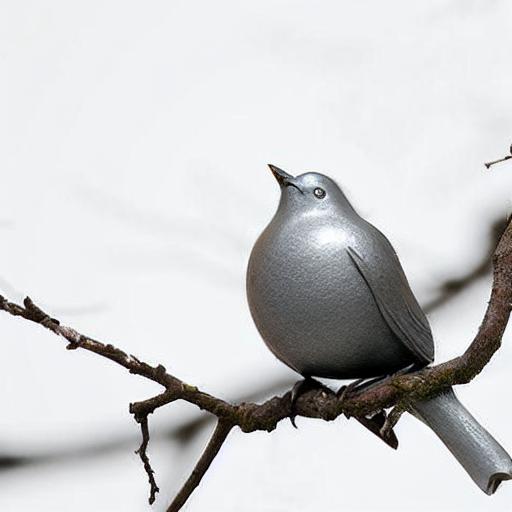}
    \end{minipage}

    \vspace{0.6em}

    \begin{minipage}{.18\textwidth}
        \centering
        (e)
    \end{minipage}
    \begin{minipage}{.72\textwidth}
        \centering
        \textit{Change the robin to a silver robin sculpture.}
    \end{minipage}

    \caption{\textbf{More Qualitative Examples.} We test our method on different tasks: (a) modifying multiple objects, (b) changing color, (c) adding, (d) removing objects, and (e) changing texture. %
    The integration of {\textbf{{\model{}}}} enhances the performance of all models, enabling localized edits while maintaining the integrity of the remaining image areas.} 
    \label{fig:qual_examples}
\end{figure*}

%% file: tables/quantitative_magicbrush.tex
  \centering

  \resizebox{\linewidth}{!}{%
  \begin{tabular}{@{}lccccccccccc@{}}
\toprule
\multicolumn{1}{c}{\multirow{2}{*}{\textbf{Methods}}} & &\multicolumn{5}{c}{\textbf{Single-turn}}& \multicolumn{5}{c}{\textbf{Multi-turn}} \\
\cmidrule(lr){3-7} \cmidrule(lr){8-12}
 & \textbf{MB} & L1~$\downarrow$ & L2~ $\downarrow$ & CLIP-I~$\uparrow$ & DINO~$\uparrow$  &CLIP-T~$\uparrow$& L1~$\downarrow$ & L2~ $\downarrow$ & CLIP-I~$\uparrow$ & DINO~$\uparrow$ & CLIP-T~$\uparrow$\\
\midrule
Open-Edit~\cite{Liu2020Open-Edit}           & \xmark & 0.143& 0.043& 0.838& 0.763&0.261& 0.166& 0.055& 0.804& 0.684& 0.253 \\
VQGAN-CLIP~\cite{Crowson2022VQGAN-CLIP}     & \xmark & 0.220& 0.083& 0.675& 0.495&\textbf{0.388}& 0.247& 0.103& 0.661& 0.459& \textbf{0.385} \\
SDEdit~\cite{Meng2022SDEdit}             & \xmark & 0.101& 0.028& 0.853& 0.773&{0.278}& 0.162& 0.060& 0.793& 0.621& 0.269 \\
Text2LIVE~\cite{Bar-Tal2022Text2LIVE}       & \xmark & 0.064& \textbf{0.017}& {0.924}& {0.881}&0.242& {0.099}& \textbf{0.028}& {0.880}& {0.793}& {0.272} \\
Null-Text Inv.~\cite{Mokady2022NullTextInversion} & \xmark & {0.075}& \underline{0.020}& \underline{0.883}& {0.821}&0.274&{ 0.106}& {0.034}& {0.847}& {0.753}& 0.271 \\ %

IP2P~\cite{Brooks2022InstructPix2Pix} & \xmark & 0.112& 0.037& 0.852& 0.743&0.276& 0.158& 0.060& 0.792& 0.618& 0.273\\
IP2P~\cite{Brooks2022InstructPix2Pix} {\textbf{+ {\model{}}}} & \xmark & \underline{0.058}& \textbf{0.017}& \underline{0.935}& \underline{0.906}&{0.293}& \underline{0.094}& {0.033}& 0.883& 0.817& {0.284}\\ %
IP2P~\cite{Brooks2022InstructPix2Pix} & \cmark & 0.063& \underline{0.020}& 0.933& 0.899&0.278& 0.096& 0.035& \underline{0.892}& \underline{0.827}& 0.275\\
IP2P~\cite{Brooks2022InstructPix2Pix} {\textbf{+ {\model{}}}}& \cmark & \textbf{0.056}& \textbf{0.017}& \textbf{0.939}& \textbf{0.911}&\underline{0.297}& \textbf{0.088}& \underline{0.030}& \textbf{0.894}& \textbf{0.835}& \underline{0.294}\\
\bottomrule
  \end{tabular}
  }
\caption{\textbf{Evaluation on MagicBrush~\cite{Zhang2023MagicBrush}.} Results for single- and multi-turn settings are presented for each method and \textit{MB} stands for models fine-tuned on MagicBrush. The other approaches are sourced from \cite{Zhang2023MagicBrush}, while values for our method are computed by following the same protocol. Across both settings, the integration of {\textbf{{\model{}}}} surpasses the base model performance, \eg{,} IP2P and IP2P w/MB.}  %

\label{tab:magicbrush} %

%% file: tex_figures/ip2p_quantitative.tex
    \centering
    \includegraphics[width=0.86\linewidth]{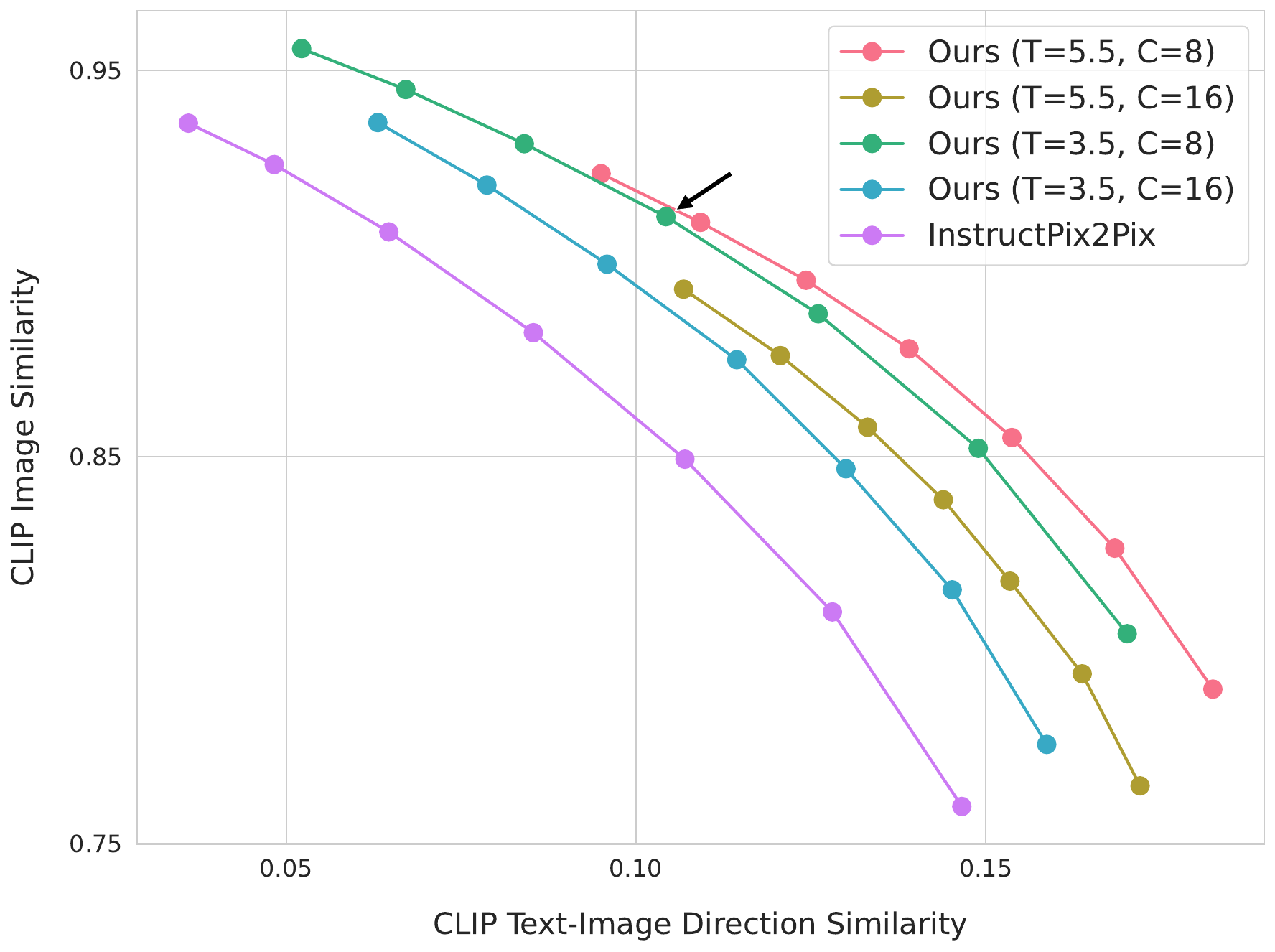}
    \caption{\textbf{Evaluation on IP2P Test Split.} Trade-off between input image and edit is showed. T and C denotes $s_T$ and \# of clusters, respectively. For all experiments, $s_I \in [1.0, 2.2]$ is fixed.} %
    \label{fig:IP2P}

%% file: tables/user_study.tex
\setlength{\columnsep}{6pt}

\begin{wraptable}{R}{0.48\linewidth}
    \scriptsize
    \centering
    \vspace{-1.6em}

    \caption{\textbf{User Study.}} %
    \vspace{-1.6em}
    
    \resizebox{\linewidth}{!}{%

    \begin{tabular}{lcc}
        \toprule
        \textbf{Options} & \textbf{(Q1)} & \textbf{(Q2)} \\ \midrule
        IP2P~\cite{Brooks2022InstructPix2Pix} & 22\% & 25\%\\
        \;\; \textcolor{LimeGreen}{\textbf{+ {\model{}}}} & \textbf{72\%}& \textbf{65\%}\\
        Neither & 6\% & 10\% \\ \midrule
        IP2P w/MB~\cite{Zhang2023MagicBrush} & 21\% &25\% \\
        \;\; \textcolor{LimeGreen}{\textbf{+ {\model{}}}} & \textbf{74\%}& \textbf{63\%}\\
        Neither & 5\% & 12\% \\ \bottomrule
    \end{tabular}}
    \vspace{-1.5em}
\label{tab:user_sduty}

\end{wraptable}

%% file: tables/ablation.tex
\begin{table}[!b]

    \vspace{-1.2em}
  \centering
  \caption{\textbf{Ablation Study.} For all experiments, IP2P is the base architecture and the evaluation is on the MagicBrush dataset. Each parameter is modified separately, while other parameters are kept fixed to isolate their impact. }
  \resizebox{0.95\linewidth}{!}{%
  \begin{tabular}{c|cccccc}
    \toprule
    & \textbf{Method} & L1~$\downarrow$  & L2~ $\downarrow$  & CLIP-I~$\uparrow$ & DINO~$\uparrow$  & CLIP-T~$\uparrow$ \\ \midrule
      & IP2P~\cite{Brooks2022InstructPix2Pix}  & 0.112 & 0.037 & 0.852 & 0.743 & 0.276 \\ \midrule
      
        \multirow{3}{*}{\rotatebox[origin=c]{90}{%
            \begin{tabular}{@{}c@{}}
                \textbf{Mask}  \\
                \textbf{Type}
            \end{tabular}%
        }} & GT    & \underline{0.063} & \textbf{0.017} & \textbf{0.935} & \underline{0.902} & \textbf{0.297} \\
        &LPM~\cite{patashnik2023localizing}  & 0.072 & \underline{0.019} & \underline{0.924} & 0.886 & 0.291 \\
        &Ours  & \textbf{0.058} & \textbf{0.017} & \textbf{0.935} & \textbf{0.906} & \underline{0.293} \\ \midrule

        \multirow{2}{*}{\rotatebox[origin=c]{90}{%
            \begin{tabular}{@{}c@{}}
                \textbf{Edit}  \\
                \textbf{Type}
            \end{tabular}%
        }} & Noise  & \underline{0.076}& \underline{0.022}& \underline{0.914}& \underline{0.864}& \underline{0.291}\\ 
        & Ours  & \textbf{0.058} & \textbf{0.017} & \textbf{0.935} & \textbf{0.906} & \textbf{0.293} \\ 
        \midrule

        \multirow{4}{*}{\rotatebox[origin=c]{90}{%
            \begin{tabular}{@{}c@{}}
                \textbf{\# of}  \\
                \textbf{Points}
            \end{tabular}%
        }} & $25$  & 0.079 & 0.023 & 0.917 & 0.874 & 0.290 \\
        &$100$ & \textbf{0.058} & \textbf{0.017} & \textbf{0.935} & \textbf{0.906} & \underline{0.293} \\
        &$225$ & \underline{0.065} & \underline{0.018} & \underline{0.932} & \underline{0.901} & \textbf{0.295} \\
        &$400$ & 0.070 & 0.020 & 0.925 & 0.889 & \textbf{0.295} \\ \midrule
        
        \multirow{4}{*}{\rotatebox[origin=c]{90}{%
            \begin{tabular}{@{}c@{}}
                \textbf{\# of}  \\
                \textbf{Clusters}
            \end{tabular}%
        }} & $4$  & 0.080 & 0.022 & 0.923 & 0.885 & \textbf{0.295} \\
        & $8$  & \textbf{0.058} & \textbf{0.017} & \textbf{0.935} & \textbf{0.906} & 0.293 \\
        & $16$ & \underline{0.062} & \underline{0.018} & \underline{0.933} & \underline{0.903} & \underline{0.294} \\
        & $32$ & 0.064 & \underline{0.018} & 0.932 & 0.901 & 0.291 \\

      \bottomrule
      \end{tabular}
      }

\label{tab:ablation}

\end{table}

%% file: sections/6_conclusion.tex
\section{Conclusion}
In this paper, we introduce, \model{}, a novel \emph{localized image editing} technique using IP2P modified with explicit segmentation of the edit area and attention regularization. This approach effectively addresses the challenges of precision and context preservation in localized editing, eliminating the need for user input or model fine-tuning/retraining. The attention regularization step of our method can also be utilized with a user-specified mask, offering additional flexibility. Our method's robustness and effectiveness are validated through empirical evaluations, outperforming existing state-of-the-art methods. This advancement contributes to the continuous evolution of LDMs in image editing, pointing toward exciting possibilities for future research.

\input{tex_figures/limitations}

\noindent\paragraph{Limitations.} Our method uses IP2P and inherits its entanglement issue discussed in \cite{Brooks2022InstructPix2Pix}, see \cref{fig:limitations} (left), although \model{} significantly helps with disentanglement. ~\cref{fig:limitations} (right) illustrates specific limitations introduced by \model{} connected to prompt quality. During the editing process, all tokens except \textit{SoT}, \textit{stop words}, and \textit{padding}, affect the RoI, leading to feature mixing. This limitation could be overcome by altering the edit instruction by using LLMs.

%% file: tex_figures/limitations.tex
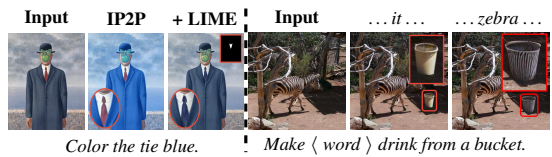
\begin{figure}[ht!]

    \vspace{-1em}
    
    \centering
    \scriptsize

    \tikz[remember picture,overlay] \node [anchor=base] (linebase) at ([xshift=-0.053\linewidth]0,0) {};

    \begin{minipage}{.12\linewidth}
        \centering
        \textbf{Input}
    \end{minipage}
    \begin{minipage}{.12\linewidth}
        \centering
        \textbf{IP2P}
    \end{minipage}
    \begin{minipage}{.12\linewidth}
        \centering
        \textbf{{+ \model{}}}
    \end{minipage}
    \begin{minipage}{.155\linewidth}
        \centering
        \textbf{Input}
    \end{minipage}
    \begin{minipage}{.155\linewidth}
        \centering
        \textit{\dots it \dots}
    \end{minipage}
    \begin{minipage}{.155\linewidth}
        \centering
        \textit{\dots zebra \dots}
    \end{minipage}

    \vspace{0.2em}

    \begin{minipage}{.12\linewidth}
        \centering
        \includegraphics[width=\textwidth]{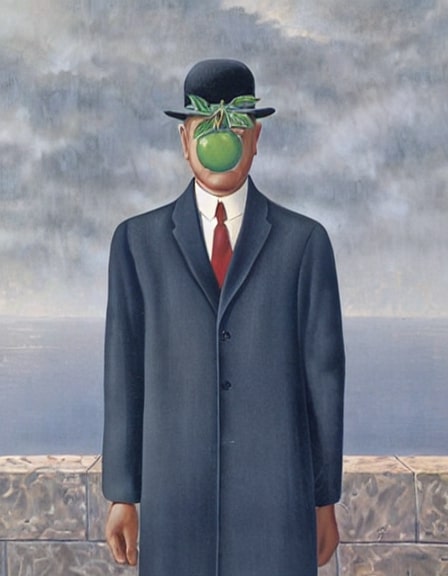}
    \end{minipage}
    \begin{minipage}{.12\linewidth}
        \centering
        \includegraphics[width=\textwidth]{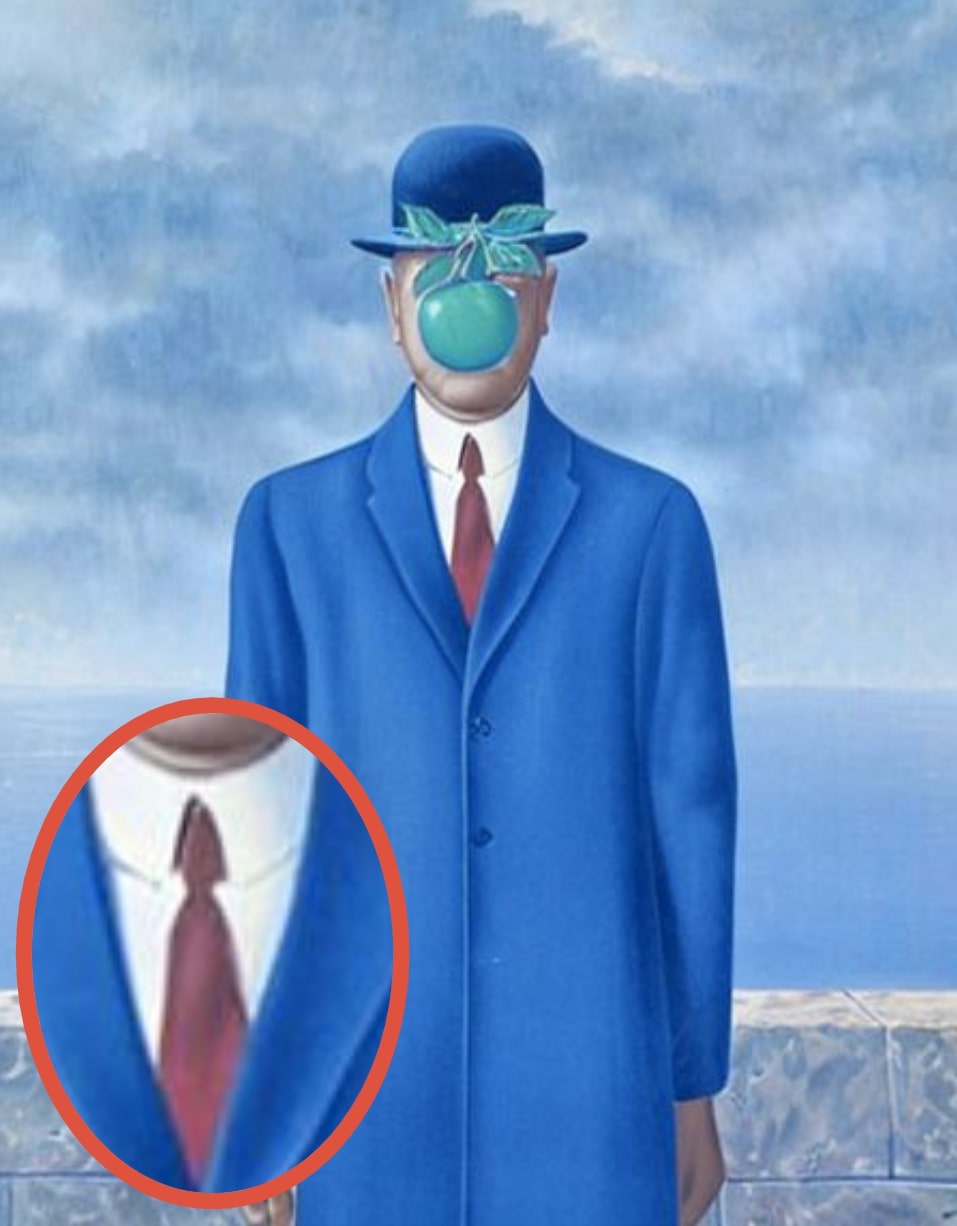}
    \end{minipage}
    \begin{minipage}{.12\linewidth}
        \centering
        \includegraphics[width=\textwidth]{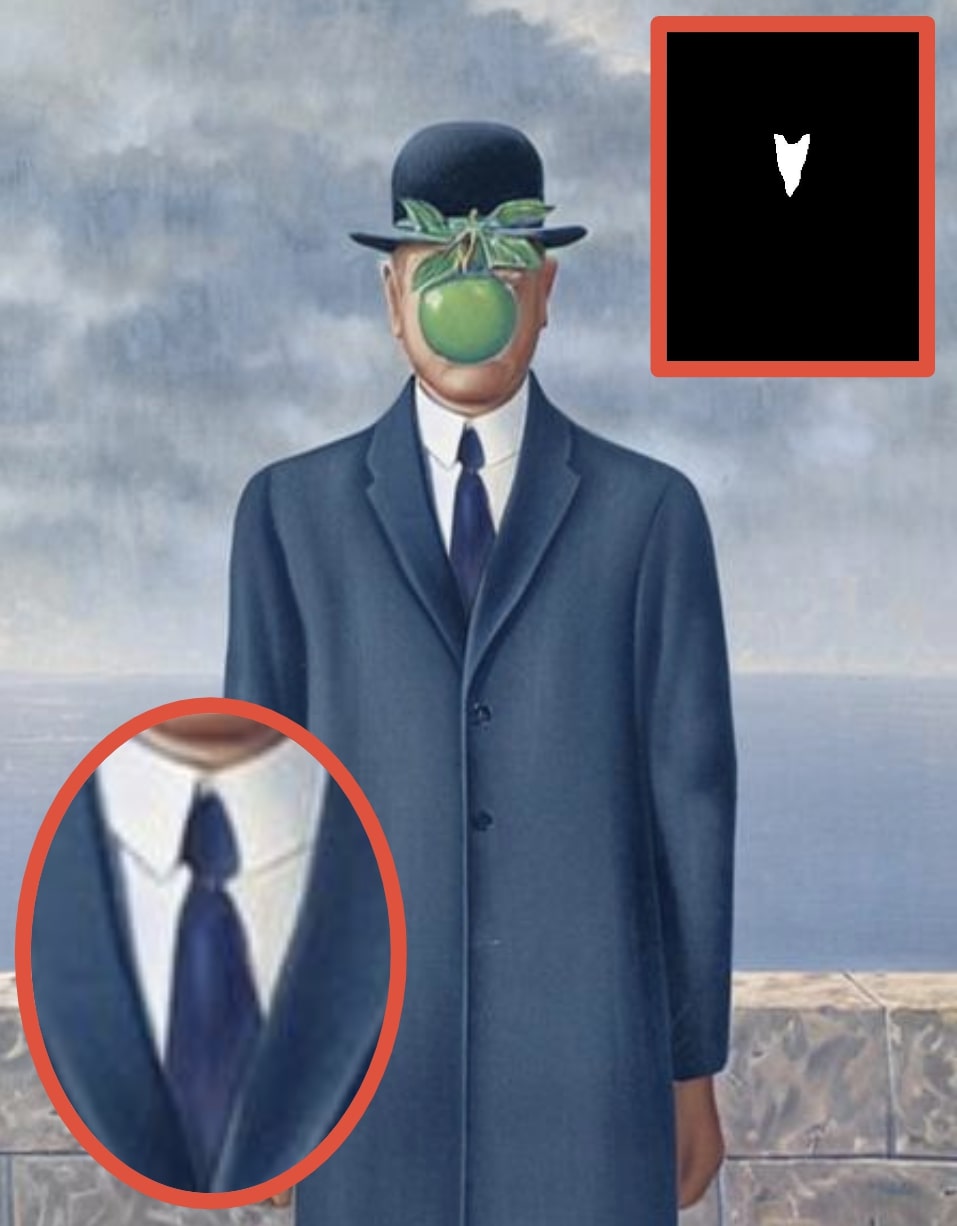}
    \end{minipage}
    \begin{minipage}{.155\linewidth}
        \centering
        \includegraphics[width=\textwidth]{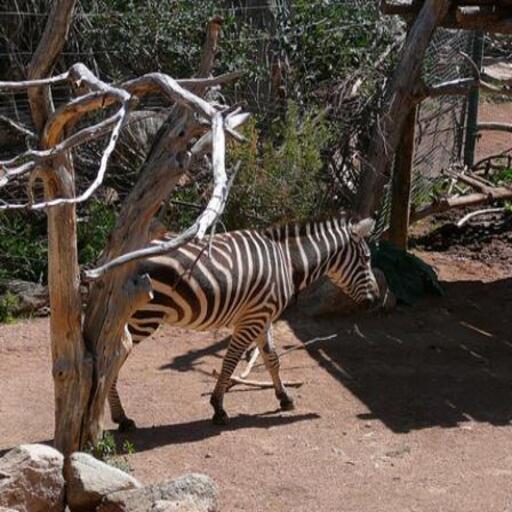}
    \end{minipage}
    \begin{minipage}{.155\linewidth}
        \centering
        \includegraphics[width=\textwidth]{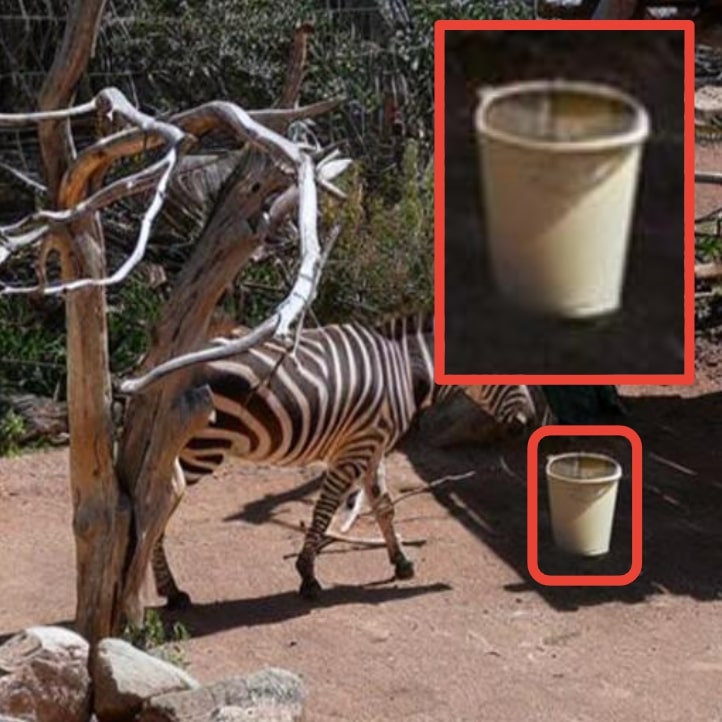}
    \end{minipage}
    \begin{minipage}{.155\linewidth}
        \centering
        \includegraphics[width=\textwidth]{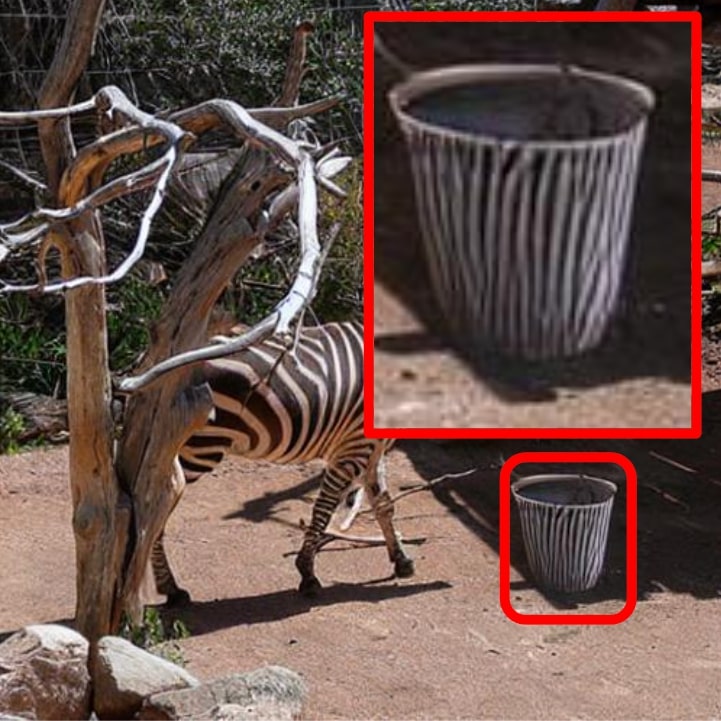}
    \end{minipage}

    \vspace{0.2em}

    \begin{minipage}{.36\linewidth}
        \centering
        \textit{Color the tie blue.}
    \end{minipage}
    \begin{minipage}{.465\linewidth}
        \centering
        \textit{Make $\langle$ word $\rangle$ drink from a bucket.}
    \end{minipage}

    \begin{tikzpicture}[remember picture, overlay]
        \draw[dashed, very thick] (linebase) -- ++(0,-2); %
    \end{tikzpicture}

    \vspace{-1.4em}
    
    \caption{\textbf{Failure Cases \& Limitations.}} %
     \label{fig:limitations} \vspace{-1.4em}
\end{figure}

%% file: sections/7_suppl.tex
\counterwithin*{subsection}{section}
\renewcommand{\thesection}{\Alph{section}}

\setcounter{page}{1}

\setcounter{figure}{7}
\setcounter{table}{2}
\setcounter{section}{0}
\definecolor{eccvblue}{rgb}{0.12,0.49,0.85}

We further explore \model{}'s applicability to other models like Prompt-to-Prompt~\cite{Hertz2022Prompt2prompt}, HIVE~\cite{Zhang2023HIVE},  and InstructDiffusion~\cite{geng2023instructdiffusion}, showing improved localized editing through quantitative and qualitative results across datasets like MagicBrush~\cite{Zhang2023MagicBrush}, PIE-Bench~\cite{ju2023direct}, and EdilVal~\cite{basu2023editval}. We provide ablation studies and implementation details and discuss broader impacts like potential misuse risks balanced against benefits like enhanced creative expression. The material underscores \model{}'s ability to enable precise localized image edits while preserving surrounding context.

\startcontents[appendices]
\printcontents[appendices]{l}{1}{\setcounter{tocdepth}{2}}

\section{Additional Experiments}

\subsection{Applicability of \model{} to other models}
The core concepts behind \model{} make it broadly applicable to a variety of image editing models, including Prompt-to-Prompt~\cite{Hertz2022Prompt2prompt}, HIVE~\cite{Zhang2023HIVE}, and InstructDiffusion~\cite{geng2023instructdiffusion}. Integrating \model{} into these methods offers the potential for enhanced performance in the following ways:

\paragraph{Prompt-to-Prompt} is a prompt-based editing method~\cite{Hertz2022Prompt2prompt}. Unlike instruction-based editing techniques such as IP2P, an input caption and an output caption are needed to execute the desired image edit. Moreover, since there is no condition on the input image, an image inversion step is required before applying the edit. Prompt-to-Prompt does offer localized editing capabilities, with an extension of the \textit{Blend} option, which mixes the diffusion processes of two images (original and edited). However, as seen in~\cref{fig:p2p_integration}, integrating the edit application of \model{} into Prompt-to-Prompt enables even more precise localized edits. This component, see ~\cref{sec:edit_application}, ensures that changes are confined to the RoI while preserving the shapes and context of surrounding elements. The official code base\footnote{\href{https://github.com/google/prompt-to-prompt/}{\textit{https://github.com/google/prompt-to-prompt/}}} is used for comparison.

\input{tex_figures/p2p_integration}

\paragraph{HIVE} is a fine-tuned version of IP2P on an expanded dataset. Further refinement is achieved through fine-tuning with a reward model, which is developed based on human-ranked data. 
\Cref{tab:MB_GIVE} shows the results on the MagicBrush dataset. HIVE~\cite{Zhang2023HIVE} improves the performance of IP2P (compare the first and third rows), and further improvement is achieved by fine-tuning HIVE on MagicBrush training set, MB with \cmark{} stands for it in \cref{tab:MB_GIVE}, (compare the third and fifth rows). For both base HIVE and the version fine-tuned on MB, \model{} can further significantly improve performance (compare the third and fourth rows and fifth and sixth rows.). 

\input{tables/quantitative_MB_HIVE}

\cref{fig:hive_integration} presents a qualitative comparison before and after integrating \model{} into the HIVE model on samples from the MagicBrush dataset. \Cref{fig:hive_integration}-(a) displays the color change of an object in a scene. While HIVE can effectively implement the edit, it inadvertently alters the structure of the vase and the color of another vase in the background, as highlighted by the red circle. However, with our model integrated, HIVE accurately targets the plants for the desired edit without affecting unrelated areas.
In \cref{fig:hive_integration}-(b), HIVE fails to recognize one of the fingernails and does not correctly apply the intended edit, indicated by the red circle and arrows. Additionally, HIVE alters the background color to blue, which was mentioned in the edit prompt. In contrast, HIVE, with our model integration, precisely applies edits to the region of interest, such as the fingernails, without missing any parts or altering areas outside the region of interest.
Lastly, \cref{fig:hive_integration}-(c) demonstrates another instance where HIVE performs an entangled edit, changing the skin color of a woman in the scene when the intended edit was to change the outfit color, as shown by the red arrow. The integration of our model enables localized and separate edits on the input image based on the edit instructions, ensuring that only the specified changes are made.

\input{tex_figures/hive_integration}

\paragraph{InstructionDiffusion (IDiff)} is another IP2P-based method~\cite{geng2023instructdiffusion}. Integrating \model{} into IDiff enhances its performance. IDiff achieved scores of $0.085$, $0.03$, $0.90$, $0.83$, and $0.30$ while the integration of \model{} into IDiff improves the scores to $0.071$, $0.02$, $0.92$, $0.86$, and $0.30$ for L1, L2, CLIP-I, DINO, and CLIP-T, respectively, on MagicBrush test dataset. The results can also be comparable with \cref{tab:MB_GIVE}.

\begin{figure}[ht!]
    \centering
    \begin{minipage}{.3\linewidth}
        \centering
        \textbf{Input}
    \end{minipage}
    \begin{minipage}{.3\linewidth}
        \centering
        \textbf{IDiff}
    \end{minipage}
    \begin{minipage}{.3\linewidth}
        \centering
        \textbf{+ \model{}}
    \end{minipage}
    
    \vspace{0.5em}
    
    \begin{minipage}{.3\linewidth}
        \centering
        \includegraphics[width=\linewidth]{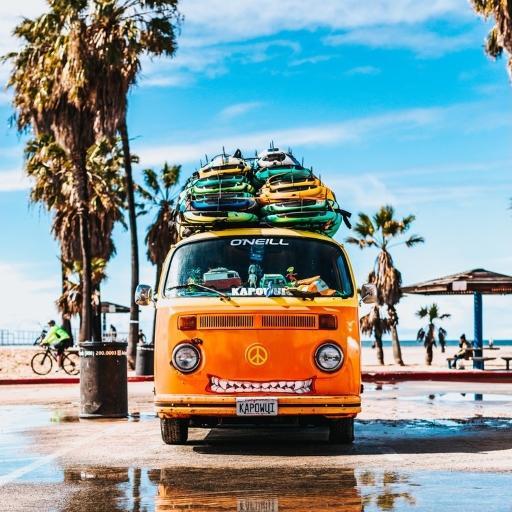}
    \end{minipage}
    \begin{minipage}{.3\linewidth}
        \centering
        \includegraphics[width=\linewidth]{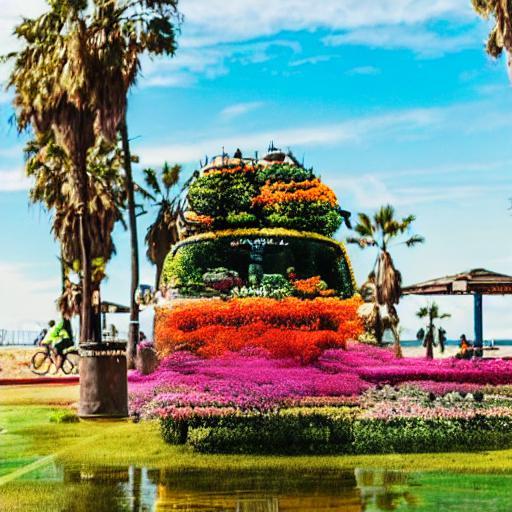}
    \end{minipage}
    \begin{minipage}{.3\linewidth}
        \centering
        \includegraphics[width=\linewidth]{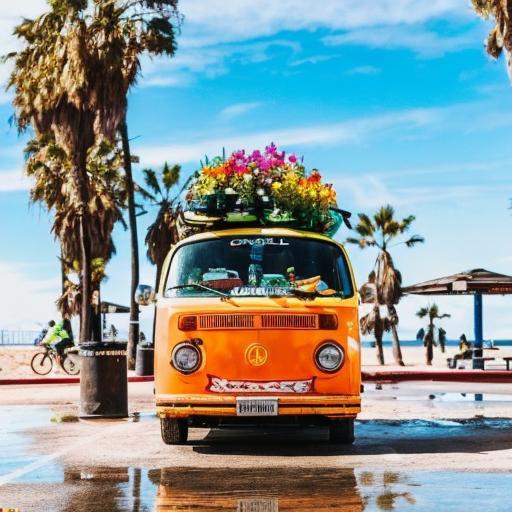}
    \end{minipage}

    \vspace{0.5em}

    \begin{minipage}{.9\linewidth}
        \centering
        Instruction: \textit{Replace the surfboards with flowers.}
    \end{minipage}
    
\end{figure}

\subsection{MagicBrush Mask Annotations}
As mentioned in \cref{sec:ablation}, the mask annotations for the MagicBrush dataset~\cite{Zhang2023MagicBrush} are not very tight around the edit area which might result in worse edit quality when we use them rather than the segmentation extracted by \model{}. We show qualitative results highlighting the problem in ~\cref{fig:magicbrush_anno}. Our method directly uses the identified mask during the editing process, therefore, it is important for the masks to be as tight as possible around the edit area to apply localized edits. The loose GT masks of MagicBrush explain why our model achieves worse performance in ~\cref{tab:ablation} when using GT masks. 
We highlight the significance of precise masks with red circles in \cref{fig:magicbrush_anno}. When precise masks are provided to \model{}, localized edits can be achieved. For the first row - (a), the handle of the racket can be preserved if the mask has a precise boundary between the handle and outfit in the occluded area. Moreover, the second row - (b) shows that if the mask in the MagicBrush dataset is used during the edit, the method changes the color of the blanket as well. However, with the precise mask extracted by our method, the edit can distinguish the objects in the area and apply localized edits.
\input{tables/quantitative_piebench}

\input{tex_figures/magicbrush_annotations}

These results highlight the quality of the edit masks extracted by our method in an entirely self-supervised way and hint at possible further developments where our contribution could be used to refine the annotations of existing datasets or speed up the creation of new ones.

\subsection{More Quantitative Results}
\paragraph{PIE-Bench~\cite{ju2023direct}} The benchmark includes 700 images in 10 editing categories with input/output captions, editing instructions, input images, and RoI annotations. Metrics for \textit{structural integrity} and \textit{background preservation} are derived from cosine similarity measures and image metrics like \textit{PSNR}, \textit{LPIPS}, \textit{MSE}, and \textit{SSIM}, while text-image consistency is evaluated via \textit{CLIP Similarity}.

Quantitative analysis on PIE-Bench~\cite{ju2023direct} demonstrates the effectiveness of our proposed method. Compared to baseline models like IP2P~\cite{Brooks2022InstructPix2Pix} and the fine-tuned version on MagicBrush~\cite{Zhang2023MagicBrush} and HIVE~\cite{Zhang2023HIVE}, our method achieves significantly better performance on metrics measuring structure and background preservation. This indicates that our approach makes localized edits according to the instructions while avoiding unintended changes to unaffected regions. \textit{Edited} measures CLIP similarity between edit prompt and edited area, and with \textit{Background Preservation} provides a measure of localized edits. As seen in~\cref{tab:pie_bench}, our method is better if both metrics are considered together. At the same time, our method obtains comparable results to base models on the CLIP similarity score, showing that edits are faithfully applied based on the textual instruction. A comprehensive comparison is presented in ~\cref{tab:pie_bench}. Overall, the quantitative results validate that our method can enable text-guided image editing by making precise edits solely based on the given edit instruction without altering unrelated parts.

\paragraph{EditVal~\cite{basu2023editval}} The benchmark offers 648 image editing operations spanning 19 classes from the MS-COCO dataset~\cite{Lin2014COCO}. The benchmark assesses the success of each edit with a binary score that indicates whether the edit type was successfully applied. The OwL-ViT~\cite{minderer2022simple} model is utilized to detect the object of interest, and detection is used to assess the correctness of the modifications.

\begin{table}[!htb]
    \centering
    \input{tables/quantitative_editval}
\end{table}

Our method exhibits superior performance across various edit types in EditVal benchmark~\cite{basu2023editval}, particularly excelling in \textit{Object Addition~(O.A.)}, \textit{Position Replacement~(P.R.)}, and \textit{Positional Addition~(P.A.)}, while achieving second-best in \textit{Object Replacement~(O.R.)}. It performs on par with other methods for edits involving \textit{Size~(S.)} and \textit{Alter Parts~(A.P.)}. \model{} advances the state-of-the-art by improving the average benchmark results by a margin of $5\%$ over the previous best model, see ~\cref{tab:editval}. 

\subsection{Visual Comparison to state-of-the-art-methods}
\subsubsection{VQGAN-CLIP} 
As shown in \cref{tab:magicbrush}, VQGAN-CLIP~\cite{Crowson2022VQGAN-CLIP} has better results on the \emph{CLIP-T} metric. This is expected since it directly fine-tunes the edited images using CLIP embeddings. However, as seen in \cref{fig:vqgan}, the edited images from VQGAN-CLIP fail to preserve the details of the input image. On the other hand, our method successfully performs the desired edit by preserving the structure and fine details of the scene and results in a similar \textit{CLIP-T} score as the one of the ground truth edited images in the MagicBrush dataset.

\input{tex_figures/vqgan}

\subsubsection{Diffusion Disentanglement} Wu \etal~\cite{wu2023uncovering} propose a disentangled attribute editing method. Since it also claims disentangled (localized) edits, we visually compare our method + IP2P with Diffusion Disentanglement. This method is not disegned for instruction-based editing, so we use input and output captions during the comparison.
\Cref{fig:disentanglement} shows edit types such as (a) texture editing and (b) replacing the object with a similar one. \textit{Diffusion Disentanglement} on (a) alters the background objects in the image, \eg{,} adding snow on and changing the shape of the branch, and also changes the features of the object of interest, \eg{,} removing the tail of the bird. On (b), it fails to perform the desired edit altogether. Moreover, it requires a GPU with \textgreater{} 48 GB RAM\footnote{\href{https://github.com/UCSB-NLP-Chang/DiffusionDisentanglement\#requirements}{\textit{https://github.com/UCSB-NLP-Chang/DiffusionDisentanglement}}}, and one image takes approximately 10 minutes on an NVIDIA A100 80GB to generate the edited version. In comparison, \model{} achieves higher visual quality and takes 25 seconds to complete on NVIDIA A100 40GB with a GPU RAM usage of 25 GB.

\input{tex_figures/disentanglement}

\subsubsection{Blended Latent Diffusion} As shown in \cref{tab:pie_bench}, Blended Latent Diffusion~\cite{avrahami2023blended} has better results than baselines and our method. However, as shown in \cref{fig:blended}, even if their method can perform the desired edit on the RoI from the user, (a) it distorts the location of the features, \eg{,} heads of the birds, and (b) it loses the information of the object in the input image and creates a new object in the RoI, \eg{,} blanket in (b). On the other hand, our method performs visually appealing edits on the input images considering the given edit instructions by preserving as many details from the input image as possible. This is also highlighted by a significantly lower Distance metric for our method in \cref{tab:pie_bench}.  

\input{tex_figures/blendeddiff}

\subsection{Qualitative comparison on segmentation maps}
Our method proposes a segmentation method based on the clustering of intermediate features of the diffusion process. In this section, we provide a qualitative comparison to other segmentation methods that could be used as an alternative to this strategy.
LPM~\cite{patashnik2023localizing} uses self-attention features from one resolution, such as $32 \times 32$, while our method leverages the intermediate features from different resolutions to enhance the segmentation map. Then, both apply a clustering method to find the segments in the input image. Another way to find segments is by using large segmentation models, \eg{,} SAM~\cite{Kirillov2023SAM}, ODISE~\cite{xu2023odise} \dots. As seen in \cref{fig:segmentation} (i), large segmentation models cannot detect the transparent fin of the fish, while LPM and ours can. Moreover, LPM utilizes only one resolution, so it cannot find rocks in the river separately. As seen in \cref{fig:segmentation} (ii), ODISE~\cite{xu2023odise} and SAM~\cite{Kirillov2023SAM} fail to segment minute object parts, like fingernails, while LPM and ours can find those segments. Furthermore, our method provides precise boundaries and segments in higher resolutions than LPM. Moreover, LPM uses Stable Diffusion~\cite{Rombach2022StableDiffusion} and requires real image inversion to find segments, while our method does not since it is based on IP2P~\cite{Brooks2022InstructPix2Pix}. For this reason, LPM requires more than 1 minute per image, while our proposal takes only 10-15 seconds per image. As a result, in a direct comparison to LPM, our method has the advantage of having higher-resolution segmentation maps with more details, and it is significantly faster. The publicly available official implementations of LPM\footnote{\href{https://github.com/orpatashnik/local-prompt-mixing}{\textit{https://github.com/orpatashnik/local-prompt-mixing}}}, SAM\footnote{\href{https://segment-anything.com/demo}{\textit{https://segment-anything.com/demo}}} and ODISE\footnote{\href{https://github.com/NVlabs/ODISE}{\textit{https://github.com/NVlabs/ODISE}}} are used for the results in ~\cref{fig:segmentation}. Additionally, the same number of clusters is used for LPM and ours to achieve a fair comparison.

\input{tex_figures/segmentation}

\subsection{Comparison with open-vocabulary segmentors}
An alternative to our proposed edit localization step, \cref{sec:edit_localization}, would be to use off-the-shelves Open Vocabulary Segmentation models~(OVS) and combine them with our \textit{edit application}, see \cref{sec:edit_application}. A key difference between OVS and our localization method is that OVS requires an additional input, which is the \textit{object of interest} to segment, and this could be a strong limitation in an instruction edit setting because the target is not always obvious from the instruction. As seen in \cref{tab:ablation_ovs}, both, SEEM~\cite{zou2024segment} and OV-SEG~\cite{liang2023open}, significantly underperform compared to \model{}.

\input{tables/ablation_ovs}

In addition to quantitative analysis in \cref{tab:ablation_ovs}, we provide a qualitative comparison for localizing the edit. As seen in \cref{fig:ovs_roi}, the OVS methods do not provide precise RoI for the edit instructions even when provided with a curated additional input, \eg{,} object of interest. On the other hand, our proposed localization method in~\cref{sec:edit_localization} provides precise and relevant RoIs without additional requirements.

\input{tex_figures/ovs_RoI}

\subsection{Ablation study}
\subsubsection{Related Token Rewarding}
In addition to the ablation study in \cref{sec:ablation}, we also analyze token selection during cross-attention regularization as defined in \cref{sec:edit_application}. Instead of regularizing the attention of unrelated tokens, such as \textit{\textless start of text\textgreater}, \textit{padding}, and \textit{stop words}, by penalizing it, we could think of doing the opposite and give high values to relevant tokens (denoted as $\tilde{S}$) within the RoI as reported in the following equation:

\vspace{-4pt}

\begin{equation}
\begin{split}
    R(QK^T, M) = \begin{cases} 
    QK^T_{ijt} \; \textcolor{eccvblue}{\textbf{+}} \; \infty, & \text{if } M_{ij} = 1 \text{ and } t \in \tilde{S}\\
    QK^T_{ijt}, & \text{otherwise},
    \end{cases}
    \label{eq:attention_penalization2}
\end{split}
\end{equation}

\vspace{-4pt}

This assignment guarantees that relevant tokens related to edit instructions have high scores after the softmax operation. As seen in \cref{tab:ablation2}, there is no significant improvement if the unrelated tokens are penalized instead of awarding the related tokens. However, penalizing the unrelated tokens gives the freedom to distribute the attention scores unequally among relevant tokens. Thus, it allows for a soft assignment of areas of the image among the related tokens.

\input{tables/ablation2}

In addition to quantitative analysis on the MagicBrush dataset, \cref{fig:ablation_tokens} shows the attention scores for rewarding related tokens vs regularizing unrelated tokens. Even though the final edit results and the last column are not significantly different, unrelated token regularization results in uneven attention scores among related tokens, while related token regularization leads to equal attention scores.

\input{tex_figures/ablation_tokens}

\subsubsection{Clustering Method Alternatives} Since our implementation uses KMeans, it requires a \textit{number of clusters} parameter to be decided and fixed. While in the main paper, we show that a good tuning of this hyperparameter works robustly across all the evaluated datasets, in this section, we perform a preliminary exploration of alternative clustering techniques that would relax this assumption. 
We replace K-means with Agglomerative Clustering, which does not require this parameter. For this implementation, we use the cosine similarity metric between features since \cite{tang2023emergent} shows that the cosine similarity provides significant information about semantics. We provide a minimum distance threshold, which can be between 0-1, and it automatically determines the number of clusters. The higher the threshold, the more clusters. As seen in~\cref{tab:ablation_clustering}, this variation achieves similar performance by properly tuning the distance threshold. This ablation study proves that our method is robust for selecting the clustering method, and any clustering method can be combined with \model{}.

\input{tables/ablation_clustering}
\input{tex_figures/usecase} %

\subsection{More Qualitative Results}
This section presents additional qualitative results derived from our method, emphasizing its improved effectiveness against established baselines, such as IP2P~\cite{Brooks2022InstructPix2Pix} and IP2P w/MB~\cite{Zhang2023MagicBrush}. \Cref{fig:usecase} illustrates the application of our method in localized image editing tasks. Specifically, it demonstrates our method's proficiency in altering the color of specific objects: (a) \textit{ottoman}, (b) \textit{lamp}, (c) \textit{carpet}, and (d) \textit{curtain}. Unlike the baseline methods, which tend to entangle the object of interest with surrounding elements, our approach achieves precise, disentangled edits. This is not achieved by the baseline, which tends to alter multiple objects simultaneously rather than isolating changes to the targeted region. The disentangled and localized edits showcased in \cref{fig:usecase} highlight the potential of \model{} in end-user applications where object-specific edits are crucial.

\Cref{fig:more_qualitatives1} demonstrates additional examples of our method's performance on the Emu-Edit test set~\cite{sheynin2023emu}, the MagicBrush~\cite{Zhang2023MagicBrush} test set and the PIE-Bench~\cite{ju2023direct} dataset. Our approach effectively executes various tasks, such as (a) replacing an animal, (b) modifying parts of animals, and (c) changing the color of multiple objects. As illustrated in \cref{fig:more_qualitatives1}, our method demonstrates significant improvements over existing baselines. For instance, while baseline models like IP2P w/MB in (a) achieve reasonable edits, they often inadvertently modify areas outside the RoI, as observed in cases (b) and (c). Notably, our method helps focus the baseline models on the RoI, as seen in (b) and (c), where baselines struggle to preserve the original image. Although our method is dependent on the baseline and may occasionally induce unintended changes in peripheral areas, \eg{,} the floor's color, it consistently outperforms the baseline models in terms of targeted and localized editing.

\section{Implementation Details}
We obtain the results using an NVIDIA A100 40GB GPU machine. For $512 \times 512$ images the IP2P-based baselines (\eg{,} IP2P, IP2P w/MB, HIVE, and HIVE w/MB) take approximately 15 seconds per edit, while for \model{} integrated models, it takes $\approx$25 seconds.

\subsection{User Study Setting} 
\begin{figure}[!hb]
    \centering
    \includegraphics[width=\linewidth]{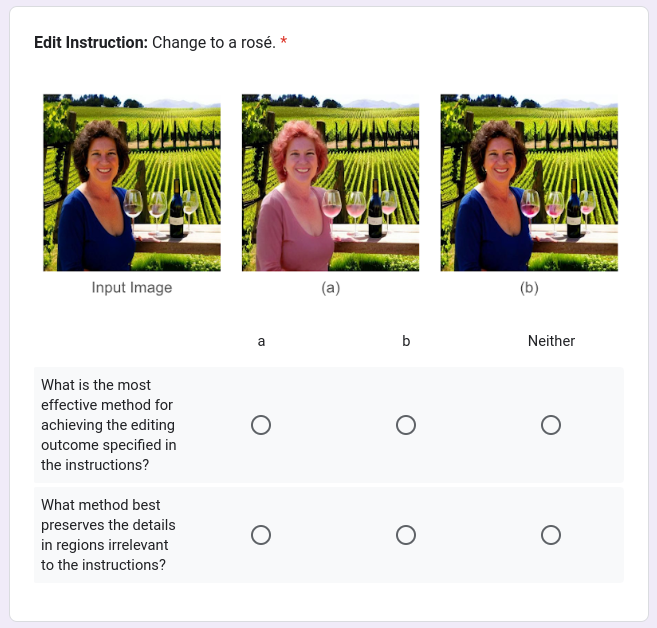}
    \caption{\textbf{User study setting.} The example with the edit instruction with the corresponding input image and randomly ordered two edited images from baseline and baseline + \model{}.}
    \label{fig:user_study}
\end{figure}

We carry out a study with 54 questions involving users, asking 53 anonymous individuals on the crowd-sourcing platform Prolific~\cite{prolific}. In our user study, participants will be presented with two alternative edited images alongside their corresponding input images and editing instructions. They will be tasked with evaluating the effectiveness of the edits in achieving the specified outcome and the ability of the editing method to preserve the details in areas not targeted by the instruction. Using the example provided in \cref{fig:user_study}, where the editing instruction is to \textit{Change to a rosé}, participants must discern which edited image (a, b or neither) not only best satisfies this directive but also maintains the fidelity of the scene's irrelevant aspects. The aggregated data from participant responses will yield insights into the preferred methods for both accurate editing and detail preservation, thereby influencing the development of advanced image editing methods.

\input{tex_figures/more_qualitatives1}

\subsection{Reproducibility Statement}
We guarantee that all the results presented in the main manuscript and the supplementary materials can be reproduced. 
We will make our code base containing examples and hyperparameters public so that the results can be reproduced. 

\subsection{Baselines}
\noindent\textbf{Open-Edit~\cite{Liu2020Open-Edit}:} This GAN-based approach uses a reconstruction loss for pre-training and incorporates a consistency loss during fine-tuning on specific images. Its unique feature is the arithmetic manipulation of word embeddings within a shared space of visual and textual features.

\noindent\textbf{VQGAN-CLIP~\cite{Crowson2022VQGAN-CLIP}:} Enhancing the VQGAN~\cite{Esser2021VQGAN} framework with CLIP embeddings~\cite{Radford2021CLIP}, this method fine-tunes VQGAN using the similarity of CLIP embeddings between the generated image and the target text, leading to optimized image generation.

\noindent\textbf{SDEdit~\cite{Meng2022SDEdit}:} Leveraging the capabilities of Stable Diffusion~\cite{Rombach2022StableDiffusion}, SDEdit introduces a tuning-free approach. It uses stochastic differential equation noise, adding it to the source image and subsequently denoising to approximate the target image, all based on the target caption.

\noindent\textbf{Text2LIVE~\cite{Bar-Tal2022Text2LIVE}:} It propose a Vision Transformer~\cite{Dosovitskiy2021ViT} for generating edited objects on an additional layer. It incorporates data augmentation and CLIP~\cite{Radford2021CLIP} supervision, ultimately alpha-blending the edited layer with the original to create the target image.

\noindent\textbf{Null Text Inversion~\cite{Mokady2022NullTextInversion}:} By optimizing the DDIM~\cite{Song2021DDIM} trajectory, this method initially inverts the source image. After, it performs image editing during the denoising process guided by cross-attention~\cite{Hertz2022Prompt2prompt} between text and image.

\noindent\textbf{SINE~\cite{zhang2022sine}:} Real images are edited using model-based guidance and patch-based fine-tuning process.

\noindent\textbf{DreamBooth~\cite{Ruiz2022DreamBooth}:} It fine-tunes a diffusion model by learning special text tokens and adjusting model parameters on a set of images for editing.

\noindent\textbf{Textual-Inversion~\cite{gal2022image}:} It fine-tunes a token embedding within the text-encoder space using a set of images.

\noindent\textbf{Imagic~\cite{Kawar2022Imagic}:} It edits images through a three-step process: first fine-tuning a token embedding, then fine-tuning the parameters of a text-guided image diffusion model using the fine-tuned token embedding, and finally performing interpolation to generate various edits based on a target prompt.

\noindent\textbf{DiffEdit~\cite{couairon2023diffedit}:} It identifies the region to edit in images by contrasting between a conditional and unconditional diffusion model based on query and reference texts. Then, it reconstructs the edited image by collecting the features from the text-query by combining the features in the noise/latent space, considering the region to edit.

\noindent\textbf{Blended Latent Diffusion~\cite{avrahami2023blended}:} This method uses a text-to-image Latent Diffusion Model (LDM) to edit the user-defined mask region. It extracts features for the mask region from the edit text, and for the rest of the image, it uses features from the original image in the noise/latent space.

\noindent\textbf{DirectDiffusion~\cite{ju2023direct}:} It inverts the input image into the latent space of Stable Diffusion~\cite{Rombach2022StableDiffusion} and then applies Prompt2Prompt~\cite{Hertz2022Prompt2prompt} to obtain the desired edit without making any changes to the edit diffusion branch.

\noindent\textbf{Diffusion Disentanglement~\cite{wu2023uncovering}:} It finds the linear combination of the text embeddings of the input caption and the desired edit to be performed. Since it does not fine-tune Stable Diffusion parameters, they claim that the method performs disentangled edits.

\noindent\textbf{InstructPix2Pix~(IP2P)~\cite{Brooks2022InstructPix2Pix}:} Starting from the foundation of Stable Diffusion~\cite{Rombach2022StableDiffusion}, the model is fine-tuned for instruction-based editing tasks. It ensures that the edited image closely follows the given instructions while maintaining the source image without the need for test-time tuning.

\noindent\textbf{InstructPix2Pix w/MagicBrush~\cite{Zhang2023MagicBrush}:} A version of IP2P~\cite{Brooks2022InstructPix2Pix} trained on MagicBrush train set~\cite{Zhang2023MagicBrush}. Since the MagicBrush dataset has more localized edit examples, the fine-tuned version has better results, as seen in \cref{tab:magicbrush}.

\noindent\textbf{HIVE~\cite{Zhang2023HIVE}:} It extends IP2P~\cite{Brooks2022InstructPix2Pix} by fine-tuning it with an expanded dataset. Further refinement is achieved through fine-tuning with a reward model, which is developed based on human-ranked data.

\noindent\textbf{HIVE w/MagicBrush~\cite{Zhang2023MagicBrush}:} HIVE~\cite{Zhang2023HIVE} fine-tuned on MagicBrush train set~\cite{Zhang2023MagicBrush}. Since the MagicBrush dataset has more localized edit examples, the fine-tuned version has better results, as seen in \cref{tab:magicbrush}.

\section{Broader Impact \& Ethical Considerations}

The advancement in localized image editing technology holds significant potential for enhancing creative expression and accessibility in digital media and virtual reality applications. However, it also raises critical ethical concerns, particularly regarding its misuse for creating deceptive imagery like deepfakes~\cite{korshunov2018deepfakes} and the potential impact on job markets in the image editing sector. Ethical considerations must focus on promoting responsible use, establishing clear guidelines to prevent abuse, and ensuring fairness and transparency, especially in sensitive areas like news media. Addressing these concerns is vital for maximizing the technology's positive impact while mitigating its risks.

%% file: tex_figures/p2p_integration.tex
\begin{figure}[!ht]
    
    \centering
    \scriptsize{
        \begin{minipage}{.24\linewidth}
            \centering
            \textbf{Inverted Image}
        \end{minipage}
        \hfill
        \begin{minipage}{.24\linewidth}
            \centering
            \textbf{Prompt-to-Prompt w/o Blend}
        \end{minipage}
        \hfill
        \begin{minipage}{.24\linewidth}
            \centering
            \textbf{Prompt-to-Prompt w/ Blend}
        \end{minipage}
        \hfill
        \begin{minipage}{.24\linewidth}
            \centering
            \textbf{Prompt-to-Prompt + \model{}}
        \end{minipage}
    }

    \footnotesize

    \vspace{0.8em}

    \begin{minipage}{.24\linewidth}
        \centering
        \includegraphics[width=\textwidth]{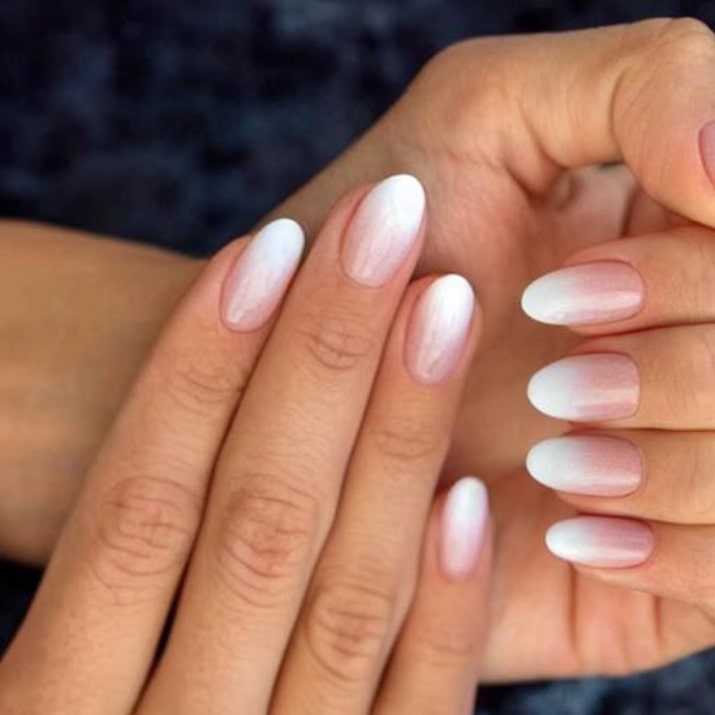}
    \end{minipage}
    \hfill
    \begin{minipage}{.24\linewidth}
        \centering
        \includegraphics[width=\textwidth]{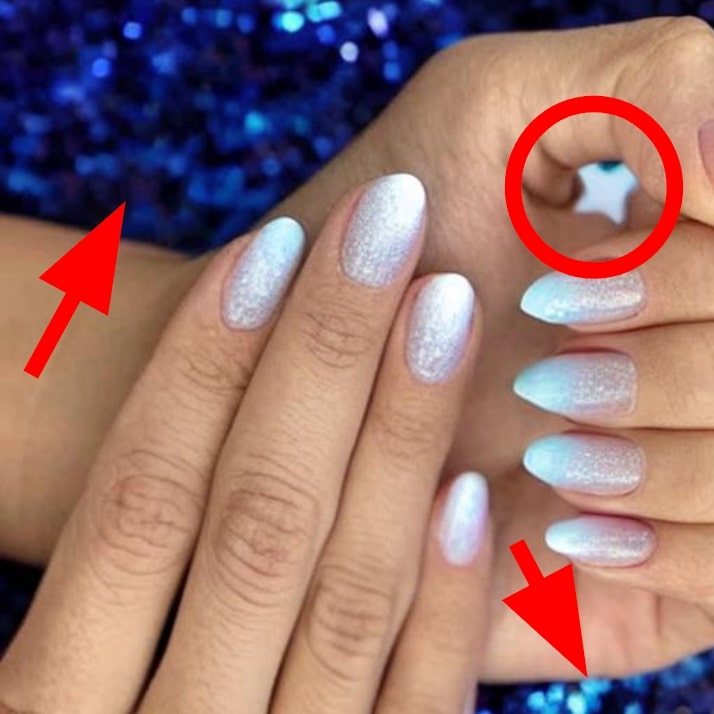}
    \end{minipage}
    \hfill
    \begin{minipage}{.24\linewidth}
        \centering
        \includegraphics[width=\textwidth]{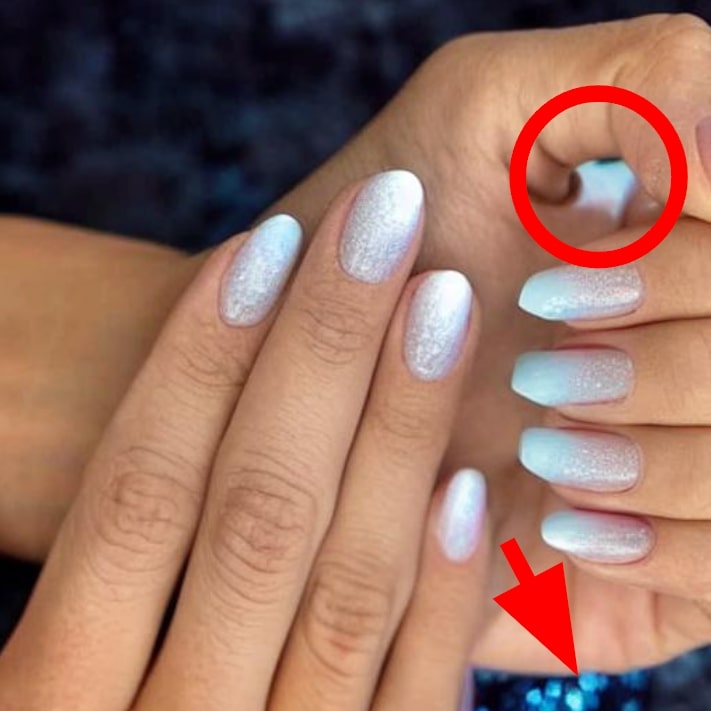}
    \end{minipage}
    \hfill
    \begin{minipage}{.24\linewidth}
        \centering
        \includegraphics[width=\textwidth]{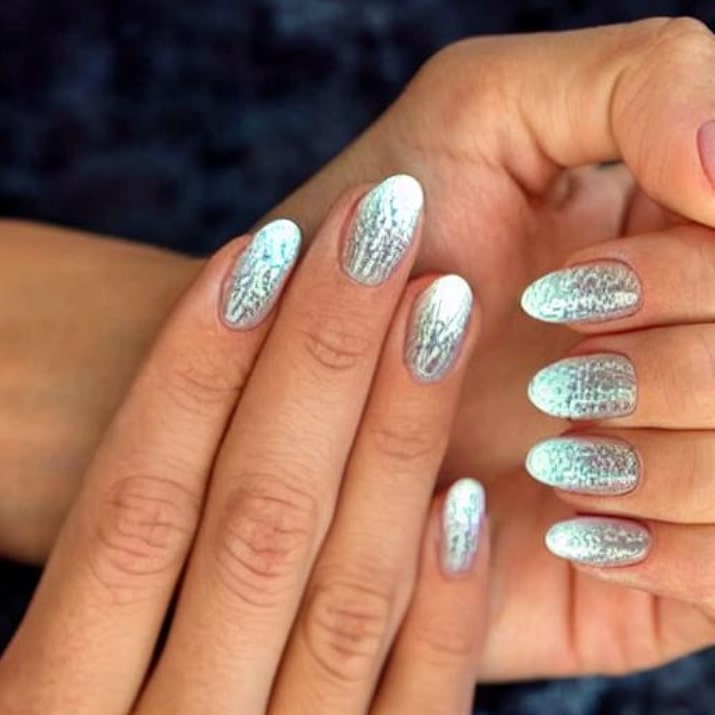}
    \end{minipage}

    \caption{\textbf{The integration of \model{} into Prompt-to-Prompt~\cite{Hertz2022Prompt2prompt}.} Red circles and arrows emphasize the localization issues of Prompt-to-Prompt model. The following textual information is used: \textbf{input caption:} \textit{A photo of fingernails} and \textbf{output caption:} \textit{A photo of glitter fingernails}.}
    \label{fig:p2p_integration}
    \vspace{-1.6em}

\end{figure}

%% file: tables/quantitative_MB_HIVE.tex
\begin{table*}[!tbh]
  \centering

  \caption{\textbf{HIVE + \model{} Evaluation on MagicBrush~\cite{Zhang2023MagicBrush}.} The numbers for others are sourced from \cite{Zhang2023MagicBrush}, while values for our method are computed by following the same protocol.The integration of {\textbf{{\model{}}}} surpasses the base model performance, \eg{,} HIVE and HIVE w/MB.}  %

  \resizebox{\linewidth}{!}{%
  \begin{tabular}{@{}lccccccccccc@{}}
\toprule
\multicolumn{1}{c}{\multirow{2}{*}{\textbf{Methods}}} & &\multicolumn{5}{c}{\textbf{Single-turn}}& \multicolumn{5}{c}{\textbf{Multi-turn}} \\
\cmidrule(lr){3-7} \cmidrule(lr){8-12}
 & \textbf{MB} & L1~$\downarrow$ & L2~ $\downarrow$ & CLIP-I~$\uparrow$ & DINO~$\uparrow$  &CLIP-T~$\uparrow$& L1~$\downarrow$ & L2~ $\downarrow$ & CLIP-I~$\uparrow$ & DINO~$\uparrow$ & CLIP-T~$\uparrow$\\
\midrule
IP2P~\cite{Brooks2022InstructPix2Pix} & \xmark & 0.112& 0.037& 0.852& 0.743&0.276& 0.158& 0.060& 0.792& 0.618& 0.273\\
IP2P~\cite{Brooks2022InstructPix2Pix} & \cmark & 0.063& {0.020}& 0.933& 0.899&0.278& 0.096& 0.035& {0.892}& {0.827}& 0.275\\
\midrule
HIVE~\cite{Zhang2023HIVE} & \xmark & 0.109& 0.034& 0.852& 0.750&0.275& 0.152& 0.056& 0.800& 0.646& 0.267\\
HIVE~\cite{Zhang2023HIVE} {\textbf{+ {\model{}}}} & \xmark & \textbf{0.051} & \textbf{0.016} & \textbf{0.940} & \textbf{0.909} & \underline{0.293} & \textbf{0.080} & \underline{0.029} & \underline{0.894} & \textbf{0.829} &\underline{ 0.283} \\ %
HIVE~\cite{Zhang2023HIVE} & \cmark & 0.066& \underline{0.022}& 0.919& 0.866&0.281& \underline{0.097}& 0.037& 0.879& \underline{0.789}& 0.280\\
HIVE~\cite{Zhang2023HIVE} {\textbf{+ {\model{}}}} & \cmark & \underline{0.053} & \textbf{0.016} & \underline{0.939} & \underline{0.906} & \textbf{0.300} & \textbf{0.080} & \textbf{0.028} & \textbf{0.899} & \textbf{0.829} & \textbf{0.295 }\\

\bottomrule
  \end{tabular}
   }  %

\label{tab:MB_GIVE} %
\end{table*}

%% file: tex_figures/hive_integration.tex
\begin{figure}[!htb]
    
    \centering

    \scriptsize{
        \begin{minipage}{.3\linewidth}
            \centering
            \textbf{Input Image}
        \end{minipage}
        \begin{minipage}{.3\linewidth}
            \centering
            \textbf{HIVE}
        \end{minipage}
        \begin{minipage}{.3\linewidth}
            \centering
            \textbf{HIVE + \model{}}
        \end{minipage}
    }

    \footnotesize

    \vspace{0.4em}

    \begin{minipage}{.3\linewidth}
        \centering
        \includegraphics[width=\textwidth]{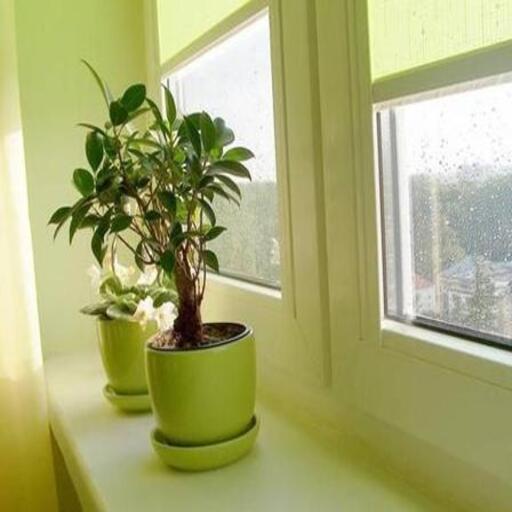}
    \end{minipage}
    \begin{minipage}{.3\linewidth}
        \centering
        \includegraphics[width=\textwidth]{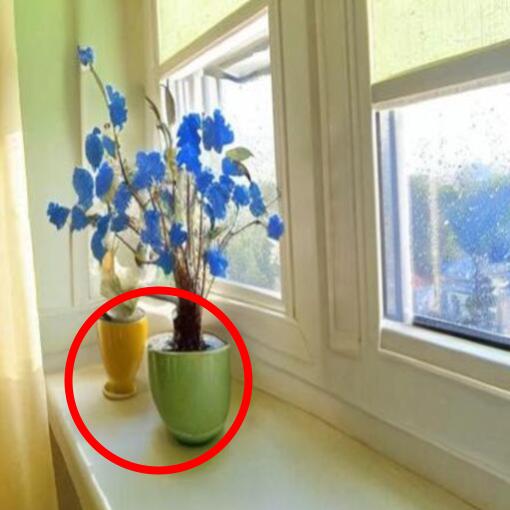}
    \end{minipage}
    \begin{minipage}{.3\linewidth}
        \centering
        \includegraphics[width=\textwidth]{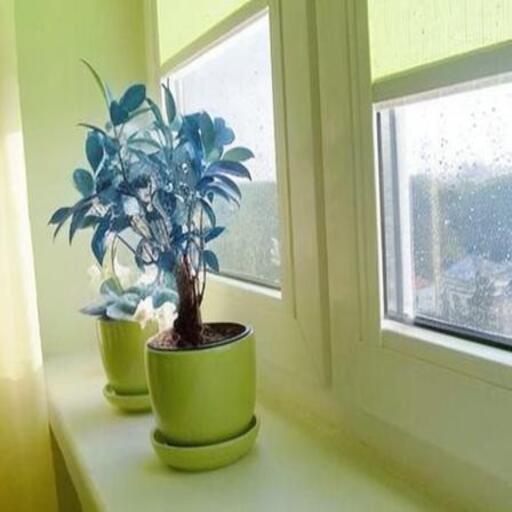}
    \end{minipage}

    \vspace{0.4em}

    \begin{minipage}{.3\linewidth}
        \centering
        (a)
    \end{minipage}
    \begin{minipage}{.60\linewidth}
        \centering
        \textit{Change the plants color to blue.}
    \end{minipage}

    \vspace{0.4em}

    \begin{minipage}{.3\linewidth}
        \centering
        \includegraphics[width=\textwidth]{figures/qualitative/ip2p/fingernails/input.jpg}
    \end{minipage}
    \begin{minipage}{.3\linewidth}
        \centering
        \includegraphics[width=\textwidth]{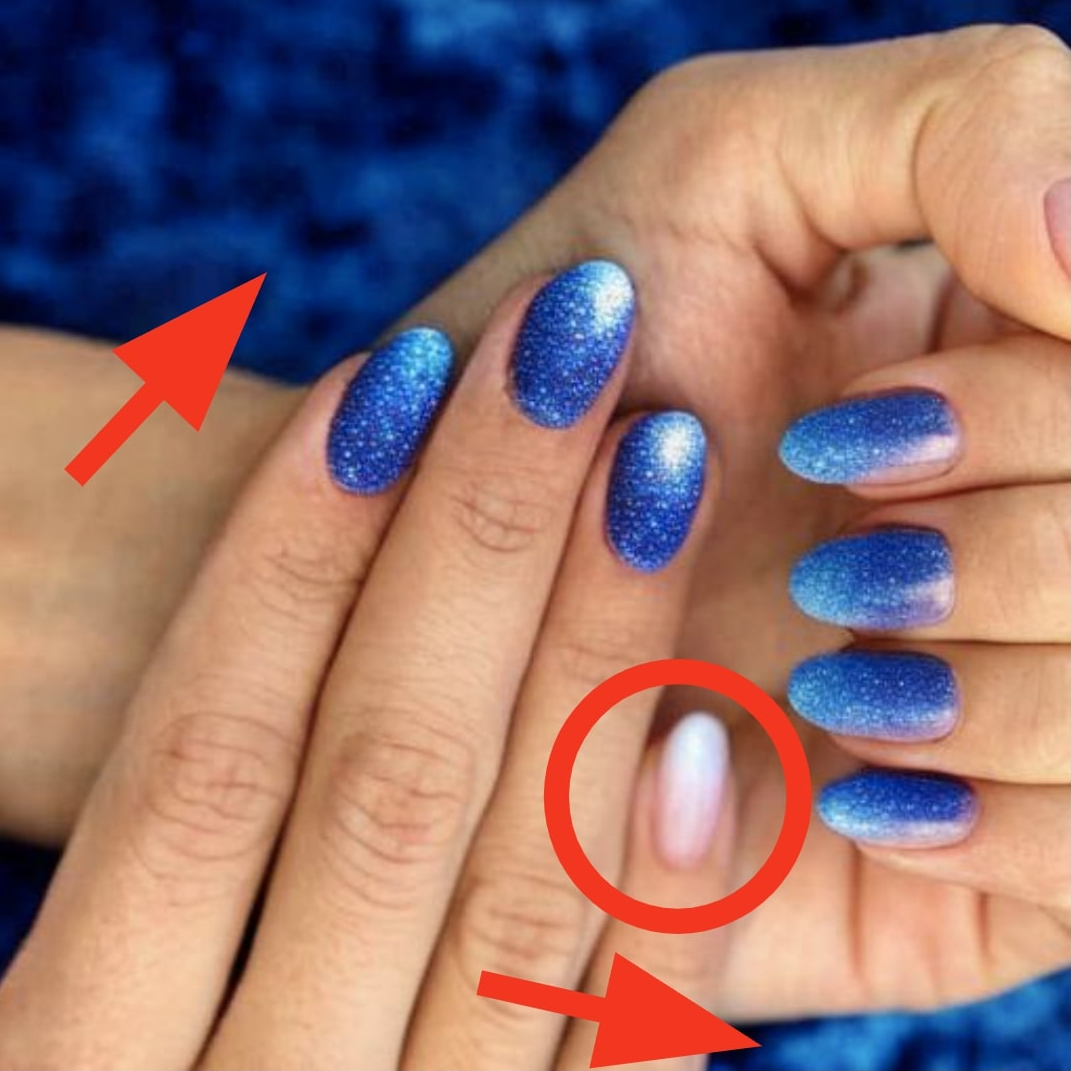}
    \end{minipage}
    \begin{minipage}{.3\linewidth}
        \centering
        \includegraphics[width=\textwidth]{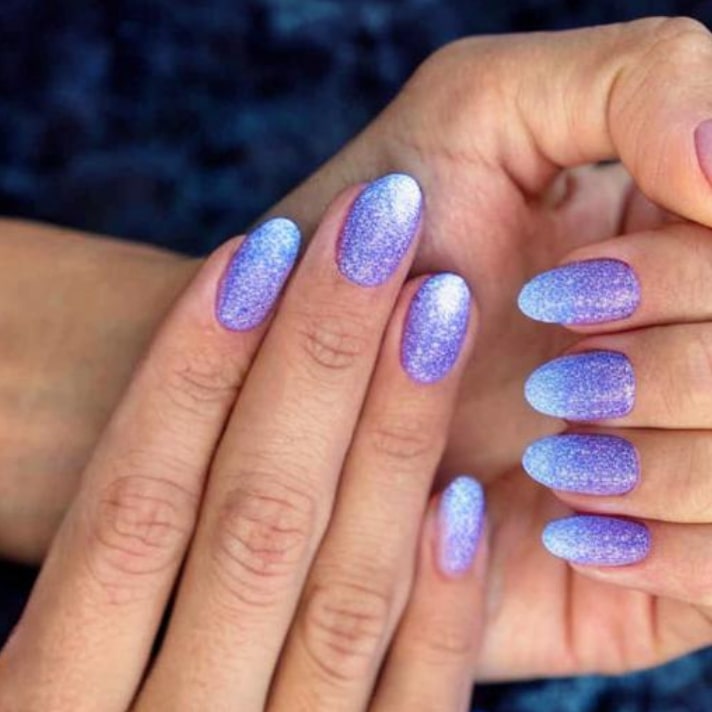}
    \end{minipage}

    \vspace{0.4em}

    \begin{minipage}{.3\linewidth}
        \centering
        (b)
    \end{minipage}
    \begin{minipage}{.60\linewidth}
        \centering
        \textit{Put blue glitter on fingernails.}
    \end{minipage}

    \vspace{0.4em}

    \begin{minipage}{.3\linewidth}
        \centering
        \includegraphics[width=\textwidth]{figures/qualitative/ip2p/outfit/input.jpg}
    \end{minipage}
    \begin{minipage}{.3\linewidth}
        \centering
        \includegraphics[width=\textwidth]{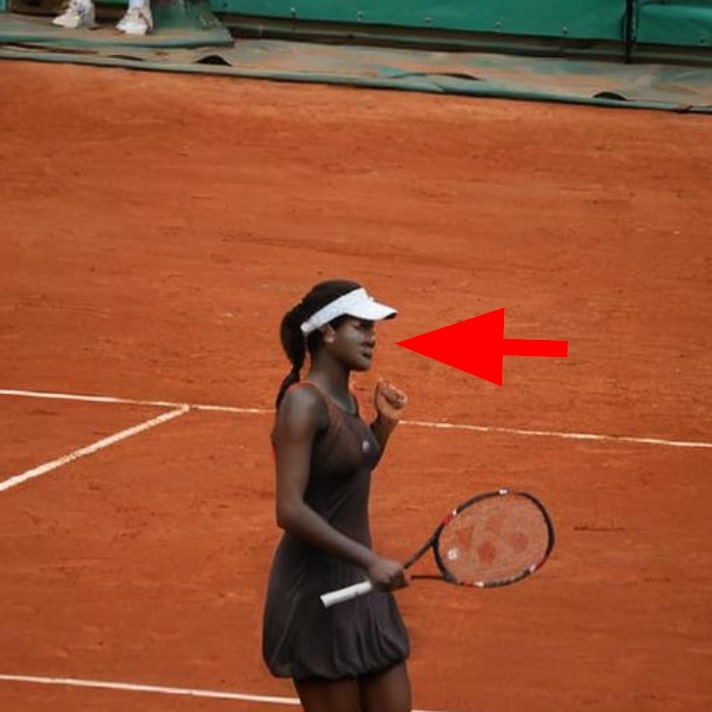}
    \end{minipage}
    \begin{minipage}{.3\linewidth}
        \centering
        \includegraphics[width=\textwidth]{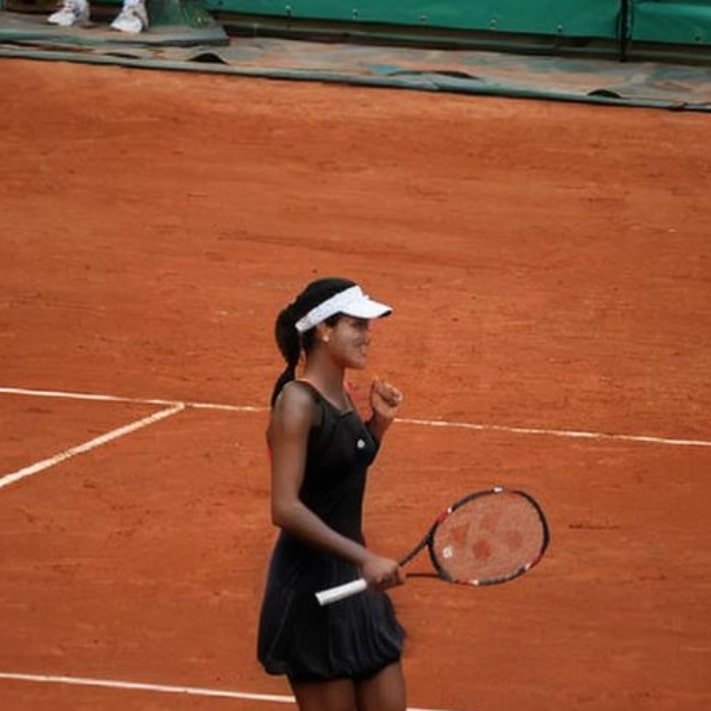}
    \end{minipage}

    \vspace{0.4em}

    \begin{minipage}{.3\linewidth}
        \centering
        (c)
    \end{minipage}
    \begin{minipage}{.60\linewidth}
        \centering
        \textit{Make her outfit black.}
    \end{minipage}

    \caption{\textbf{The integration of \model{} into HIVE~\cite{Zhang2023HIVE}.} Red circles and arrows emphasize the localization issues of HIVE model. HIVE + \model{} enables localized and effective edits.} \vspace{-2.8em}
    \label{fig:hive_integration}

\end{figure}

%% file: tables/quantitative_piebench.tex
\begin{table*}[!hbt]
    \centering
    \vspace{-0.8em}

    \caption{\textbf{Evaluation on PIE-Bench~\cite{ju2023direct}.} Comparison across ten edit types shows the integration of {\textbf{{\model{}}}} outperforming base models on instruction-based editing models. \emph{GT Mask} stands for ground-truth regions of interest masks.} %
    \vspace{-0.8em}
    \resizebox{0.75\linewidth}{!}{%
    \setlength{\tabcolsep}{0.5mm}{
    \begin{tabular}{l|c|cccc|cc}
    \toprule
    & \textbf{Structure} & \multicolumn{4}{c|}{\textbf{Background Preservation}} & \multicolumn{2}{c}{\textbf{CLIP Similarity}} \\ \midrule
    
    \textbf{Methods} & \textbf{Distance}$_{^{\times 10^3}}$ $\downarrow$ & \textbf{PSNR} $\uparrow$ & \textbf{LPIPS}$_{^{\times 10^3}}$ $\downarrow$ & \textbf{MSE}$_{^{\times 10^4}}$ $\downarrow$ & \textbf{SSIM}$_{^{\times 10^2}}$ $\uparrow$ & \textbf{Whole} $\uparrow$ & \textbf{Edited} $\uparrow$ \\ \midrule
    
    InstructDiffusion~\cite{geng2023instructdiffusion} & {75.44} & 20.28 & 155.66 & 349.66 & 75.53 & 23.26 & 21.34 \\
    DirectInversion + P2P~\cite{ju2023direct} & \underline{11.65} & {27.22} & {54.55} & \underline{32.86} & {84.76} & \textbf{25.02} & \textbf{22.10} \\

    IP2P~\cite{Brooks2022InstructPix2Pix} & 57.91 & 20.82 & 158.63 & 227.78 & 76.26 & 23.61 & {21.64} \\
    IP2P~\cite{Brooks2022InstructPix2Pix} {\textbf{+ {\model{}}}} & {32.80} & {21.36} & {110.69} & {159.93} & {80.20} & {23.73} & {21.11} \\

    IP2P w/MB~\cite{Zhang2023MagicBrush} & 22.25 & \underline{27.68} & \underline{47.61} & 40.03 & \underline{85.82} & {23.83} & {21.26} \\
    IP2P w/MB~\cite{Zhang2023MagicBrush} {\textbf{+ {\model{}}}} & \textbf{10.81} & \textbf{28.80} & \textbf{41.08} & \textbf{27.80} & \textbf{86.51} & 23.54 & 20.90 \\

    HIVE~\cite{Zhang2023HIVE} & 56.37 & 21.76 & 142.97 & 159.10 & 76.73 & 23.30 & {21.52} \\
    HIVE~\cite{Zhang2023HIVE} {\textbf{+ {\model{}}}} & {37.05} & {22.90} & {112.99} & {107.17} & {78.67} & {23.41} & 21.12 \\

    HIVE w/MB~\cite{Zhang2023MagicBrush} & 34.91 & 20.85 & 158.12 & 227.18 & 76.47 & 23.90 & \underline{21.75} \\
    HIVE w/MB~\cite{Zhang2023MagicBrush} {\textbf{+ {\model{}}}} & {26.98}& {26.09}& {68.28}& {63.70}& {84.58}& \underline{23.96} & 21.36 \\
    
    \bottomrule
    \end{tabular} }
    \vspace{-2em}
    }
    
    \label{tab:pie_bench} 
\end{table*}

%% file: tex_figures/magicbrush_annotations.tex
\begin{figure}[!htb]
    
    \footnotesize
    \centering
    
    \begin{minipage}{.19\linewidth}
        \centering
        \textbf{Input Image}
    \end{minipage}
    \hfill
    \begin{minipage}{.19\linewidth}
        \centering
        \textbf{GT~}
    \end{minipage}
    \hfill
    \begin{minipage}{.19\linewidth}
        \centering
        \textbf{+ Edit}
    \end{minipage}
    \hfill
    \begin{minipage}{.19\linewidth}
        \centering
        \textbf{RoI}
    \end{minipage}
    \hfill
    \begin{minipage}{.19\linewidth}
        \centering
        \textbf{+ Edit}
    \end{minipage}

    \vspace{0.6em}

    \begin{minipage}{.19\linewidth}
        \centering
        \includegraphics[width=\textwidth]{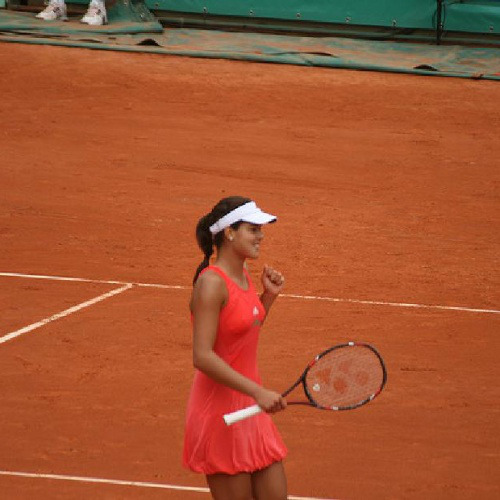}
    \end{minipage}
    \hfill
    \begin{minipage}{.19\linewidth}
        \centering
        \includegraphics[width=\textwidth]{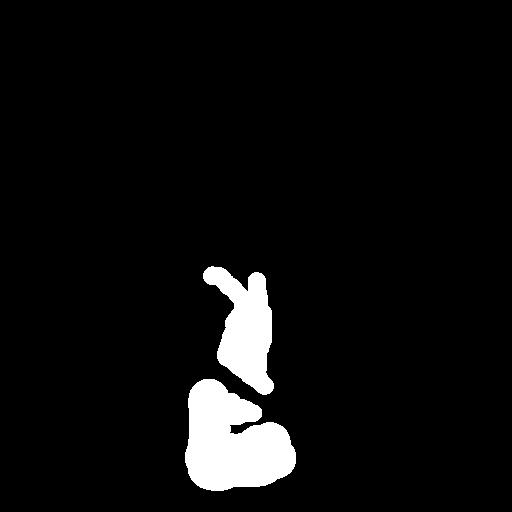}
    \end{minipage}
    \hfill
    \begin{minipage}{.19\linewidth}
        \centering
        \includegraphics[width=\textwidth]{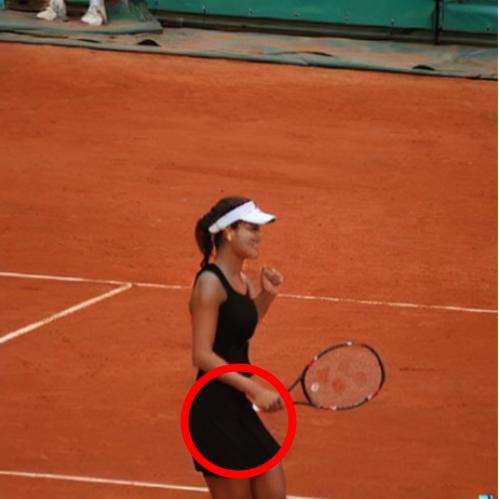}
    \end{minipage}
    \hfill
    \begin{minipage}{.19\linewidth}
        \centering
        \includegraphics[width=\textwidth]{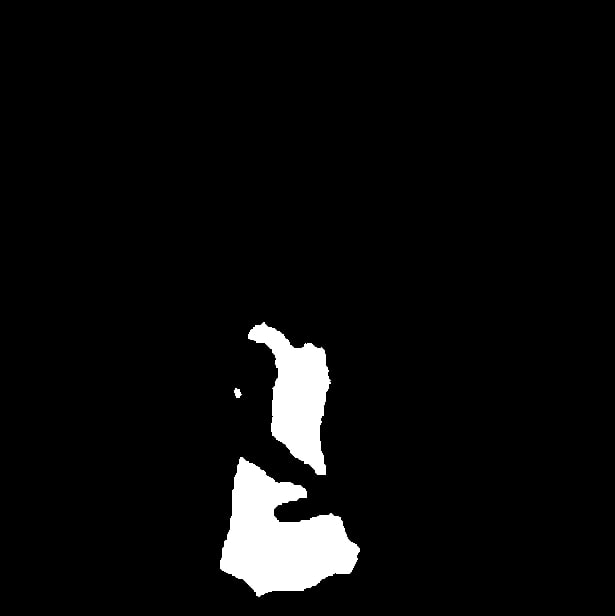}
    \end{minipage}
    \hfill
    \begin{minipage}{.19\linewidth}
        \centering
        \includegraphics[width=\textwidth]{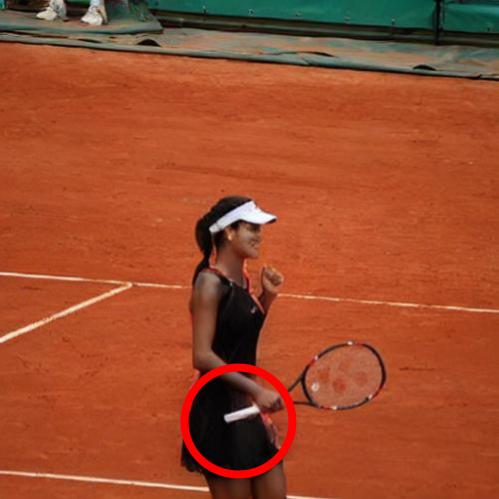}
    \end{minipage}

    \vspace{0.6em}

    \begin{minipage}{.19\linewidth}
        \centering
        (a)
    \end{minipage}
    \hfill
    \begin{minipage}{.76\linewidth}
        \centering
        \textit{Make her outfit black.}
    \end{minipage}
    
    \vspace{0.6em}

    \begin{minipage}{.19\linewidth}
        \centering
        \includegraphics[width=\textwidth]{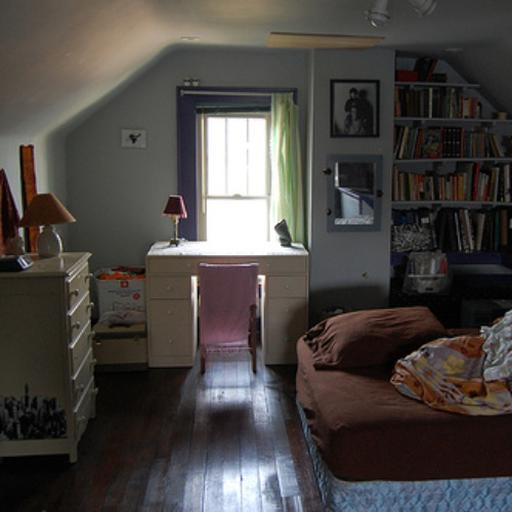}
    \end{minipage}
    \hfill
    \begin{minipage}{.19\linewidth}
        \centering
        \includegraphics[width=\textwidth]{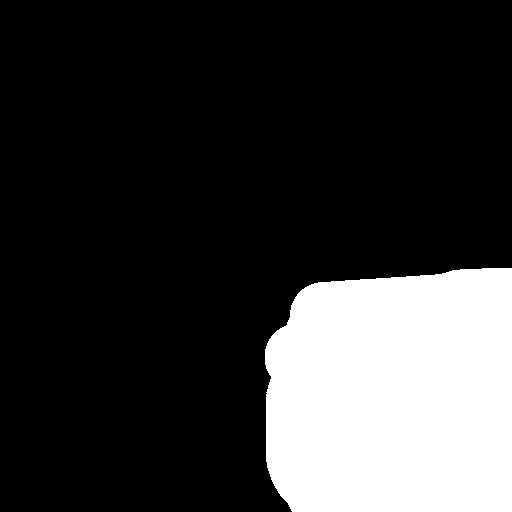}
    \end{minipage}
    \hfill
    \begin{minipage}{.19\linewidth}
        \centering
        \includegraphics[width=\textwidth]{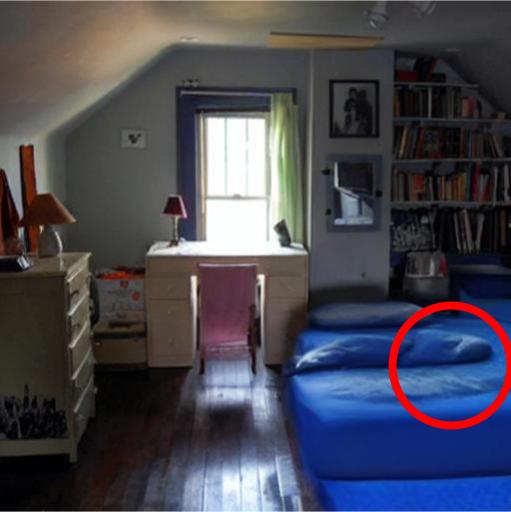}
    \end{minipage}
    \hfill
    \begin{minipage}{.19\linewidth}
        \centering
        \includegraphics[width=\textwidth]{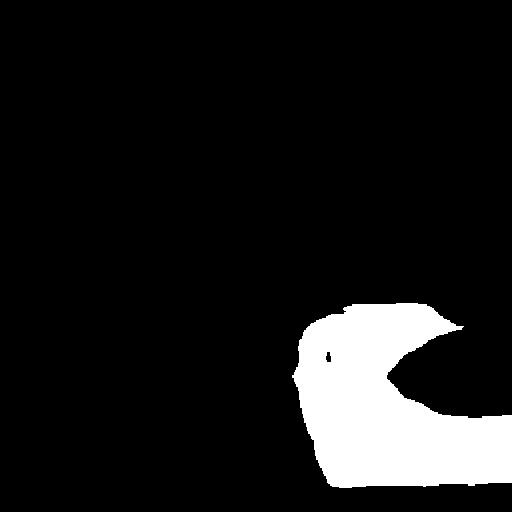}
    \end{minipage}
    \hfill
    \begin{minipage}{.19\linewidth}
        \centering
        \includegraphics[width=\textwidth]{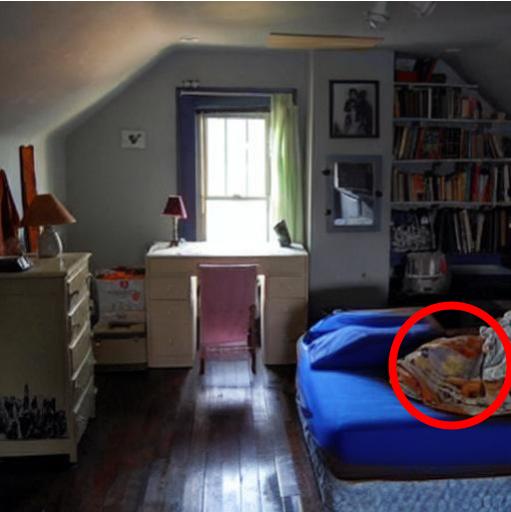}
    \end{minipage}

    \vspace{0.6em}

    \begin{minipage}{.19\linewidth}
        \centering
        (b)
    \end{minipage}
    \hfill
    \begin{minipage}{.76\linewidth}
        \centering
        \textit{Can the bed be blue?}
    \end{minipage}

    \caption{\textbf{MagicBrush Mask Annotations.} Ground truth~(GT) refers to mask annotations in MagicBrush~\cite{Zhang2023MagicBrush}. RoI indicates inferred masks from our proposed method. Red circles on the edited images~(+ Edit) highlight area where the precise localization of the edits can be appreciated.} \vspace{-2em}
    \label{fig:magicbrush_anno}
\end{figure}

%% file: tables/quantitative_editval.tex
  \centering
  \caption{\textbf{Evaluation on EditVal~\cite{basu2023editval}.} Comparison across six edit types shows our method outperforming eight state-of-the-art text-guided image editing models. The numbers for other methods are directly taken from the benchmark dataset~\cite{basu2023editval}. %
  } %

  \resizebox{\linewidth}{!}{%
  \begin{tabular}{@{}lccccccc@{}}
    \toprule
    \textbf{Method} & \textbf{O.A.} & \textbf{O.R.} & \textbf{P.R.} & \textbf{P.A.} & \textbf{S.} & \textbf{A.P.} & \textbf{Avg.} \\
    \midrule
    SINE~\cite{zhang2022sine}        & \underline{0.47} & \textbf{0.59} & 0.02 & 0.16 & 0.46 & 0.30 & \underline{0.33} \\
    NText.~\cite{Mokady2022NullTextInversion}      & 0.35 & 0.48 & 0.00 & 0.20 & \textbf{0.52} & \textbf{0.34} & 0.32 \\
    IP2P~\cite{Brooks2022InstructPix2Pix}        & 0.38 & 0.39 & 0.07 & \underline{0.25} & \underline{0.51} & 0.25 & 0.31 \\
    Imagic~\cite{Kawar2022Imagic}      & 0.36 & \underline{0.49} & 0.03 & 0.08 & 0.49 & 0.21 & 0.28 \\
    SDEdit~\cite{Meng2022SDEdit}      & 0.35 & 0.06 & 0.04 & 0.18 & 0.47 & \underline{0.33} & 0.24 \\
    DBooth~\cite{Ruiz2022DreamBooth}      & 0.39 & 0.32 & \underline{0.11} & 0.08 & 0.28 & 0.22 & 0.24 \\
    TInv.~\cite{gal2022image}       & 0.43 & 0.19 & 0.00 & 0.00 & 0.00 & 0.21 & 0.14 \\
    DiffEdit~\cite{couairon2023diffedit}    & 0.34 & 0.26 & 0.00 & 0.00 & 0.00 & 0.07 & 0.11 \\
    \midrule
    IP2P~\cite{Brooks2022InstructPix2Pix} {\textbf{+ {\model{}}}} & \textbf{0.48} & \underline{0.49} & \textbf{0.21} & \textbf{0.34} & 0.49 & 0.28 & \textbf{0.38} \\
    \bottomrule
  \end{tabular}
  }
  \label{tab:editval}

%% file: tex_figures/vqgan.tex
\begin{figure}[!ht]
    
    \centering
\vspace{-0.6em}
    \scriptsize{
        \begin{minipage}{.24\linewidth}
            \centering
            \textbf{Input Image}
        \end{minipage}
        \hfill
        \begin{minipage}{.24\linewidth}
            \centering
            \textbf{Ground Truth}
        \end{minipage}
        \hfill
        \begin{minipage}{.24\linewidth}
            \centering
            \textbf{VQGAN-CLIP~\cite{Crowson2022VQGAN-CLIP}}
        \end{minipage}
        \hfill
        \begin{minipage}{.24\linewidth}
            \centering
            \textbf{Ours}
        \end{minipage}
    }

    \footnotesize

    \vspace{0.4em}

    \begin{minipage}{.24\linewidth}
        \centering
        \includegraphics[width=\textwidth]{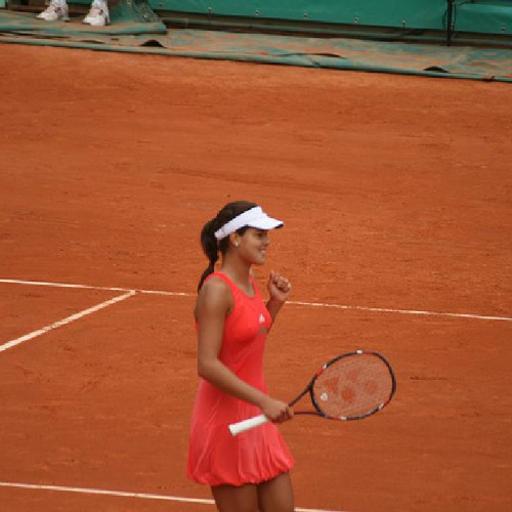}
    \end{minipage}
    \hfill
    \begin{minipage}{.24\linewidth}
        \centering
        \includegraphics[width=\textwidth]{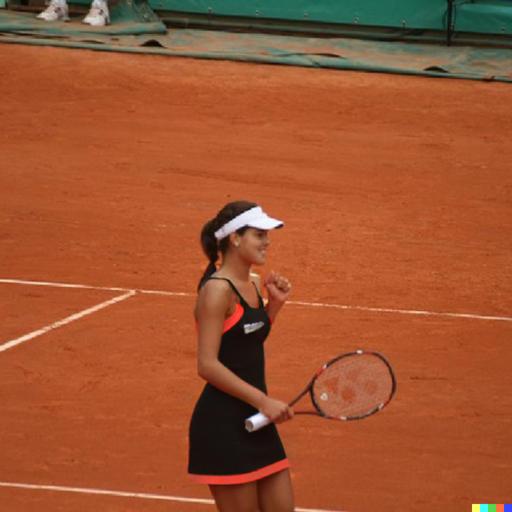}
    \end{minipage}
    \hfill
    \begin{minipage}{.24\linewidth}
        \centering
        \includegraphics[width=\textwidth]{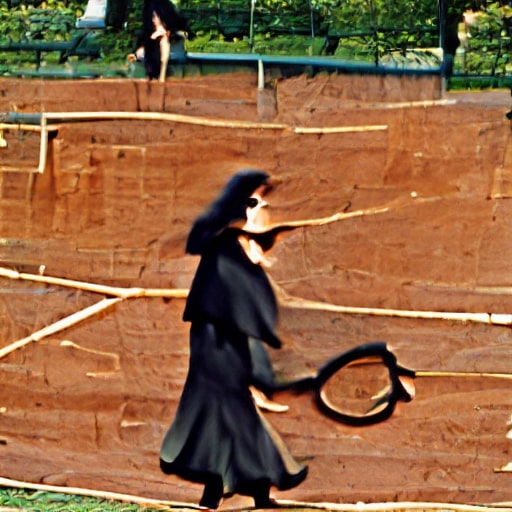}
    \end{minipage}
    \hfill
    \begin{minipage}{.24\linewidth}
        \centering
        \includegraphics[width=\textwidth]{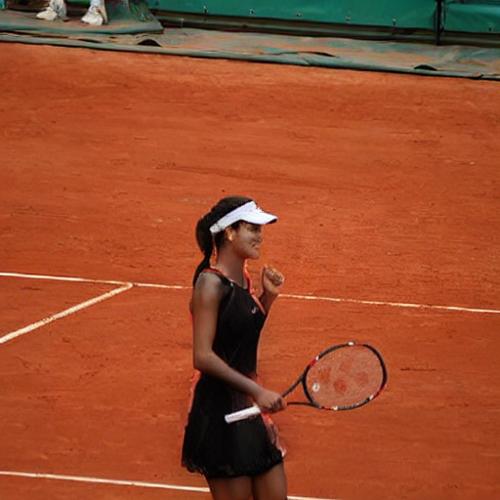}
    \end{minipage}

    \vspace{0.4em}

    \begin{minipage}{.24\linewidth}
        \centering
        \textit{Make her outfit black.}
    \end{minipage}
    \hfill
    \begin{minipage}{.24\linewidth}
        \centering
        {CLIP-T:} \textit{0.306}
    \end{minipage}
    \hfill
    \begin{minipage}{.24\linewidth}
        \centering
        {CLIP-T:} \textit{0.486}
    \end{minipage}
    \hfill
    \begin{minipage}{.24\linewidth}
        \centering
        {CLIP-T:} \textit{0.314}
    \end{minipage}

    \vspace{0.4em}

    \begin{minipage}{.24\linewidth}
        \centering
        \includegraphics[width=\textwidth]{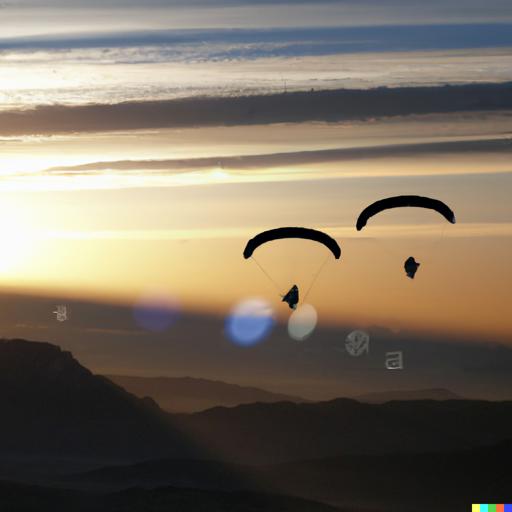}
    \end{minipage}
    \hfill
    \begin{minipage}{.24\linewidth}
        \centering
        \includegraphics[width=\textwidth]{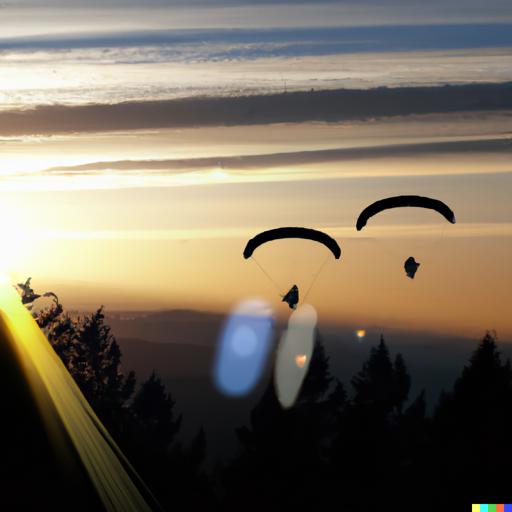}
    \end{minipage}
    \hfill
    \begin{minipage}{.24\linewidth}
        \centering
        \includegraphics[width=\textwidth]{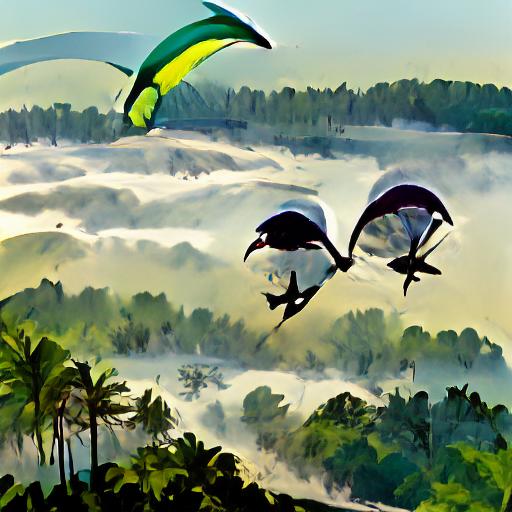}
    \end{minipage}
    \hfill
    \begin{minipage}{.24\linewidth}
        \centering
        \includegraphics[width=\textwidth]{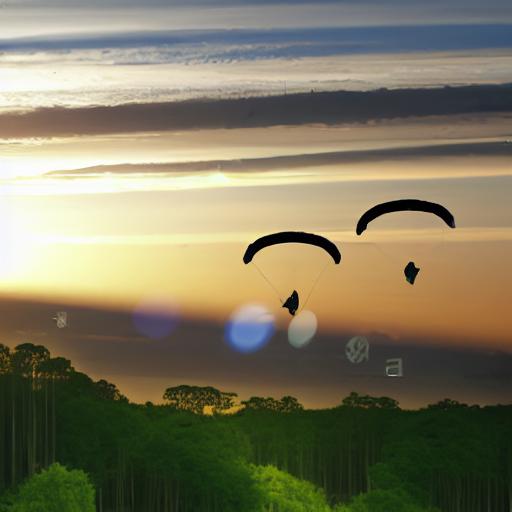}
    \end{minipage}

    \vspace{0.4em}

    \begin{minipage}{.24\linewidth}
        \centering
        \textit{Replace the ground with a forest.}
    \end{minipage}
    \hfill
    \begin{minipage}{.24\linewidth}
        \centering
        {CLIP-T:} \textit{0.311}
    \end{minipage}
    \hfill    
    \begin{minipage}{.24\linewidth}
        \centering
        {CLIP-T:} \textit{0.420}
    \end{minipage}
    \hfill
    \begin{minipage}{.24\linewidth}
        \centering
        {CLIP-T:} \textit{0.318}
    \end{minipage}
    \vspace{-0.8em}
    \caption{\textbf{Investigating CLIP-T for VQGAN-CLIP~\cite{Crowson2022VQGAN-CLIP}.} CLIP-T metrics are reported below each image and calculated between the output caption and the corresponding image. Input images and edit instructions are pictured in the first column. Ground truth edit images are taken from the MagicBrush dataset.}
    \label{fig:vqgan}
    \vspace{-2.5em}
\end{figure}

%% file: tex_figures/disentanglement.tex
\begin{figure}[!ht]
    
    \centering

    \small{
        \begin{minipage}{.32\linewidth}
            \centering
            \textbf{Input Image}
        \end{minipage}
        \hfill
        \begin{minipage}{.32\linewidth}
            \centering
            \textbf{DiffusionDisent.~\cite{wu2023uncovering}}
        \end{minipage}
        \hfill
        \begin{minipage}{.32\linewidth}
            \centering
            \textbf{Ours}
        \end{minipage}
    }

    \footnotesize

    \vspace{0.4em}

    \begin{minipage}{.32\linewidth}
        \centering
        \includegraphics[width=\textwidth]{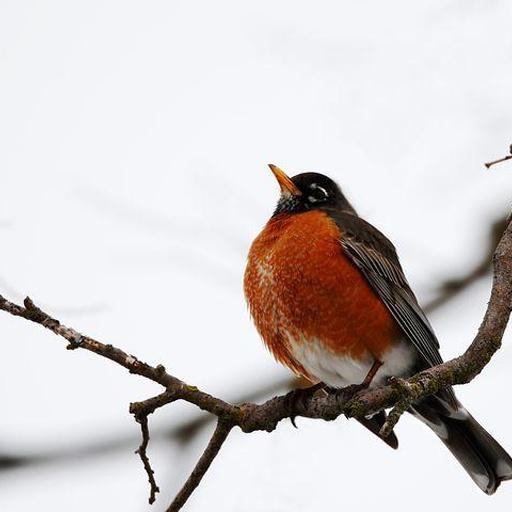}
    \end{minipage}
    \begin{minipage}{.32\linewidth}
        \centering
        \includegraphics[width=\textwidth]{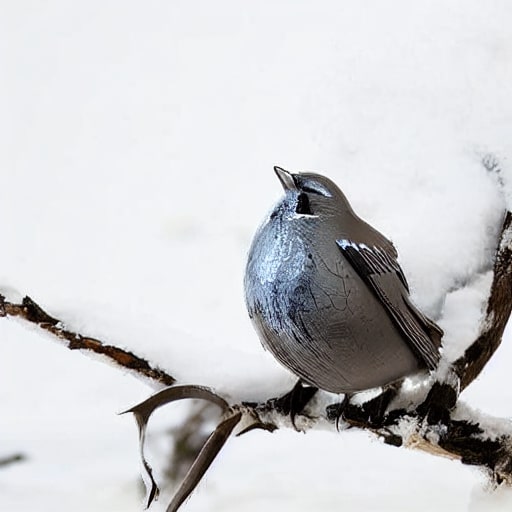}
    \end{minipage}
    \begin{minipage}{.32\linewidth}
        \centering
        \includegraphics[width=\textwidth]{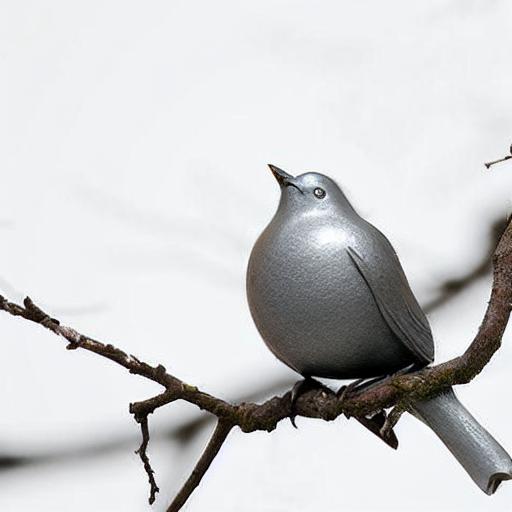}
    \end{minipage}

    \vspace{0.4em}

    \begin{minipage}{.32\linewidth}
        \centering
        {(a)}
    \end{minipage}
    \begin{minipage}{.64\linewidth}
        \centering
        \textit{Change the robin to a silver robin sculpture.}
    \end{minipage}

    \vspace{0.4em}

    \begin{minipage}{.32\linewidth}
        \centering
        \includegraphics[width=\textwidth]{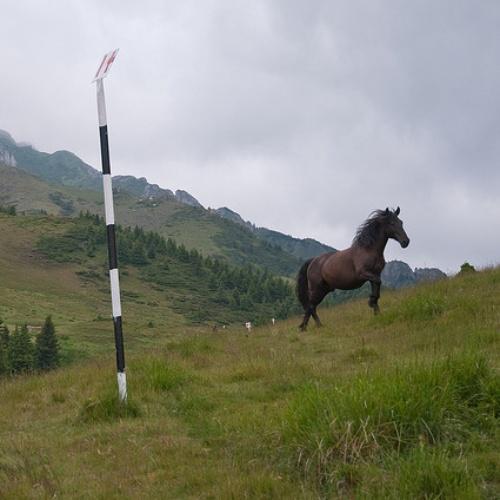}
    \end{minipage}
    \begin{minipage}{.32\linewidth}
        \centering
        \includegraphics[width=\textwidth]{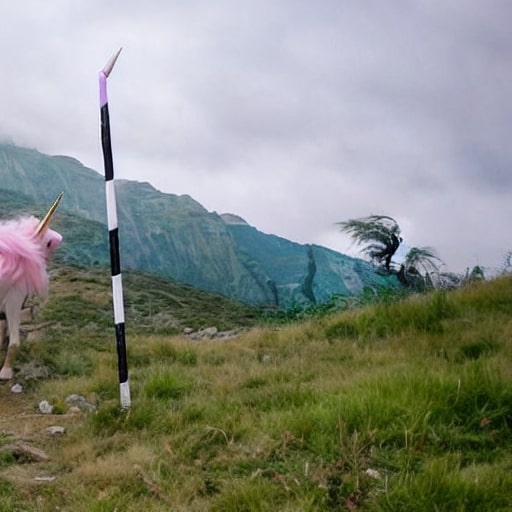}
    \end{minipage}
    \begin{minipage}{.32\linewidth}
        \centering
        \includegraphics[width=\textwidth]{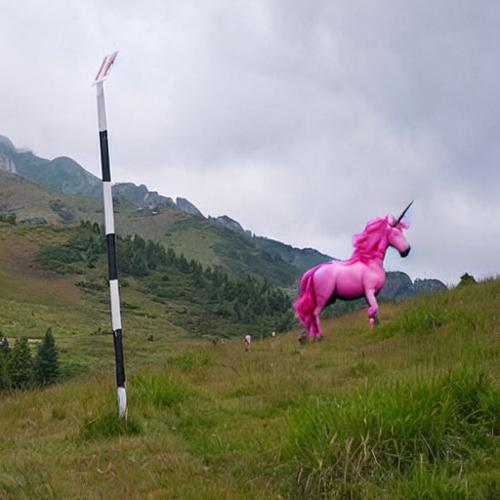}
    \end{minipage}

    \vspace{0.4em}

    \begin{minipage}{.32\linewidth}
        \centering
        {(b)}
    \end{minipage}
    \begin{minipage}{.64\linewidth}
        \centering
        \textit{Turn the brown horse into a pink unicorn.}
    \end{minipage}
    \vspace{-0.8em}

    \caption{\textbf{Diffusion Disentanglement~\cite{wu2023uncovering} Qualitative Comparison.} Edits are obtained by using the global description of the input image and the desired edit by concatenating them with ','.} 
        \vspace{-1.8em}
\label{fig:disentanglement}
\end{figure}

%% file: tex_figures/blendeddiff.tex
\begin{figure}[!ht]
    
    \centering

    \scriptsize{
        \begin{minipage}{.3\linewidth}
            \centering
            \textbf{Input Image}
        \end{minipage}
        \begin{minipage}{.3\linewidth}
            \centering
            \textbf{BlendedDiffusion~\cite{avrahami2023blended}}
        \end{minipage}
        \begin{minipage}{.3\linewidth}
            \centering
            \textbf{Ours}
        \end{minipage}
    }

    \footnotesize

    \vspace{0.4em}

    \begin{minipage}{.3\linewidth}
        \centering
        \includegraphics[width=\textwidth]{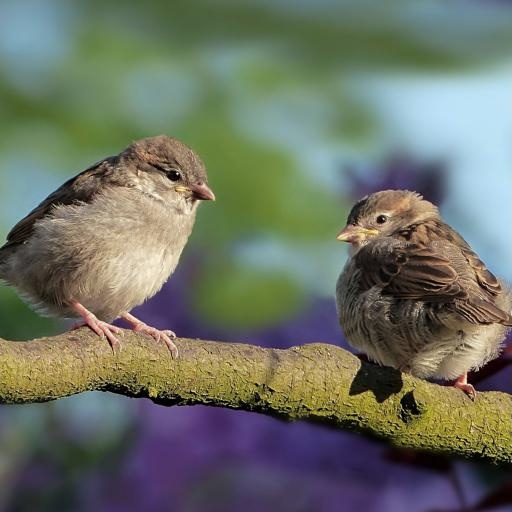}
    \end{minipage}
    \begin{minipage}{.3\linewidth}
        \centering
        \includegraphics[width=\textwidth]{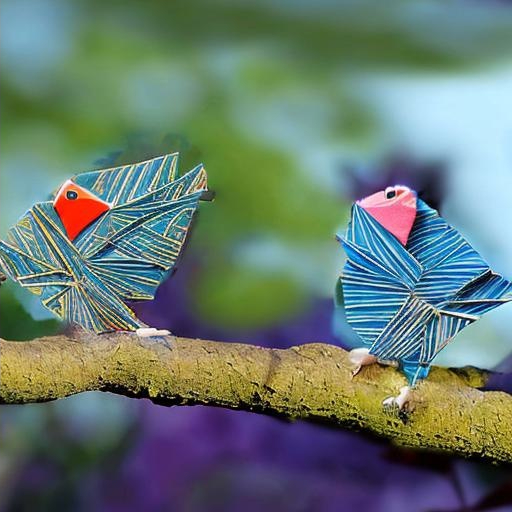}
    \end{minipage}
    \begin{minipage}{.3\linewidth}
        \centering
        \includegraphics[width=\textwidth]{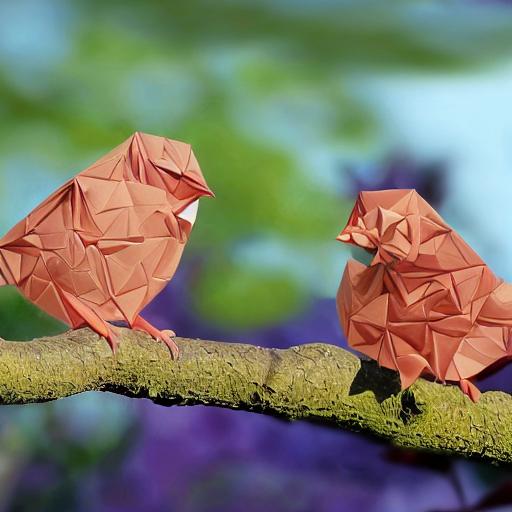}
    \end{minipage}

    \vspace{0.4em}

    \begin{minipage}{.3\linewidth}
        \centering
        {(a)}
    \end{minipage}
    \begin{minipage}{.60\linewidth}
        \centering
        \textit{Turn the real birds into origami birds.}
    \end{minipage}

    \vspace{0.4em}

    \begin{minipage}{.3\linewidth}
        \centering
        \includegraphics[width=\textwidth]{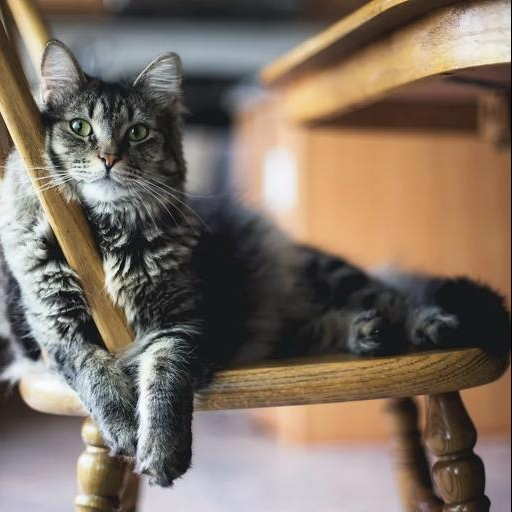}
    \end{minipage}
    \begin{minipage}{.3\linewidth}
        \centering
        \includegraphics[width=\textwidth]{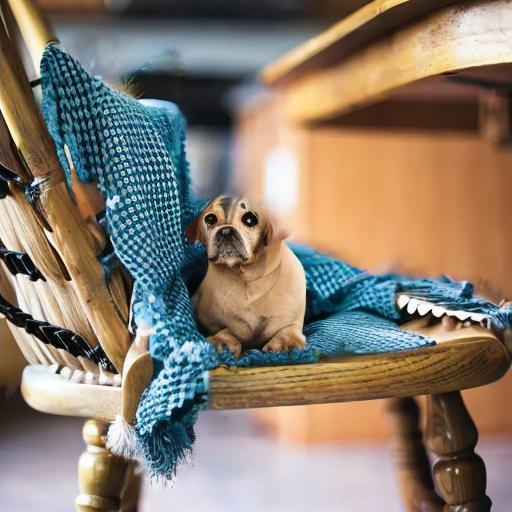}
    \end{minipage}
    \begin{minipage}{.3\linewidth}
        \centering
        \includegraphics[width=\textwidth]{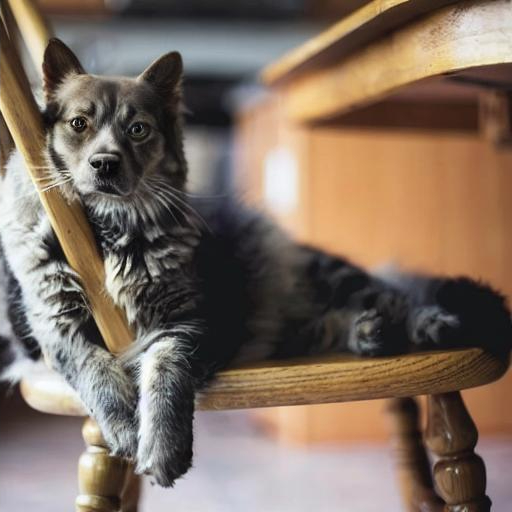}
    \end{minipage}

    \vspace{0.4em}

    \begin{minipage}{.3\linewidth}
        \centering
        {(b)}
    \end{minipage}
    \begin{minipage}{.60\linewidth}
        \centering
        \textit{Change the animal from a cat to a dog.}
    \end{minipage}

    \vspace{-0.8em}
    \caption{\textbf{BlendedDiffusion~\cite{avrahami2023blended} Qualitative Comparison.} Edited images based on input images and edit instructions reported below each row. The images for BlendedDiffusion are taken from the PIE-Bench evaluation~\cite{ju2023direct}.}
    \label{fig:blended}

\end{figure}

%% file: tex_figures/segmentation.tex
\begin{figure}[!ht]
    
    \centering

    \footnotesize{
        \begin{minipage}{.19\linewidth}
            \centering
            \textbf{Input Image}
        \end{minipage}
        \hfill
        \begin{minipage}{.19\linewidth}
            \centering
            \textbf{LPM~\cite{patashnik2023localizing}}
        \end{minipage}
        \hfill
        \begin{minipage}{.19\linewidth}
            \centering
            \textbf{SAM~\cite{Kirillov2023SAM}}
        \end{minipage}
        \hfill
        \begin{minipage}{.19\linewidth}
            \centering
            \textbf{ODISE~\cite{xu2023odise}}
        \end{minipage}
        \hfill
        \begin{minipage}{.19\linewidth}
            \centering
            \textbf{Ours}
        \end{minipage}
    }

    \footnotesize

    \vspace{0.4em}

    \begin{minipage}{.19\linewidth}
        \centering
        \includegraphics[width=\textwidth]{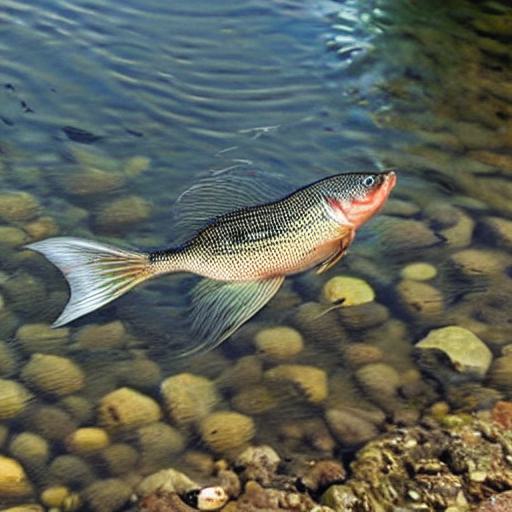}
    \end{minipage}
    \hfill
    \begin{minipage}{.19\linewidth}
        \centering
        \includegraphics[width=\textwidth]{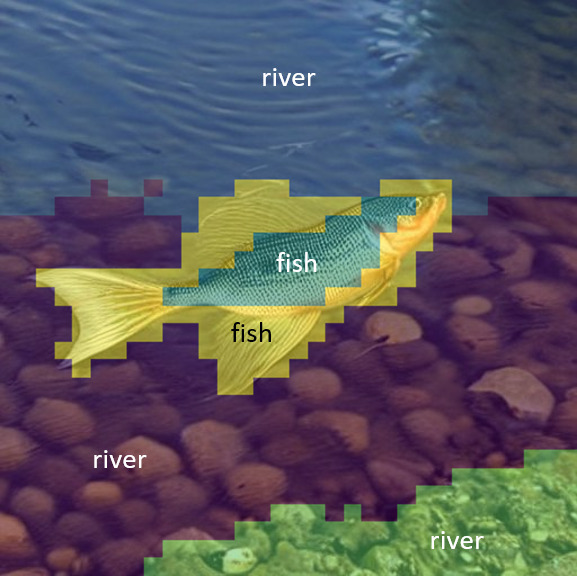}
    \end{minipage}
    \hfill
    \begin{minipage}{.19\linewidth}
        \centering
        \includegraphics[width=\textwidth]{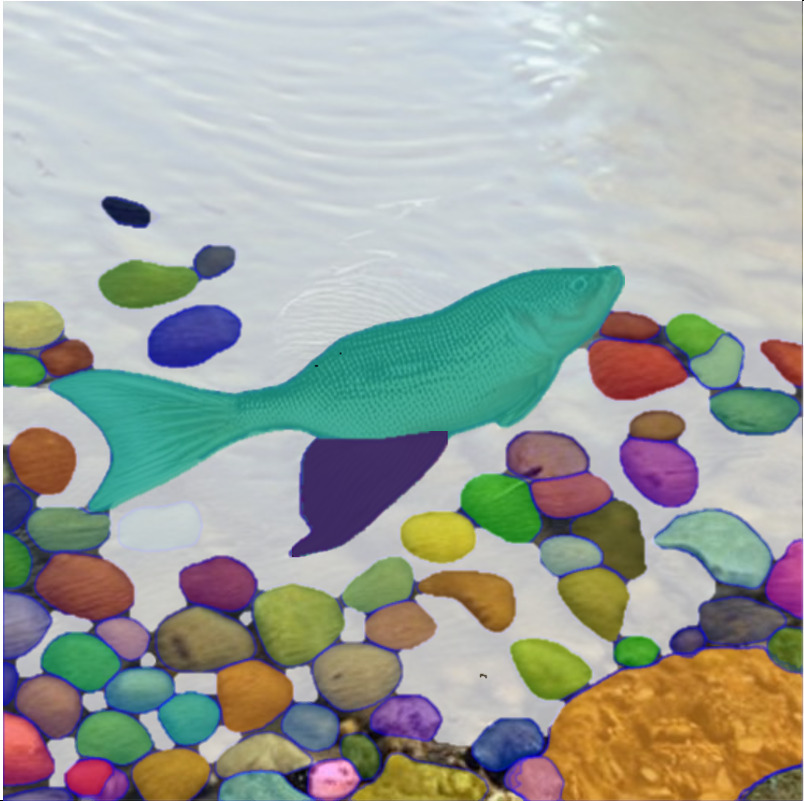}
    \end{minipage}
    \hfill
    \begin{minipage}{.19\linewidth}
        \centering
        \includegraphics[width=\textwidth]{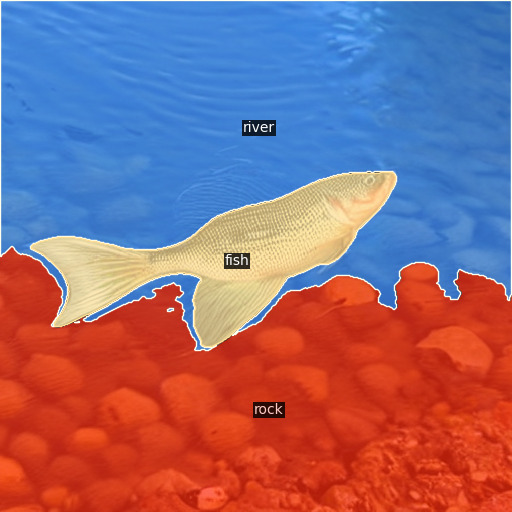}
    \end{minipage}
    \hfill
    \begin{minipage}{.19\linewidth}
        \centering
        \includegraphics[width=\textwidth]{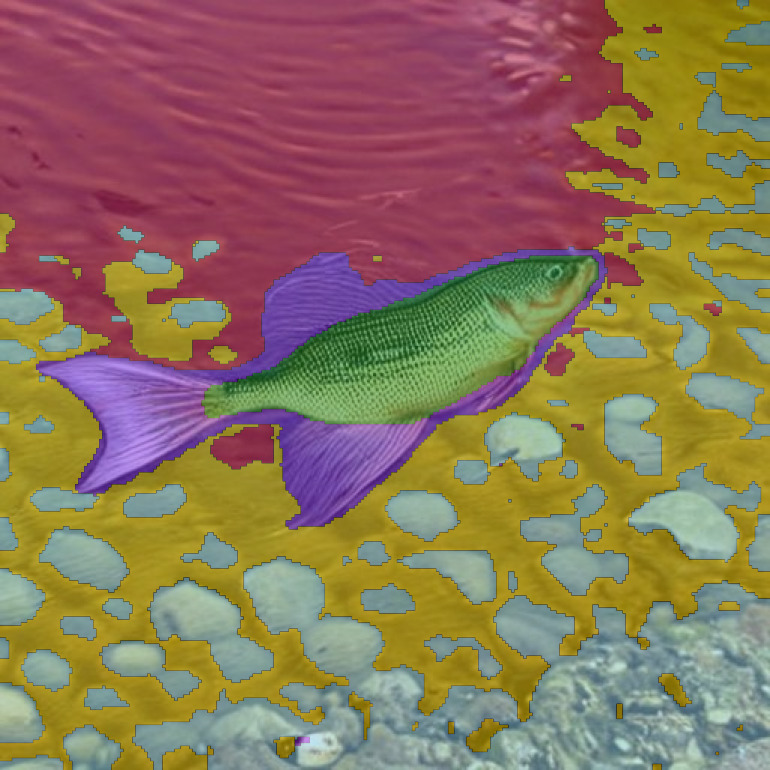}
    \end{minipage}

    \vspace{0.4em}
    
    \begin{minipage}{.19\linewidth}
        \centering
        \includegraphics[width=\textwidth]{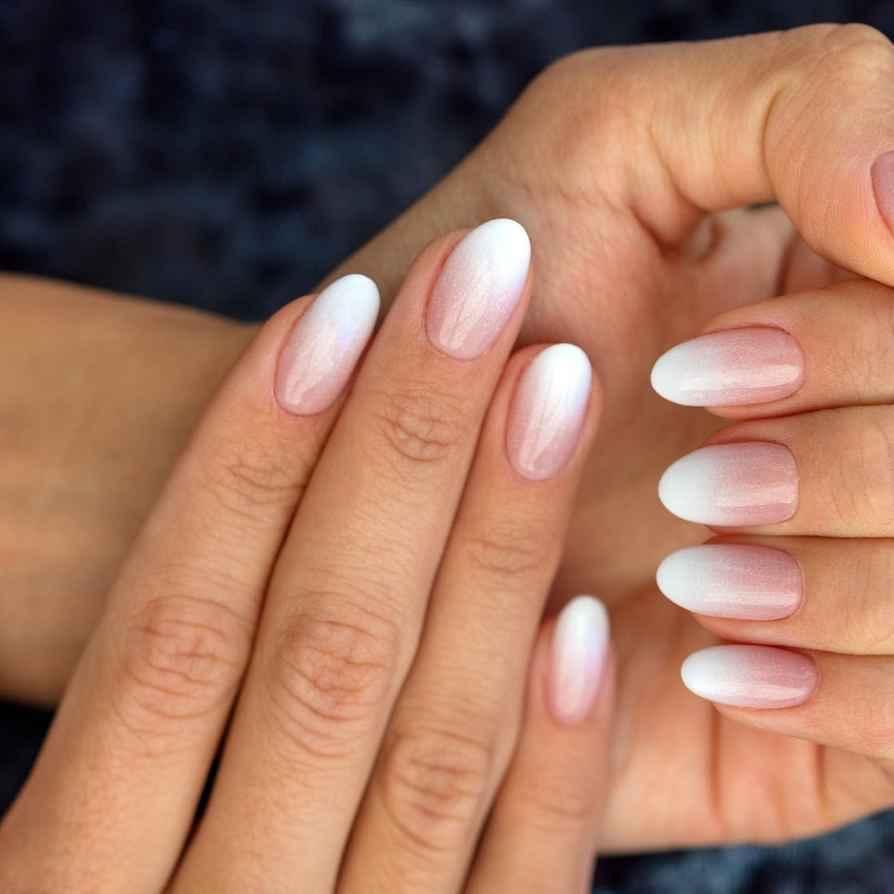}
    \end{minipage}
    \hfill
    \begin{minipage}{.19\linewidth}
        \centering
        \includegraphics[width=\textwidth]{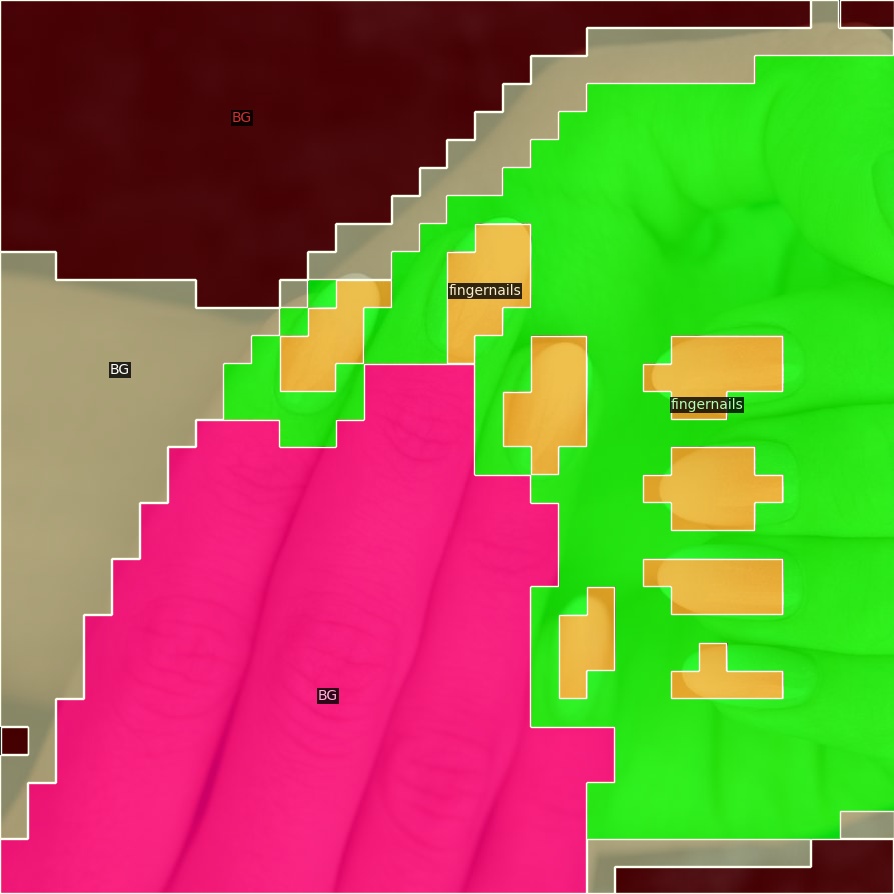}
    \end{minipage}
    \hfill
    \begin{minipage}{.19\linewidth}
        \centering
        \includegraphics[width=\textwidth]{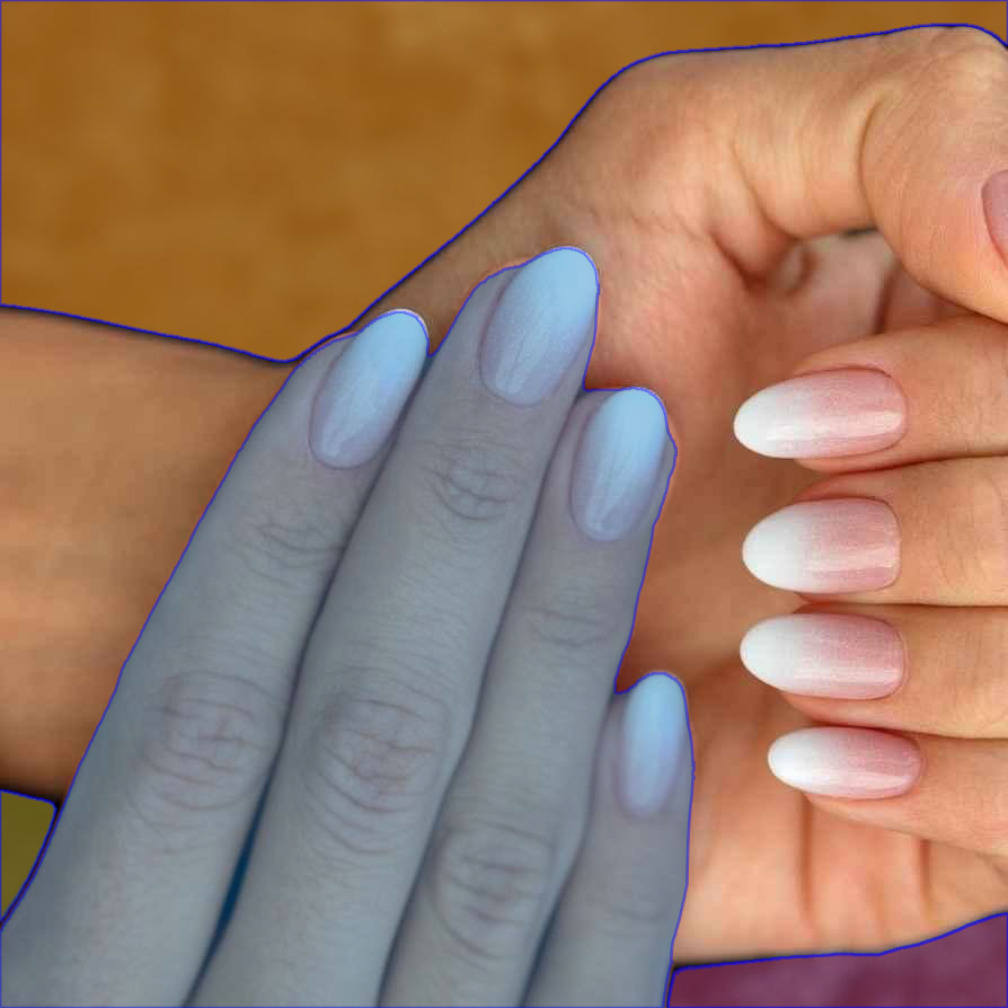}
    \end{minipage}
    \hfill
    \begin{minipage}{.19\linewidth}
        \centering
        \includegraphics[width=\textwidth]{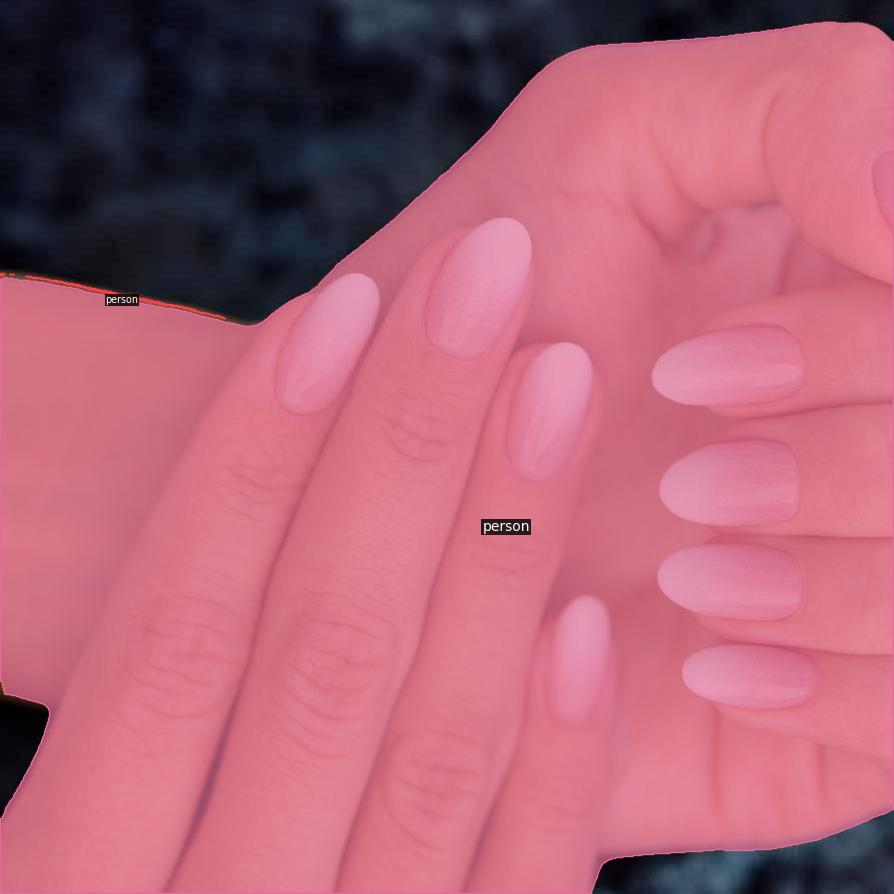}
    \end{minipage}
    \hfill
    \begin{minipage}{.19\linewidth}
        \centering
        \includegraphics[width=\textwidth]{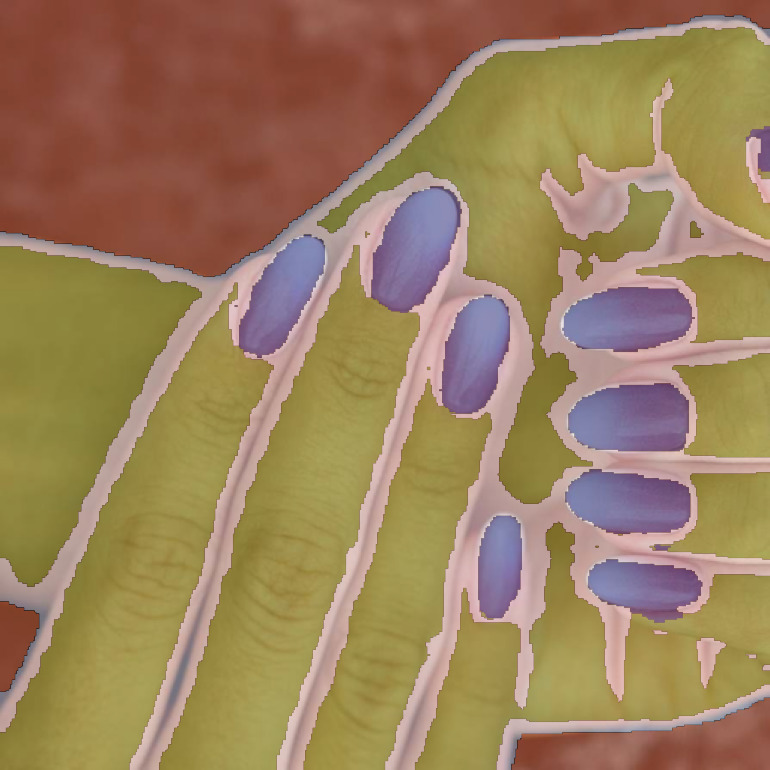}
    \end{minipage}

    \caption{\textbf{Segmentation Qualitative.} Comparison between the-state-of-art segmentation methods on challenging examples. In the first example, both LPM and \model{} can detect the fish fin, while SAM and ODISE cannot. Moreover, \model{} provides a higher resolution than LPM. In the second example, SAM and ODISE fail to identify fingernails. Although LPM can find them, it delivers low-resolution and imprecise segments. Our method, however, accurately identifies fingernails and offers precise boundaries.} %
    \label{fig:segmentation} 
\end{figure}

%% file: tables/ablation_ovs.tex
\begin{table}[!htb]

  \centering
  \caption{\textbf{OVS methods for localizing the edit.} We replace the edit localization part with OVS methods to show the significance of our proposed localization method. The methods are evaluated on MagicBrush dataset.} %
  \resizebox{\linewidth}{!}{%
  \begin{tabular}{c|cccccc}
    \toprule
    & \textbf{Method} & L1~$\downarrow$  & L2~ $\downarrow$  & CLIP-I~$\uparrow$ & DINO~$\uparrow$  & CLIP-T~$\uparrow$ \\ \midrule
      & IP2P~\cite{Brooks2022InstructPix2Pix}  & 0.112 & 0.037 & 0.852 & 0.743 & 0.276 \\ \midrule
      
        \multirow{3}{*}{\rotatebox[origin=c]{90}{%
            \begin{tabular}{@{}c@{}}
                \textbf{Mask}  \\
                \textbf{Type}
            \end{tabular}%
        }} & OV-SEG~\cite{liang2023open}  & 0.084 & 0.031 & 0.887 & 0.851 & 0.283 \\
        &SEEM~\cite{zou2024segment}  & \underline{0.079} & \underline{0.022} & \underline{0.903} & \underline{0.866} & \underline{0.289} \\
        &IP2P + \model{}  & \textbf{0.058} & \textbf{0.017} & \textbf{0.935} & \textbf{0.906} & \textbf{0.293} \\
        
      \bottomrule
      \end{tabular}
      }

\label{tab:ablation_ovs}

\end{table}

%% file: tex_figures/ovs_RoI.tex
\begin{figure}[!ht]
    
    \centering

    \small{
        \begin{minipage}{.24\linewidth}
            \centering
            \textbf{Input Image}
        \end{minipage}
        \hfill
        \begin{minipage}{.72\linewidth}
            \centering
            \hfill
            \begin{minipage}{.31\linewidth}
                \centering
                \textbf{OV-SEG~\cite{liang2023open}}
            \end{minipage}
            \hfill
            \begin{minipage}{.31\linewidth}
                \centering
                \textbf{SEEM~\cite{zou2024segment}}
            \end{minipage}
            \hfill
            \begin{minipage}{.31\linewidth}
                \centering
                \textbf{Ours}
            \end{minipage}
            \hfill
        \end{minipage}
    }

    \footnotesize

    \vspace{0.4em}

    \begin{minipage}{.24\linewidth}
        \centering
        \includegraphics[width=\textwidth]{figures/vqgan/outfit/input.jpg}
    \end{minipage}
    \hfill
    \begin{minipage}{.72\linewidth}
        \centering
        \includegraphics[width=\textwidth]{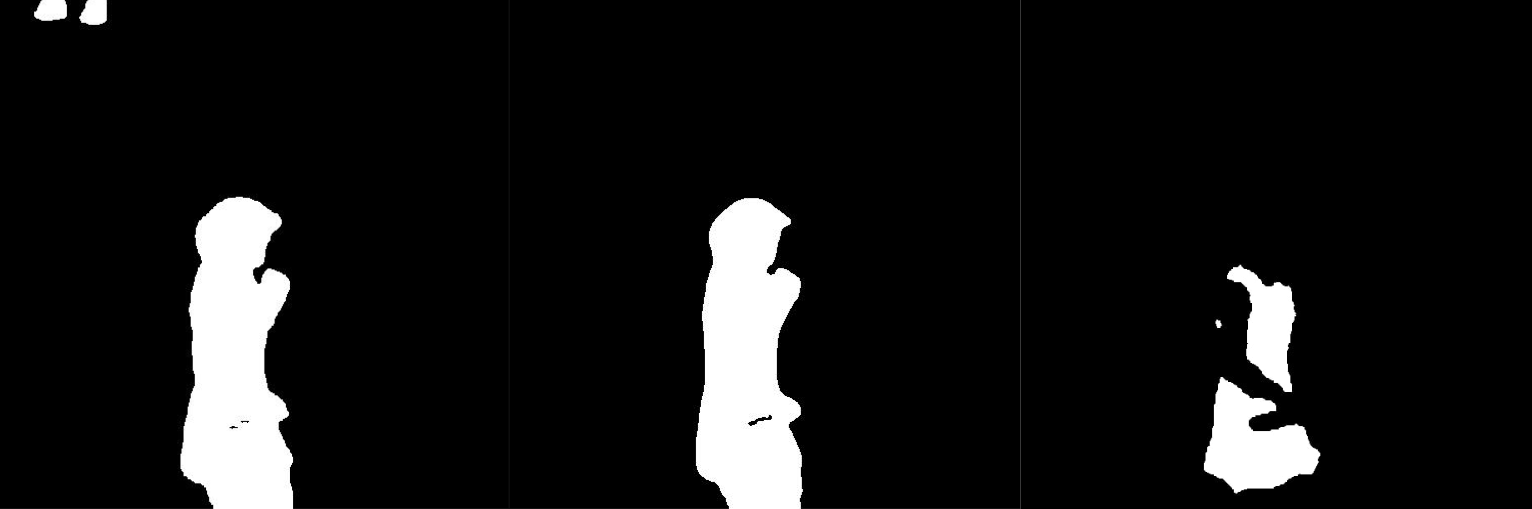}
    \end{minipage}

    \vspace{0.4em}

    \begin{minipage}{.24\linewidth}
        \centering
        \phantom{\textit{a}}
    \end{minipage}
    \hfill
    \begin{minipage}{.72\linewidth}
        \centering
        \textit{Make her \textbf{outfit} black.}
    \end{minipage}

    \vspace{0.4em}

    \begin{minipage}{.24\linewidth}
        \centering
        \includegraphics[width=\textwidth]{figures/qualitative/ip2p/fingernails/input.jpg}
    \end{minipage}
    \hfill
    \begin{minipage}{.72\linewidth}
        \centering
        \includegraphics[width=\textwidth]{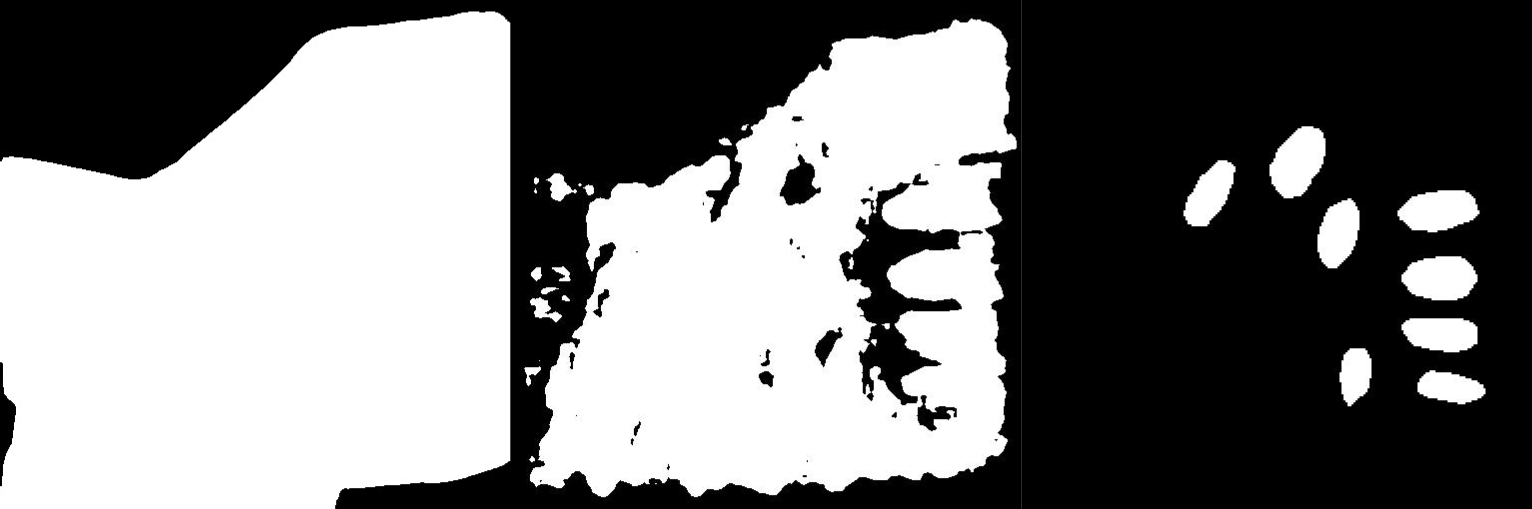}
    \end{minipage}

    \vspace{0.4em}

    \begin{minipage}{.24\linewidth}
        \centering
        \phantom{\textit{a}}
    \end{minipage}
    \hfill
    \begin{minipage}{.72\linewidth}
        \centering
        \textit{Put blue glitter on \textbf{fingernails}.}
    \end{minipage}

    \caption{\textbf{OVS Method for Localization.} RoI examples that are found by different OVS methods, OV-SEG~\cite{liang2023open} and SEEM~\cite{zou2024segment}, and our proposal in \cref{sec:edit_localization}. Since OVS methods require \textit{object of interest}, words with \textbf{bold} are used for the \textit{object of interest}.}
    \label{fig:ovs_roi}
\end{figure}

%% file: tables/ablation2.tex
\begin{table}[!ht]
  \centering
    \caption{\textbf{Ablation Study on Token Selection.} For fair comparison, all parameters are the same for all settings except the ablated parameter.} %

  \resizebox{0.9\linewidth}{!}{%
  \begin{tabular}{cccccc}
    \toprule
    \textbf{Method} & L1~$\downarrow$  & L2~ $\downarrow$  & CLIP-I~$\uparrow$ & DINO~$\uparrow$  & CLIP-T~$\uparrow$ \\ \midrule
    IP2P~\cite{Brooks2022InstructPix2Pix}  & 0.112 & 0.037 & 0.852 & 0.743 & 0.276 \\ \midrule
    Related  & \underline{0.065}& \underline{0.018}& \underline{0.930}& \underline{0.897}& \underline{0.292}\\ 
    Unrelated  & \textbf{0.058} & \textbf{0.017} & \textbf{0.935} & \textbf{0.906} & \textbf{0.293} \\ 
        
      \bottomrule
      \end{tabular}
      }

\label{tab:ablation2}
\end{table}

%% file: tex_figures/ablation_tokens.tex
\begin{figure}[ht!]
    \centering
    \footnotesize

    \begin{minipage}{\linewidth}
        \centering
        \textbf{Token-based cross-attention probabilities}
    \end{minipage}

    \vspace{0.2em}

    \begin{minipage}{.03\linewidth}
        \centering
        \rotatebox{90}{\textbf{Unrelated}}
    \end{minipage}
    \begin{minipage}{.9\linewidth}
        \centering
        \includegraphics[width=\textwidth]{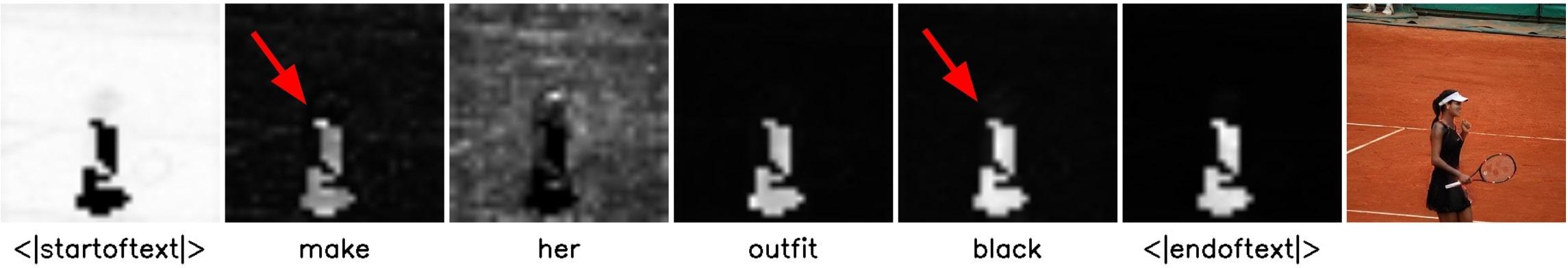}
    \end{minipage}

    \vspace{0.5em}

    \begin{minipage}{.03\linewidth}
        \centering
        \rotatebox{90}{\textbf{Related}}
    \end{minipage}
    \begin{minipage}{.9\linewidth}
        \centering
        \includegraphics[width=\textwidth]{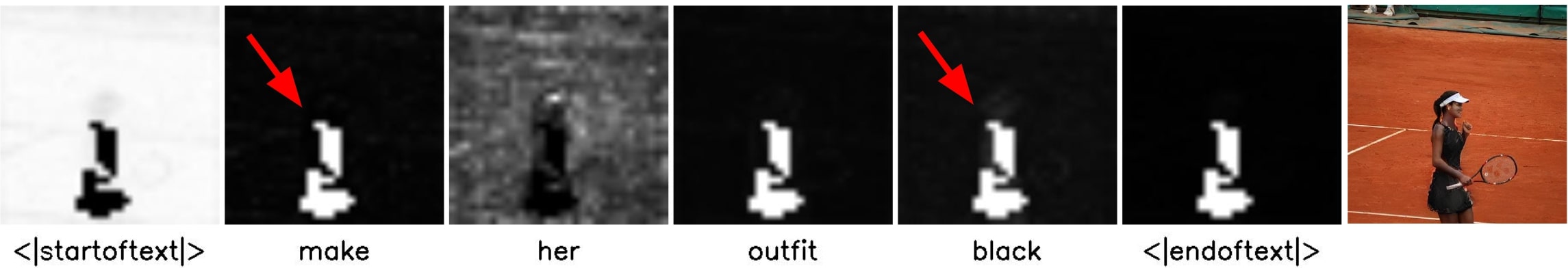}
    \end{minipage}

    \caption{\textbf{Ablation study on related token rewarding.} Instead of regularizing the unrelated tokens, see the first row, we ablate the opposite which is rewarding the related tokens by giving + $\infty$ within RoI, see the second row.}

    \label{fig:ablation_tokens}
\end{figure}

%% file: tables/ablation_clustering.tex
\begin{table}[!htb]

  \centering
  \caption{\textbf{Ablation Study on Clustering Method.} For all experiments, IP2P is the base architecture and the evaluation is on the MagicBrush dataset. Each parameter is modified separately, while other parameters are kept fixed to isolate their impact. }
  \resizebox{\linewidth}{!}{%
  \begin{tabular}{c|cccccc}
    \toprule
    & \textbf{Method} & L1~$\downarrow$  & L2~ $\downarrow$  & CLIP-I~$\uparrow$ & DINO~$\uparrow$  & CLIP-T~$\uparrow$ \\ \midrule
      & IP2P~\cite{Brooks2022InstructPix2Pix}  & 0.112 & 0.037 & 0.852 & 0.743 & 0.276 \\ \midrule

        \multirow{4}{*}{\rotatebox[origin=c]{90}{%
            \begin{tabular}{@{}c@{}}
                \textbf{\# of}  \\
                \textbf{Clusters}
            \end{tabular}%
        }} & $4$  & 0.080 & 0.022 & 0.923 & 0.885 & \textbf{0.295} \\
        & $8$  & \textbf{0.058} & \textbf{0.017} & \textbf{0.935} & \textbf{0.906} & 0.293 \\
        & $16$ & \underline{0.062} & \underline{0.018} & \underline{0.933} & \underline{0.903} & \underline{0.294} \\
        & $32$ & 0.064 & \underline{0.018} & 0.932 & 0.901 & 0.291 \\  
        
        \midrule

        \multirow{4}{*}{\rotatebox[origin=c]{90}{%
            \begin{tabular}{@{}c@{}}
                \textbf{Distance}  \\
                \textbf{Threshold}
            \end{tabular}%
        }} & $0.7$  & 0.080 & 0.022 & 0.927 & 0.893 & \textbf{0.294} \\
        & $0.6$  & {0.076} & {0.021} & {0.929} & {0.894} & \textbf{0.294} \\
        & $0.5$ & \textbf{0.063} & \textbf{0.018} & \textbf{0.933} & \textbf{0.902} & \underline{0.293} \\
        & $0.4$ & \underline{0.072} & \underline{0.020} & \underline{0.930} & \underline{0.896} & \underline{0.293} \\  
        
      \bottomrule
      \end{tabular}
      }

\label{tab:ablation_clustering}

\end{table}

%% file: tex_figures/usecase.tex
\begin{figure}[!htp]
    \centering

    \begin{minipage}{.3\linewidth}
        \centering
        \textbf{Input Image} %
    \end{minipage}
    \begin{minipage}{.3\linewidth}
        \centering
        \textbf{Baseline}
    \end{minipage}
    \begin{minipage}{.3\linewidth}
        \centering
        \textbf{+ \model{}}
    \end{minipage}

    \vspace{0.6em}

    \begin{minipage}{.3\linewidth}
        \centering
        \includegraphics[width=\textwidth]{figures/more_qualitative/livingroom/original.jpg}
    \end{minipage}
    \begin{minipage}{.3\linewidth}
        \centering
        \includegraphics[width=\textwidth]{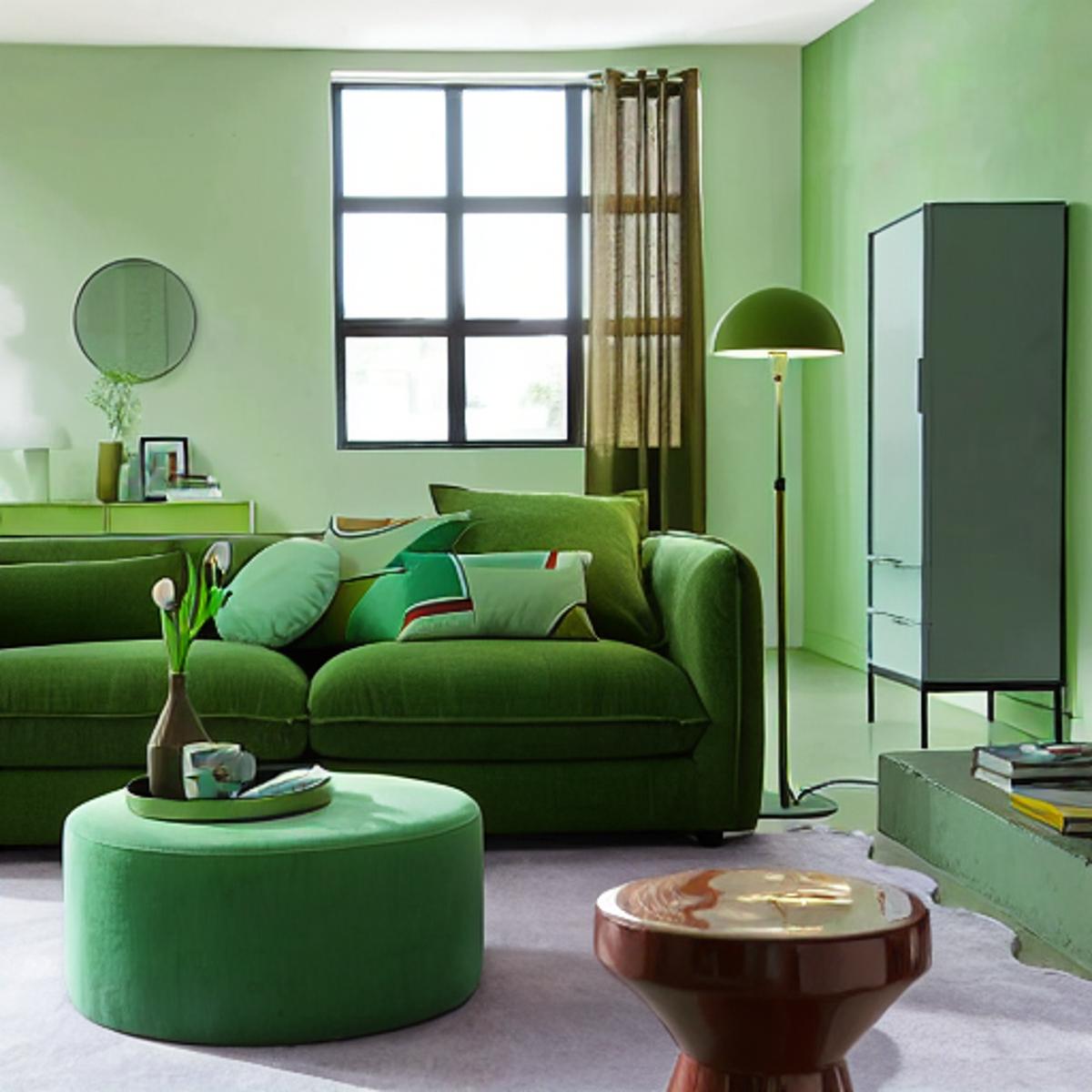}
    \end{minipage}
    \begin{minipage}{.3\linewidth}
        \centering
        \includegraphics[width=\textwidth]{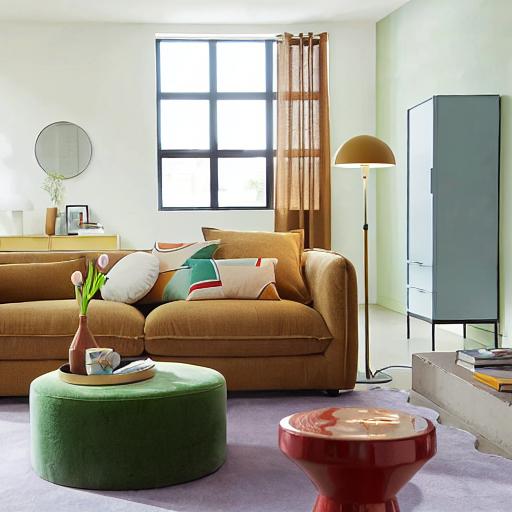}
    \end{minipage}

    \vspace{0.6em}

    \begin{minipage}{.3\linewidth}
        \centering
        {(a)}
    \end{minipage}
    \begin{minipage}{.54\linewidth}
        \centering
        \textit{Change color of \textbf{ottoman} to dark green.}
    \end{minipage} 

    \vspace{0.6em}

    \begin{minipage}{.3\linewidth}
        \centering
        \includegraphics[width=\textwidth]{figures/more_qualitative/livingroom/original.jpg}
    \end{minipage}    
    \begin{minipage}{.3\linewidth}
        \centering
        \includegraphics[width=\textwidth]{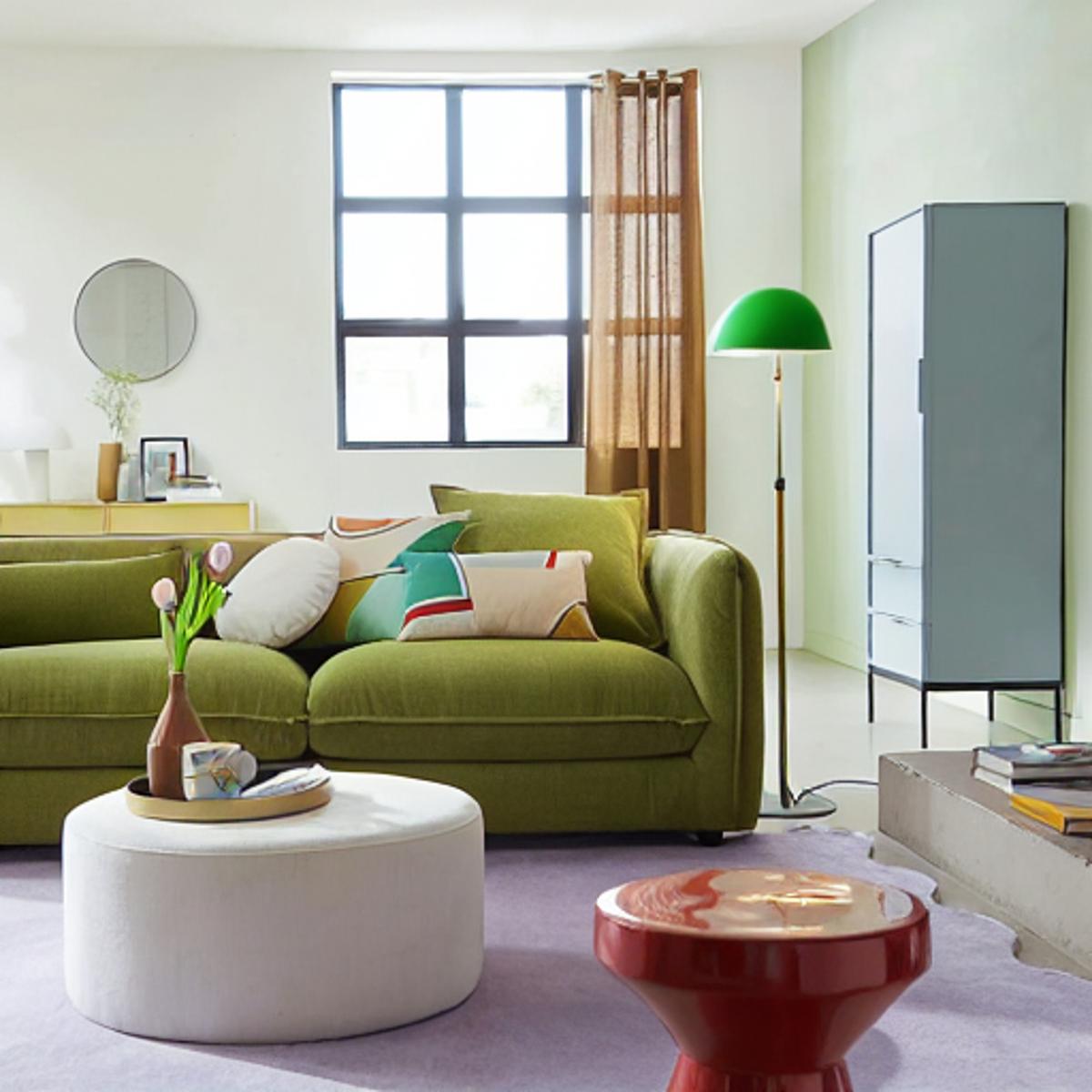}
    \end{minipage}
    \begin{minipage}{.3\linewidth}
        \centering
        \includegraphics[width=\textwidth]{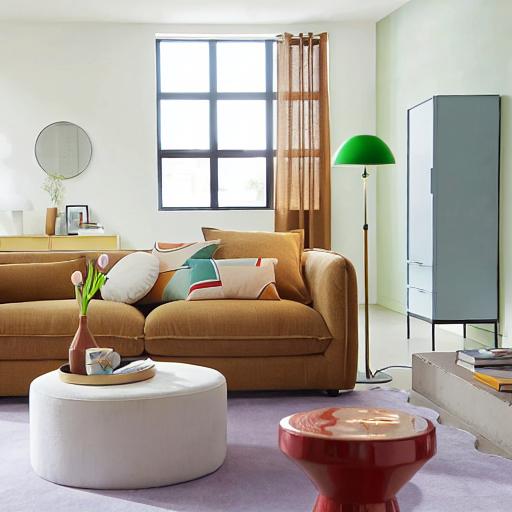}
    \end{minipage}

    \vspace{0.6em}

    \begin{minipage}{.3\linewidth}
        \centering
        {(b)}
    \end{minipage}
    \begin{minipage}{.64\linewidth}
        \centering
        \textit{Change color of \textbf{lamp} to dark green.}
    \end{minipage}

    \vspace{0.6em}

    \begin{minipage}{.3\linewidth}
        \centering
        \includegraphics[width=\textwidth]{figures/more_qualitative/livingroom/original.jpg}
    \end{minipage}    
    \begin{minipage}{.3\linewidth}
        \centering
        \includegraphics[width=\textwidth]{figures/more_qualitative/livingroom/carpet_MB.jpg}
    \end{minipage}
    \begin{minipage}{.3\linewidth}
        \centering
        \includegraphics[width=\textwidth]{figures/more_qualitative/livingroom/carpet_ours.jpg}
    \end{minipage}

    \vspace{0.6em}

    \begin{minipage}{.3\linewidth}
        \centering
        {(c)}
    \end{minipage}
    \begin{minipage}{.64\linewidth}
        \centering
        \textit{Change color of \textbf{carpet} to dark blue.}
    \end{minipage}

    \vspace{0.6em}

    \begin{minipage}{.3\linewidth}
        \centering
        \includegraphics[width=\textwidth]{figures/more_qualitative/livingroom/original.jpg}
    \end{minipage}    
    \begin{minipage}{.3\linewidth}
        \centering
        \includegraphics[width=\textwidth]{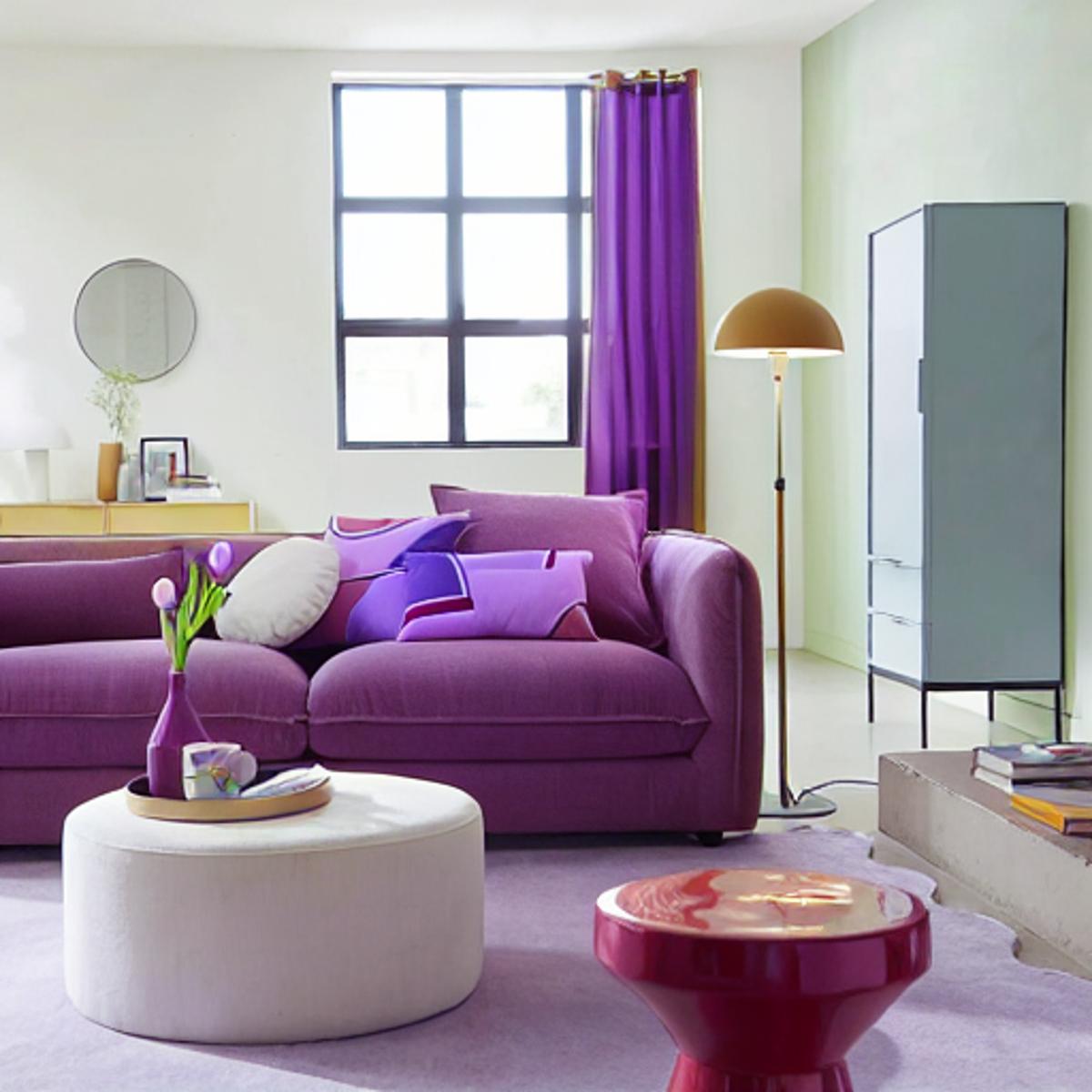}
    \end{minipage}
    \begin{minipage}{.3\linewidth}
        \centering
        \includegraphics[width=\textwidth]{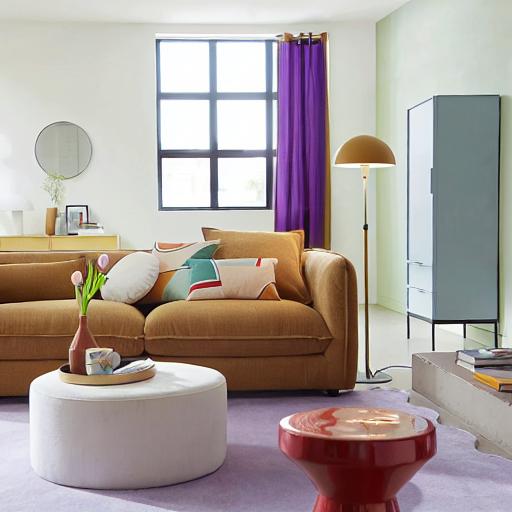}
    \end{minipage}

    \vspace{0.6em}

    \begin{minipage}{.3\linewidth}
        \centering
        {(d)}
    \end{minipage}
    \begin{minipage}{.64\linewidth}
        \centering
        \textit{Change color of \textbf{curtain} to purple.}
    \end{minipage}
    
    \caption{\textbf{A use-case of the proposed method.} Changing the color of different objects is shown by comparing baselines and our method. Our method performs disentangled and localized edits for different colors and different objects in the scene.}
    \vspace{-2.5em}

    \label{fig:usecase}
\end{figure}

%% file: tex_figures/more_qualitatives1.tex
\begin{figure*}[!hp]
    \centering

    \tikz[remember picture,overlay] \node [anchor=base] (linebase) {};

    \begin{minipage}{.19\textwidth}
        \centering
        \textbf{Input Image}
    \end{minipage}
    \begin{minipage}{.19\textwidth}
        \centering
        \textbf{IP2P~\cite{Brooks2022InstructPix2Pix}}
    \end{minipage}
    \begin{minipage}{.19\textwidth}
        \centering
        \textbf{+ \model{}}
    \end{minipage}
    \begin{minipage}{.19\textwidth}
        \centering
        \textbf{IP2P~\cite{Brooks2022InstructPix2Pix} w/MB~\cite{Zhang2023MagicBrush}}
    \end{minipage}
    \begin{minipage}{.19\textwidth}
        \centering
        \textbf{+ \model{}}
    \end{minipage}

    \vspace{0.6em}

    \begin{minipage}{.19\textwidth}
        \centering
        \includegraphics[width=\textwidth]{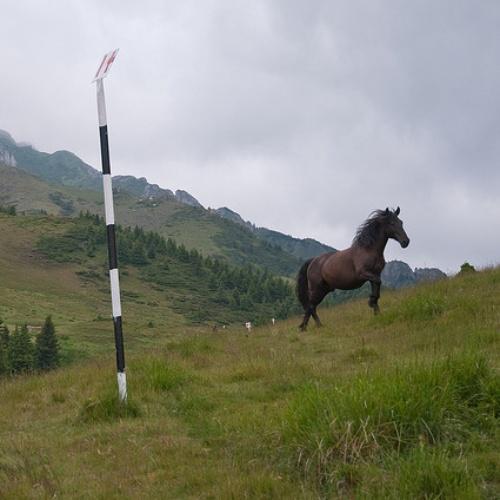}
    \end{minipage}
    \begin{minipage}{.19\textwidth}
        \centering
        \includegraphics[width=\textwidth]{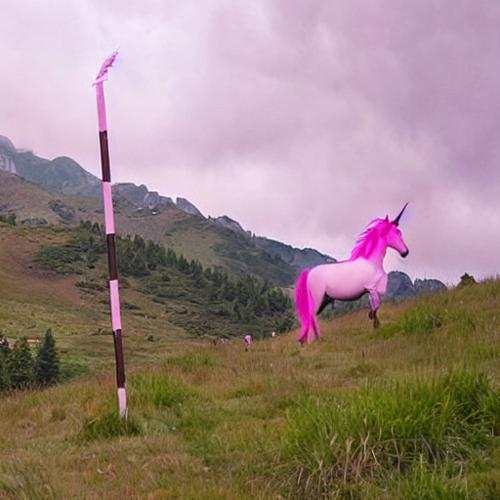}
    \end{minipage}
    \begin{minipage}{.19\textwidth}
        \centering
        \includegraphics[width=\textwidth]{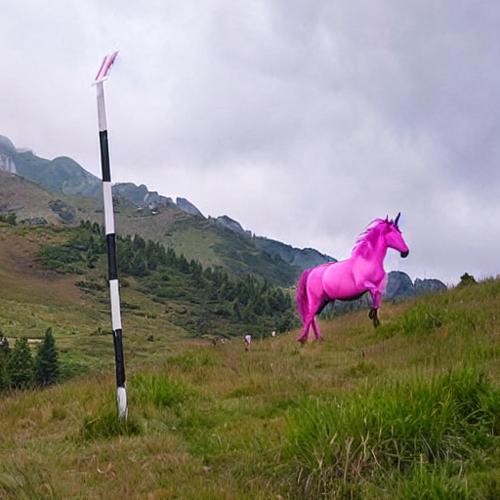}
    \end{minipage}
    \begin{minipage}{.19\textwidth}
        \centering
        \includegraphics[width=\textwidth]{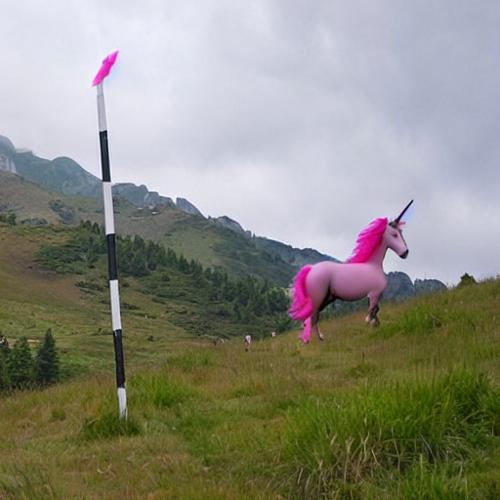}
    \end{minipage}
    \begin{minipage}{.19\textwidth}
        \centering
        \includegraphics[width=\textwidth]{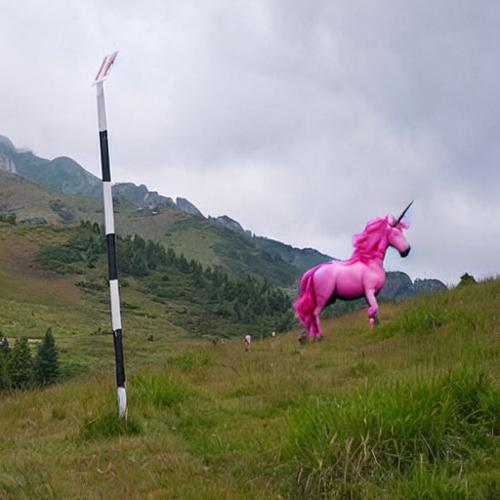}
    \end{minipage}

    \vspace{0.6em}

    \begin{minipage}{.19\textwidth}
        \centering
        (a)
    \end{minipage}
    \begin{minipage}{.76\textwidth}
        \centering
        \textit{Turn the brown horse into a pink unicorn.}
    \end{minipage}

    \vspace{0.6em}

    \begin{minipage}{.19\textwidth}
        \centering
        \includegraphics[width=\textwidth]{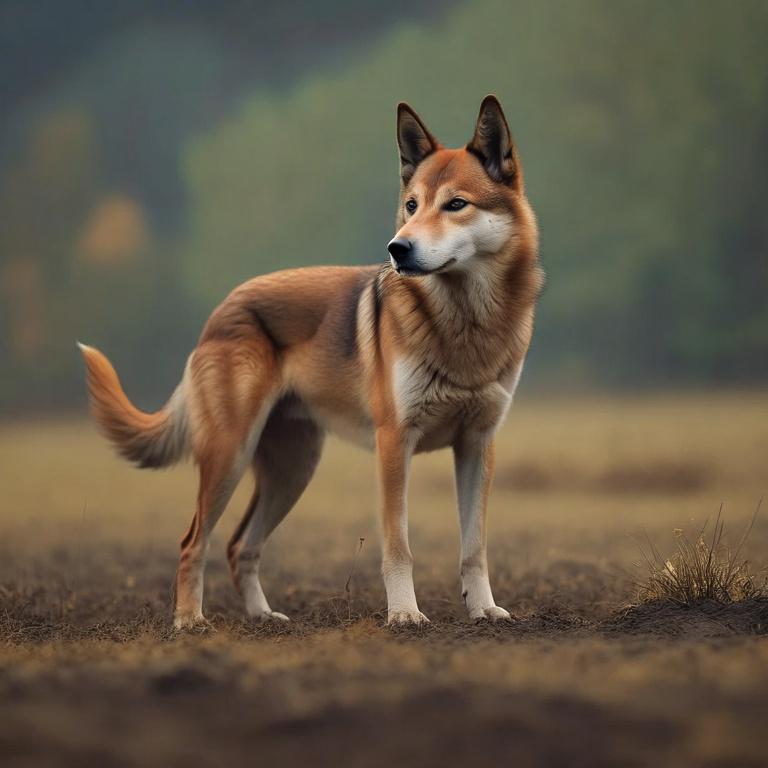}
    \end{minipage}
    \begin{minipage}{.19\textwidth}
        \centering
        \includegraphics[width=\textwidth]{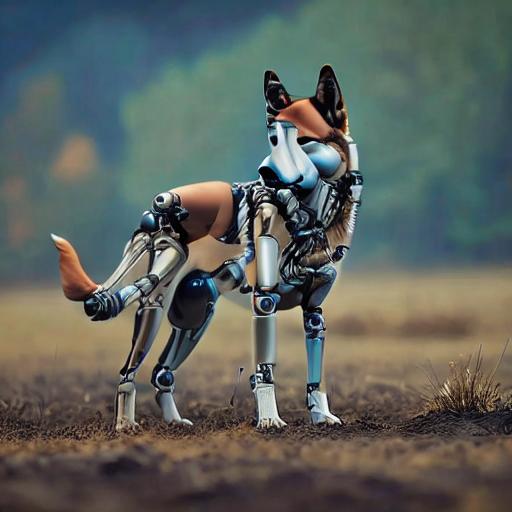}
    \end{minipage}
    \begin{minipage}{.19\textwidth}
        \centering
        \includegraphics[width=\textwidth]{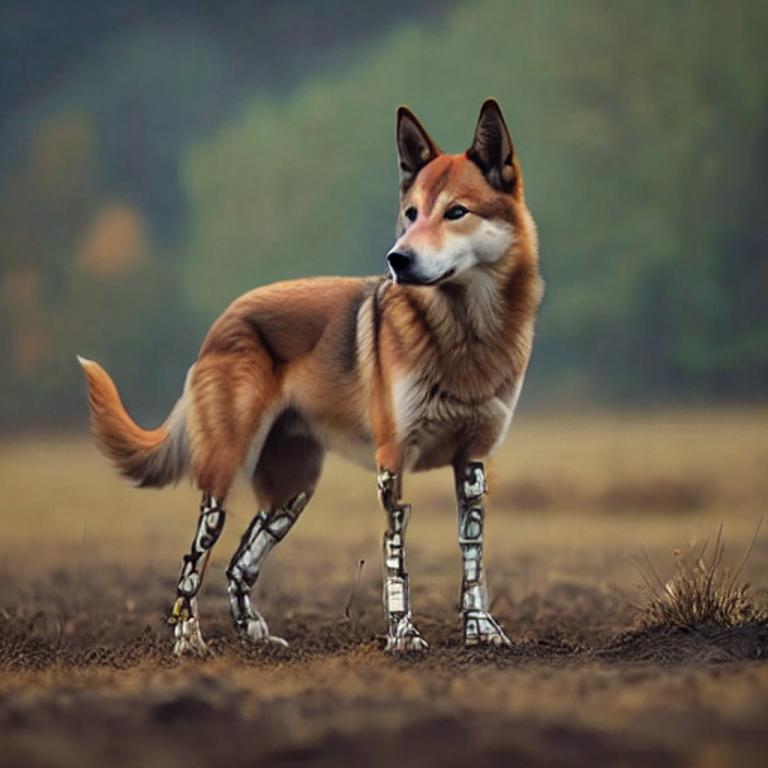}
    \end{minipage}
    \begin{minipage}{.19\textwidth}
        \centering
        \includegraphics[width=\textwidth]{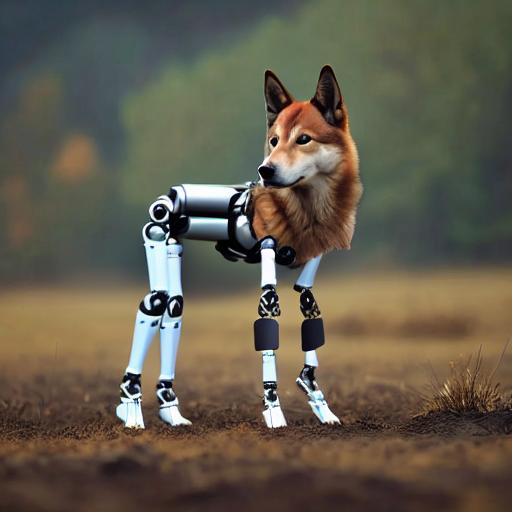}
    \end{minipage}
    \begin{minipage}{.19\textwidth}
        \centering
        \includegraphics[width=\textwidth]{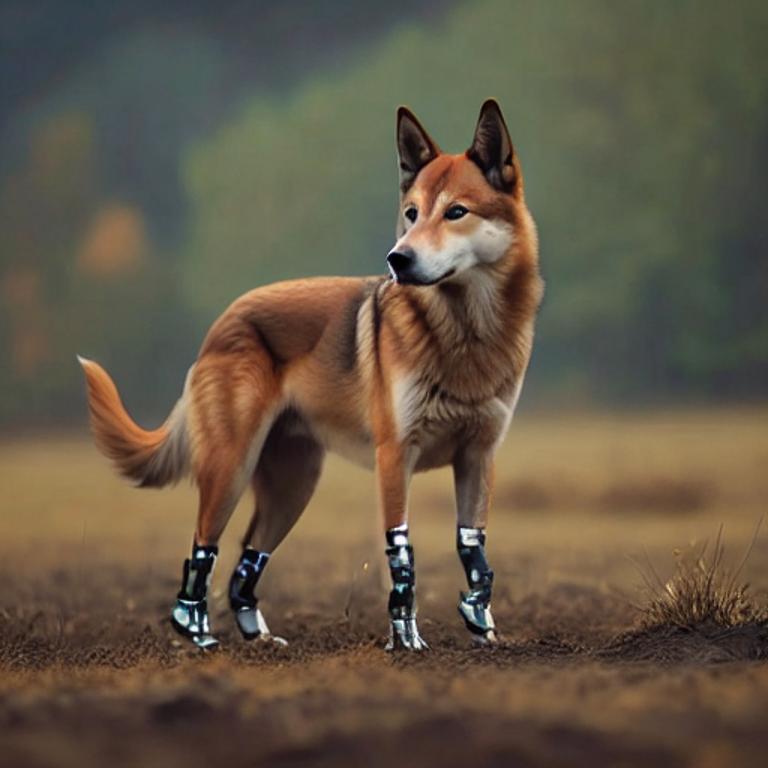}
    \end{minipage}

    \vspace{0.6em}

    \begin{minipage}{.19\textwidth}
        \centering
        (b)
    \end{minipage}
    \begin{minipage}{.76\textwidth}
        \centering
        \textit{Change the legs to be bionic.}
    \end{minipage}

    \vspace{0.6em}

    \begin{minipage}{.19\textwidth}
        \centering
        \includegraphics[width=\textwidth]{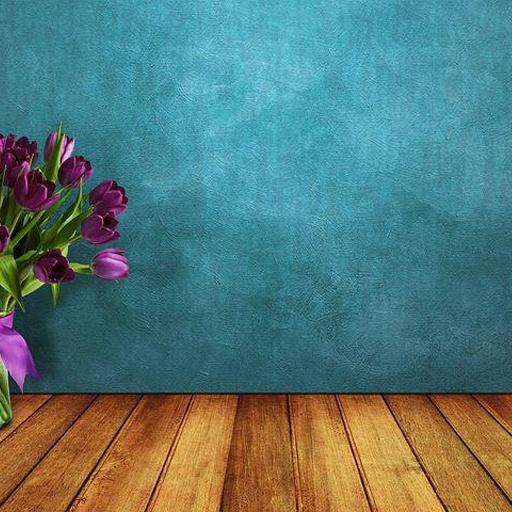}
    \end{minipage}
    \begin{minipage}{.19\textwidth}
        \centering
        \includegraphics[width=\textwidth]{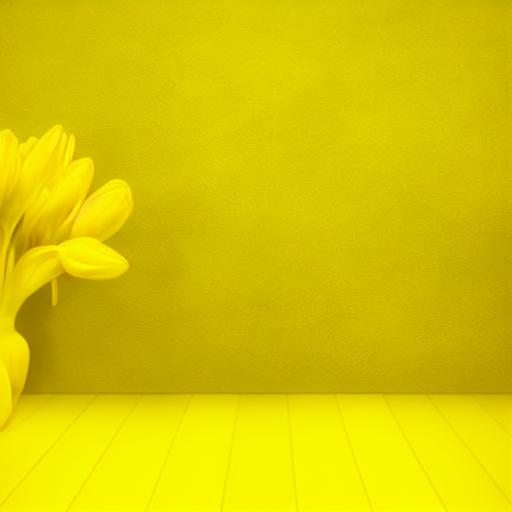}
    \end{minipage}
    \begin{minipage}{.19\textwidth}
        \centering
        \includegraphics[width=\textwidth]{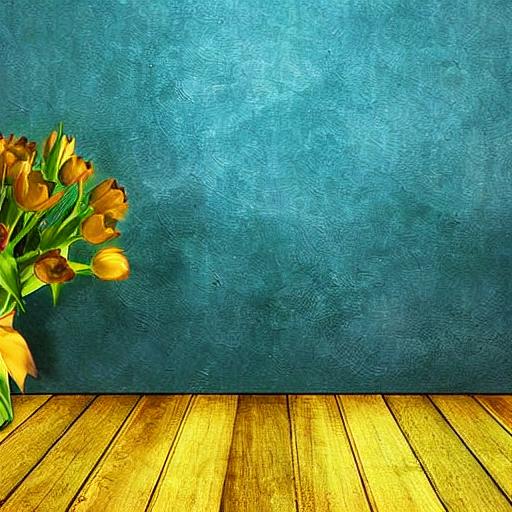}
    \end{minipage}
    \begin{minipage}{.19\textwidth}
        \centering
        \includegraphics[width=\textwidth]{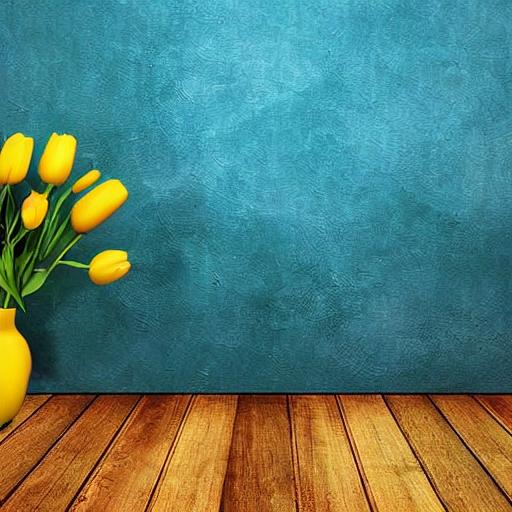}
    \end{minipage}
    \begin{minipage}{.19\textwidth}
        \centering
        \includegraphics[width=\textwidth]{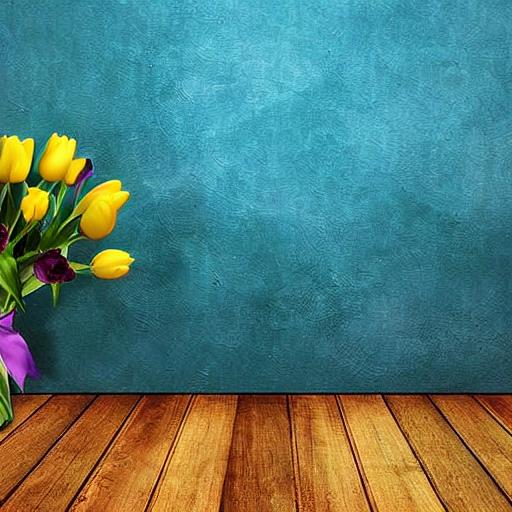}
    \end{minipage}

    \vspace{0.6em}

    \begin{minipage}{.19\textwidth}
        \centering
        (c)
    \end{minipage}
    \begin{minipage}{.76\textwidth}
        \centering
        \textit{Change the color of the tulips to yellow.}
    \end{minipage}

    \caption{\textbf{More Qualitative Examples.} We test our method on different tasks: (a) replacing an animal, (b) editing animal parts, and (c) changing the color of multiple objects. 
    The integration of \model{} enhances the performance of all models, enabling localized edits while maintaining the integrity of the remaining image areas.}
    \label{fig:more_qualitatives1}
\end{figure*}